\documentclass[10pt,twocolumn,letterpaper]{article}

\usepackage{cvpr}              %

\usepackage{paralist}

\definecolor{rblue}{rgb}{0,0.5,1}
\definecolor{awesome}{rgb}{1.0, 0.13, 0.32}
\definecolor{hollywoodcerise}{rgb}{0.96, 0.0, 0.63}
\definecolor{lasallegreen}{rgb}{0.03, 0.47, 0.19}
\definecolor{hanpurple}{rgb}{0.32, 0.09, 0.98}
\definecolor{green(pigment)}{rgb}{0.0, 0.65, 0.31}

\usepackage{pifont}
\newcommand{\cmark}{\ding{51}} 
\newcommand{\xmark}{\ding{55}} 

\usepackage{booktabs}
\usepackage{multirow}
\usepackage{colortbl}
\usepackage{makecell}
\usepackage{graphicx}
\usepackage{threeparttable} 

\usepackage{amsmath}
\usepackage[linesnumbered,ruled,vlined]{algorithm2e}
\usepackage{listings}

\usepackage{booktabs}
\usepackage{graphicx}
\usepackage{multirow}
\usepackage{makecell}
\usepackage{amssymb}
\newcommand{\car}{\includegraphics[width=4mm]{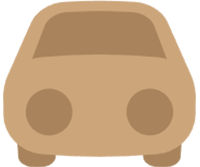}}
\newcommand{\robot}{\includegraphics[width=3.5mm]{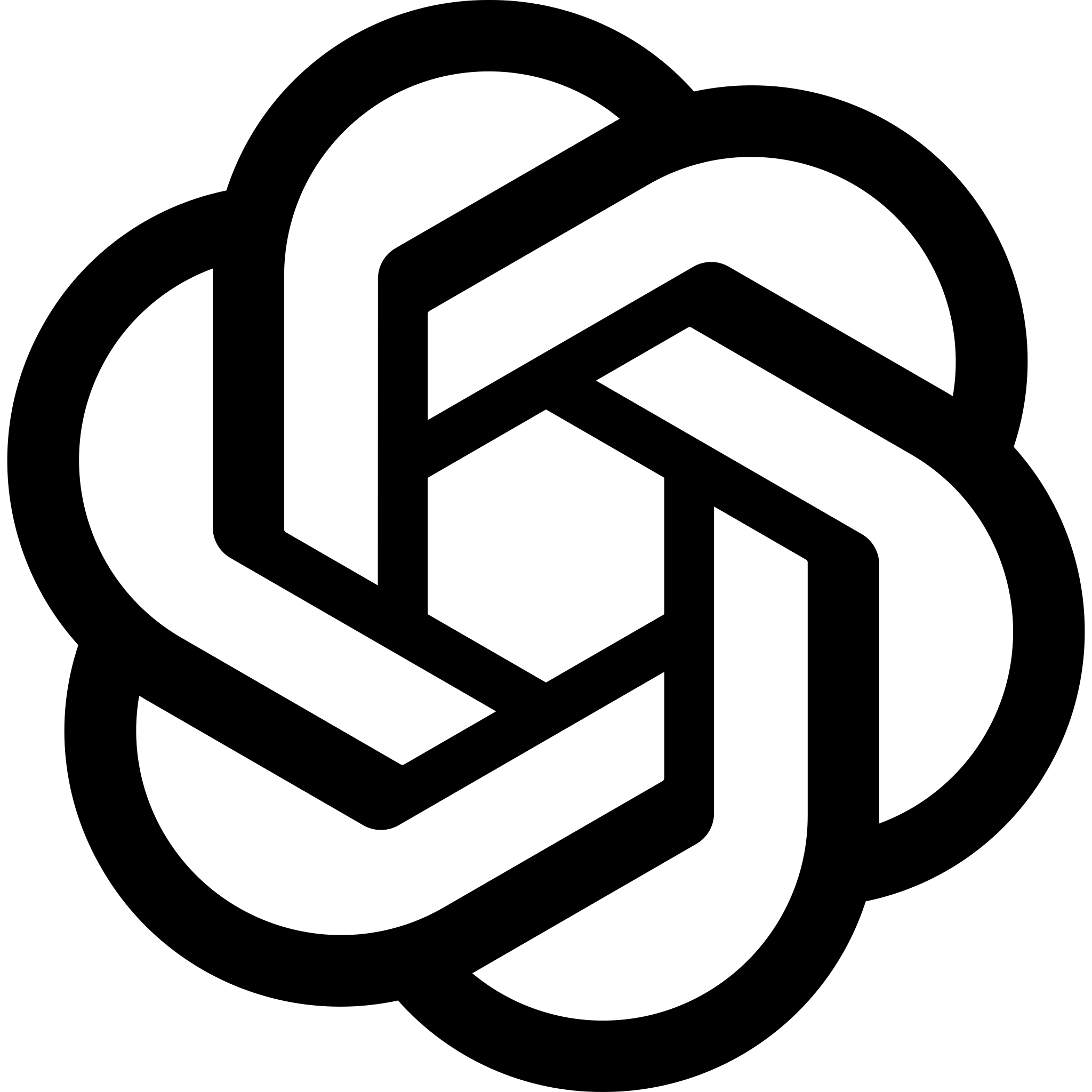}}
\newcommand{\robotdog}{\includegraphics[width=4mm]{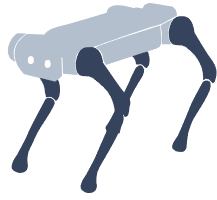}}
\newcommand{\webm}{\includegraphics[width=3.6mm]{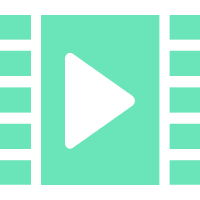}}
\newcommand{\mywebm}{\includegraphics[width=3.5mm]{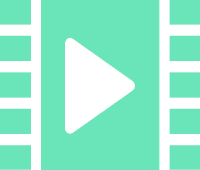}}
\newcommand{\wheels}{\includegraphics[width=4mm]{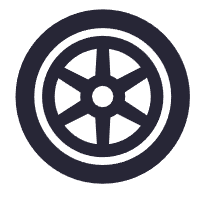}}
\newcommand{\gait}{\includegraphics[width=4mm]{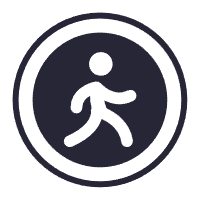}}
\newcommand{\stationary}{\includegraphics[width=4mm]{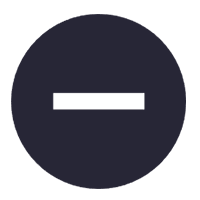}}
\newcommand{\mycheckmark}{\includegraphics[width=3.5mm]{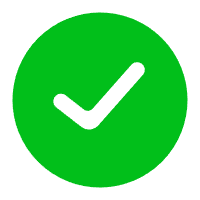}}
\newcommand{\crossmark}{\includegraphics[width=3.5mm]{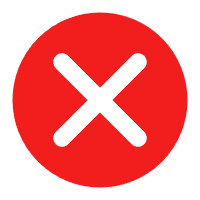}}
\newcommand{\sun}{\includegraphics[width=3.5mm]{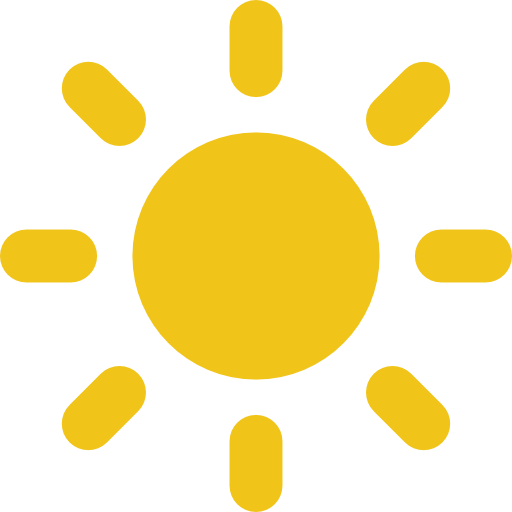}}
\newcommand{\dusk}{\includegraphics[width=3.5mm]{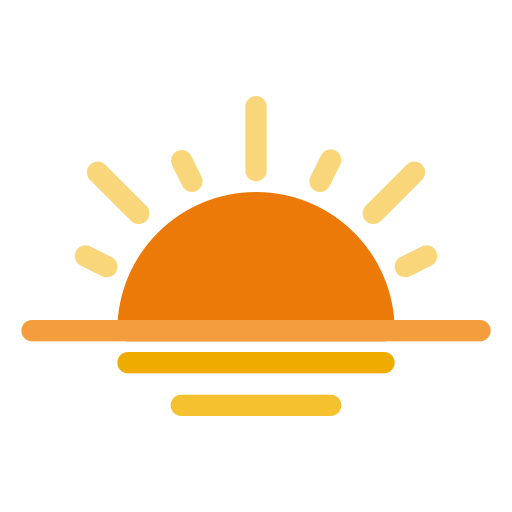}}
\newcommand{\night}{\includegraphics[width=3.5mm]{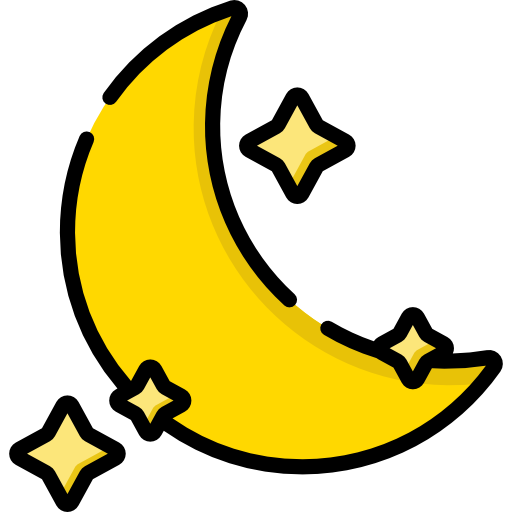}}
\newcommand{\human}{\includegraphics[width=3.5mm]{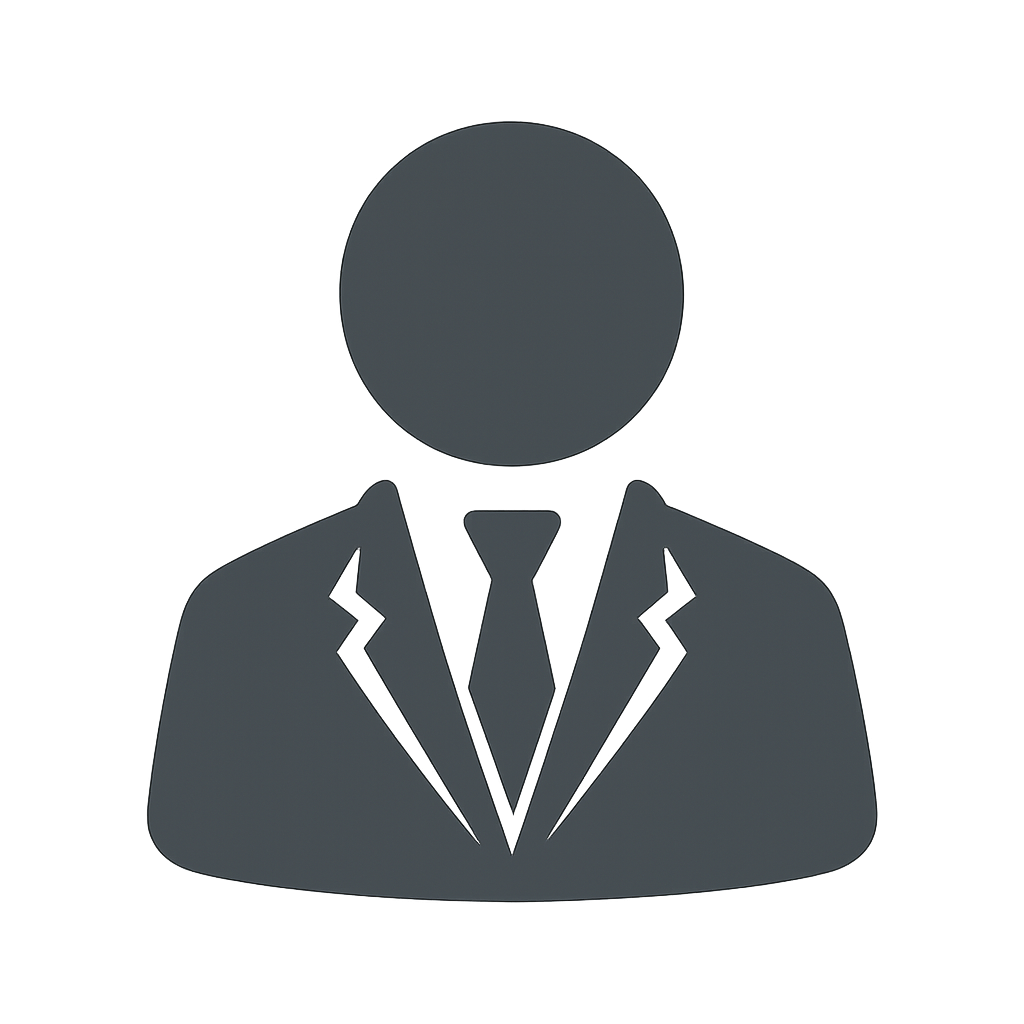}}
\newcommand{\syn}{\includegraphics[width=3.5mm]{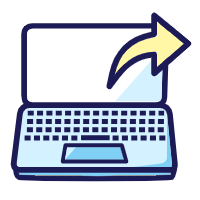}}
\newcommand{\topline}{\noalign{\hrule height 0.8 pt}} 
 
\newcommand{\bottomline}{\noalign{\hrule height 0.8 pt}} 
\definecolor{poster_color1}{rgb}{1,1,0.949}
\definecolor{poster_color2}{rgb}{0.925,0.980,0.898}
\definecolor{codebackgroundcolor}{RGB}{248,242,222}
\definecolor{colorVehicle}{RGB}{100, 150, 245}
\definecolor{colorPedestrian}{RGB}{255, 30, 30}
\definecolor{colorRoad}{RGB}{255, 0, 255}
\definecolor{colorBuilding}{RGB}{255, 200, 0}
\definecolor{colorVegetation}{RGB}{0, 175, 0}
\definecolor{colorTerrain}{RGB}{150, 240, 80}
\makeatletter
\newcommand{\vehicle@quadoccfreq}{0.90}
\newcommand{\pedestrian@quadoccfreq}{2.09} 
\newcommand{\road@quadoccfreq}{52.34}
\newcommand{\building@quadoccfreq}{14.00}
\newcommand{\vegetation@quadoccfreq}{22.82}
\newcommand{\terrain@quadoccfreq}{7.85}
\newcommand{\quadoccfreq}[1]{{\csname #1@quadoccfreq\endcsname}}
\makeatother

\definecolor{hthreeoRoad}{RGB}{128,64,128}        
\definecolor{hthreeoSidewalk}{RGB}{244,35,232}    
\definecolor{hthreeoBuilding}{RGB}{70,70,70}      
\definecolor{hthreeoVegetation}{RGB}{107,142,35}  
\definecolor{hthreeoCar}{RGB}{0,0,142}            
\definecolor{hthreeoTruck}{RGB}{0,0,70}           
\definecolor{hthreeoBus}{RGB}{0,60,100}           
\definecolor{hthreeoTwoWheeler}{RGB}{0,0,230}     
\definecolor{hthreeoPerson}{RGB}{220,20,60}       
\definecolor{hthreeoPole}{RGB}{153,153,153}       

\makeatletter
\newcommand{\road@hthreeofreqHomo}{34.04}
\newcommand{\sidewalk@hthreeofreqHomo}{26.24}
\newcommand{\building@hthreeofreqHomo}{10.00}
\newcommand{\vegetation@hthreeofreqHomo}{24.02}
\newcommand{\car@hthreeofreqHomo}{2.92}
\newcommand{\truck@hthreeofreqHomo}{0.99}
\newcommand{\bus@hthreeofreqHomo}{0.25}
\newcommand{\twoWheeler@hthreeofreqHomo}{0.07}
\newcommand{\person@hthreeofreqHomo}{0.97}
\newcommand{\pole@hthreeofreqHomo}{0.50}
\newcommand{\hthreeofreqHomo}[1]{{\csname #1@hthreeofreqHomo\endcsname}}

\newcommand{\road@hthreeofreqHeter}{34.04}
\newcommand{\sidewalk@hthreeofreqHeter}{26.24}
\newcommand{\building@hthreeofreqHeter}{10.00}
\newcommand{\vegetation@hthreeofreqHeter}{24.02}
\newcommand{\car@hthreeofreqHeter}{2.92}
\newcommand{\truck@hthreeofreqHeter}{0.99}
\newcommand{\bus@hthreeofreqHeter}{0.25}
\newcommand{\twoWheeler@hthreeofreqHeter}{0.07}
\newcommand{\person@hthreeofreqHeter}{0.97}
\newcommand{\pole@hthreeofreqHeter}{0.50}
\newcommand{\hthreeofreqHeter}[1]{{\csname #1@hthreeofreqHeter\endcsname}}
\makeatother

\newcommand{\ColHeadHomo}[3]{\rotatebox{90}{\textcolor{#1}{$\blacksquare$} \makecell[l]{#2 \\ (\hthreeofreqHomo{#3}\%)}}}
\newcommand{\ColHeadHeter}[3]{\rotatebox{90}{\textcolor{#1}{$\blacksquare$} \makecell[l]{#2 \\ (\hthreeofreqHeter{#3}\%)}}}
\usepackage{xspace}
\makeatletter
\DeclareRobustCommand\onedot{\futurelet\@let@token\@onedot}
\def\@onedot{\ifx\@let@token.\else.\null\fi\xspace}
\def\eg{\emph{e.g}\onedot}

\makeatother

\usepackage[most]{tcolorbox} 
\newtcolorbox{bluequestion}{
  enhanced, breakable,
  colback=blue!5,        
  colframe=blue!30,      
  boxrule=0pt,           
  borderline west={6pt}{0pt}{blue!60}, 
  left=8pt, right=8pt, top=6pt, bottom=6pt, 
  arc=0pt                
}

\usepackage{tikz}
\usetikzlibrary{positioning,arrows.meta,fit,calc}
\usepackage{capt-of}   %
\usepackage{cuted}
\definecolor{cvprblue}{rgb}{0.21,0.49,0.74}
\usepackage[pagebackref,breaklinks,colorlinks,allcolors=cvprblue]{hyperref}

\makeatletter
\renewcommand*{\@fnsymbol}[1]{\ensuremath{\ifcase#1\or *\or \dagger\or \ddagger\or
    \mathsection\or \mathparagraph\or \|\or **\or \dagger\dagger
    \or \ddagger\ddagger \else\@ctrerr\fi}}
\makeatother

\title{OneOcc: Semantic Occupancy Prediction for Legged Robots\\with a Single Panoramic Camera}

\author{Hao Shi$^{1,2,}$\thanks{Equal contribution.} \quad Ze Wang$^{1,3,*}$ \quad Shangwei Guo$^{1,*}$ \quad Mengfei Duan$^{4}$ \quad Song Wang$^{1}$ \quad Teng Chen$^{5}$\\Kailun Yang$^{4,}$\thanks{Corresponding authors (e-mail: {\tt kailun.yang@hnu.edu.cn, linwang@ntu.edu.sg, wangkaiwei@zju.edu.cn}).} \quad Lin Wang$^{2,\dag}$ \quad Kaiwei Wang$^{1,\dag}$\\
$^{1}$ZJU \quad $^{2}$NTU \quad $^{3}$MirrorMe Technology \quad $^{4}$HNU \quad $^{5}$Xiaomi Corporation
}

\begin{document}
\maketitle

\begin{abstract}
\vskip -0.5ex
Robust 3D semantic occupancy is crucial for legged/humanoid robots, yet most semantic scene completion (SSC) systems target wheeled platforms with forward-facing sensors. We present \textbf{OneOcc}, a vision-only panoramic SSC framework designed for gait-introduced body jitter and $360^{\circ}$ continuity. OneOcc combines: (i) {Dual-Projection fusion (DP-ER)} to exploit the annular panorama and its equirectangular unfolding, preserving $360^{\circ}$ continuity and grid alignment; (ii) Bi-Grid Voxelization (BGV) to reason in Cartesian and cylindrical-polar spaces, reducing discretization bias and sharpening free/occupied boundaries; (iii) a lightweight decoder with {Hierarchical AMoE-3D} for dynamic multi-scale fusion and better long-range/occlusion reasoning; and (iv) plug-and-play {Gait Displacement Compensation (GDC)} learning feature-level motion correction without extra sensors. We also release two panoramic occupancy benchmarks: \textbf{QuadOcc} (real quadruped, first-person $360^{\circ}$) and \textbf{Human360Occ (H3O)} (CARLA human-ego $360^{\circ}$ with RGB, Depth, semantic occupancy; standardized within-/cross-city splits). OneOcc sets a new state of the art on QuadOcc, outperforming strong vision baselines and remaining competitive with classical LiDAR baselines; on H3O it gains +3.83 mIoU (within-city) and +8.08 (cross-city). Modules are lightweight, enabling deployable full-surround perception for legged/humanoid robots. Datasets and code will be publicly available at \href{https://github.com/MasterHow/OneOcc}{OneOcc}.
\vskip -1ex
\end{abstract}

\section{Introduction}
\label{sec:intro}
\begin{figure}[!t]
\centering
\includegraphics[width=0.95\linewidth]{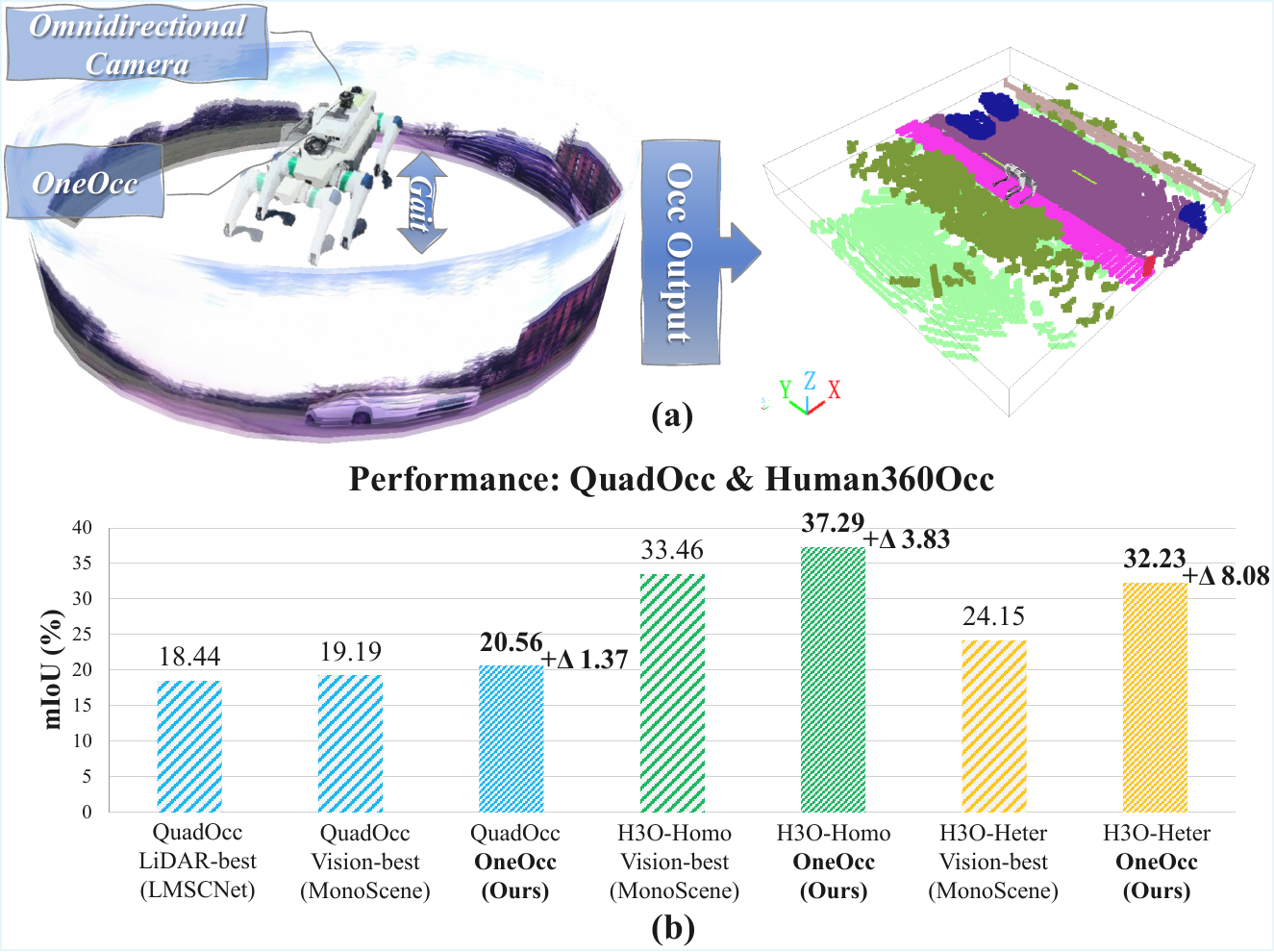}
\vskip -2.5ex
\caption{\textbf{Teaser.}
\textbf{(a)} Single-sensor panoramic semantic scene completion (SSC) on a quadruped: an omnidirectional camera observes full \(360^{\circ}\) with gait-induced motion; \emph{OneOcc} outputs voxelized semantic occupancy.
\textbf{(b)} Summary on \emph{QuadOcc} and \emph{Human360Occ (H3O)}: bars show mIoU(\%). 
On \emph{QuadOcc}, OneOcc reaches 20.56 mIoU, surpassing the best LiDAR baseline (LMSCNet, \(18.44\)) and vision baseline (MonoScene, \(19.19\)).
On \emph{H3O}, OneOcc attains 37.29/32.23 mIoU under within-/cross-city, outperforming the best vision baselines (\(33.46/24.15\)).}
\vskip -4.5ex
\label{fig:teaser}
\end{figure}

\vskip -0.5ex
A holistic 3D understanding of nearby surroundings is fundamental for robotic autonomy in navigation~\cite{wang2024omega,hoeller2022neural,wang2024he_nav,wang2024agrnav,patel2025hierarchical,li2024stereonavnet}, locomotion~\cite{niijima2025realtime_multi_plane,dong2025marg,escontrela2025gaussgym,chen2025learning_traversal}, manipulation~\cite{kartmann2021semantic,dengler2025efficient_manipulation,li2025manipdreamer3d,wang2025odyssey}, and human-robot interaction~\cite{wu2025embodiedocc,wang2025embodiedocc++,liu2024revisit_human_scene_interaction}.
Among 3D perception tasks, \emph{Semantic Scene Completion} (SSC)---predicting complete volumetric geometry and semantics from partial observations---has become central~\cite{lai2024rtonet,cao2024slcf_sequential,yan2024pointssc,zhang2025roboocc}, with most progress driven by autonomous driving where forward-facing sensors on stable wheeled platforms yield benign motion and sensing.

\begin{table*}
  \centering
  \small
  \resizebox{0.975\textwidth}{!}{
  \setlength\tabcolsep{6pt}  
  \renewcommand\arraystretch{1.0}
  \begin{tabular}{l|lc|ccccc|ccccc}
      \topline
      \rowcolor{poster_color1}
      & \multicolumn{2}{c|}{\textbf{Data}} & \multicolumn{5}{c|}{\textbf{Domain}} &&&&&\\
      \rowcolor{poster_color1}
      \multirow{-2}{*}{\textbf{Datasets}}
      & \textbf{Camera} & \textbf{Sur.}
      & \textbf{Plat.} & \textbf{Move.} & \textbf{Env.} & \textbf{Weath.} & \textbf{Lighting}
      & \multirow{-2}{*}{\textbf{Modality}} & \multirow{-2}{*}{\textbf{Volume Size}}
      & \multirow{-2}{*}{\textbf{Scenes}} & \multirow{-2}{*}{\textbf{Frames}} & \multirow{-2}{*}{\textbf{Anno.}} \\
      \hline\hline

      \textbf{NYUv2}~\cite{nyu}
      & Pinhole & \crossmark & \webm & \stationary & Indoor & \crossmark & \crossmark
      & C\&D & $(144{\times}240{\times}240)$ & 1.4K & 1.4K & \human \\

      \textbf{ScanNet}~\cite{scannet}
      & Pinhole & \crossmark & \webm & \stationary & Indoor & \crossmark & \crossmark
      & C\&D & $(31{\times}62{\times}62)$ & 1.5K & 1.5K & \human \\

      \textbf{SUNCG}~\cite{song2017semantic}
      & Pinhole & \crossmark & \webm & \wheels & Indoor & \crossmark & \crossmark
      & C\&D & $(144{\times}240{\times}240)$ & 46K & 140K & \syn \\

      \textbf{SemanticPOSS}~\cite{semanticposs}
      & \textit{n.a.} & \crossmark & \car & \wheels & Outdoor & \crossmark & \sun
      & L & $(32{\times}256{\times}256)$ & 6 & 3K & \human \\

      \textbf{SemanticKITTI}~\cite{semantickitti}
      & Pinhole & \crossmark & \car & \wheels & Outdoor & \crossmark & \sun
      & C\&L & $(32{\times}256{\times}256)$ & 22 & 9K & \human \\

      \textbf{nuScenes}~\cite{wang2023openoccupancy}
      & Pinhole+Fisheye & \mycheckmark & \car & \wheels & Outdoor & \mycheckmark & \sun \dusk \night
      & C\&L & $(40{\times}512{\times}512)$ & 850 & 200K & \human \\

      \textbf{Occ3D-Waymo}~\cite{tian2024occ3d}
      & Pinhole+Fisheye & \mycheckmark & \car & \wheels & Outdoor & \mycheckmark & \sun \dusk \night
      & C\&L & $(32{\times}200{\times}200)$ & 1K & 200K & \human \\

      \textbf{SSCBench-KITTI-360}~\cite{li2024sscbench}
      & Fisheye & \mycheckmark & \car & \wheels & Outdoor & \crossmark & \sun
      & C\&L & $(32{\times}200{\times}200)$ & 11 & 91K & \human \\

      \textbf{ScribbleSC}~\cite{wang2024label_efficient}
      & Pinhole & \crossmark & \car & \wheels & Outdoor & \crossmark & \sun
      & C\&L & $(32{\times}256{\times}256)$ & 11 & 9K & \human \\

      \textbf{OmniHD-Scenes}~\cite{zheng2024omnihd}
      & Pinhole & \mycheckmark & \car & \wheels & Outdoor & \crossmark & \sun \night
      & C\&L\&R & -- & 200 & 60K & \robot \human \\

      \textbf{WildOcc}~\cite{zhai2024wildocc}
      & Pinhole & \crossmark & \car & \wheels & Outdoor & \crossmark & \sun
      & C\&L & $(40{\times}100{\times}100)$ & 5 & 10K & \human \\

      \textbf{DSEC-SSC}~\cite{guo2025event}
      & Pinhole & \crossmark & \car & \wheels & Outdoor & \crossmark & \sun \night
      & C\&L\&E & $(16{\times}128{\times}128)$ & 6 & 3K & \human \\

      \textbf{ORAD-3D}~\cite{min2025advancing_orad} & Pinhole & \mycheckmark & \car & \wheels & Outdoor & \mycheckmark & \sun \dusk \night & C\&L & -- & 145 & 58K & \human\\

      \textbf{Co3SOP}~\cite{wu2025synthetic_v2x} & Pinhole & \crossmark & \car & \wheels & Outdoor & \crossmark & \crossmark & C\&L & $(70{\times}1000{\times}1000)$ & -- & -- & \syn \\

      \hline
      \rowcolor{poster_color2}
      \textbf{Human360Occ (ours)}
      & Panoramic & \mycheckmark & \human & \gait & Outdoor & \mycheckmark & \sun \dusk \night
      & C\&D & $(16{\times}128{\times}128)$ & 160 & 8K & \syn \\

      \rowcolor{poster_color2}
      \textbf{QuadOcc (ours)}
      & Panoramic & \mycheckmark & \robotdog & \gait & Outdoor & \crossmark & \sun \dusk \night
      & C\&L & $(8{\times}64{\times}64)$ & 10 & 24K & \robot \human \\

      \bottomline
  \end{tabular}
  }
  \vskip -2ex
  \captionsetup{font=small}
  \caption{Typical datasets for Semantic Scene Completion (SSC). 
  Abbreviations:
  \car~(Autonomous Car), \robotdog~(Quadruped Robot), \robot~(Vision Foundation Model), \human~(Human), \syn~(Synthetic), \mywebm~(Internet images/videos), \wheels~(Wheels), \gait~(Gait), \stationary~(Stationary), \sun~(Day), \dusk~(Dusk), \night~(Night), 
  Sur. (Surround), Plat. (Platform), Move. (Movement), Env. (Environment), Weath. (Weather diversity), Anno. (Annotation), 
  C (Camera), D (Depth), L (LiDAR), R (Radar).}
  \vskip -4ex
  \label{tab:comparison dataset}
\end{table*}

\noindent\textbf{Why legged and humanoid platforms are different.}
Directly transferring SSC is non-trivial: agile gaits cause \emph{body jitter} that corrupts evidence and breaks temporal coherence; these platforms require full \(360^{\circ}\) situational awareness in cluttered, uneven terrain; and tight payload/power budgets favor lightweight, single-sensor, low-latency solutions. 
Hence, we need a panoramic SSC pipeline explicitly robust to gait-induced motion.

\noindent\textbf{Panoramic SSC with a single sensor.}
A single panoramic camera provides compact, true \(360^{\circ}\) coverage but introduces annular distortions and seam artifacts after unwrapping~\cite{lin2025one_flight}—issues poorly handled by perspective-camera SSC pipelines. 
This motivates a design that preserves raw panoramic cues while exploiting grid-friendly projections.

\noindent\textbf{Method: OneOcc.}
We propose OneOcc, a vision-only panoramic SSC framework for severe body jitter and \(360^{\circ}\) continuity, integrating:
(i) \emph{Dual-Projection fusion (DP-ER)} to jointly process the raw annular panorama and its equirectangular unfolding;
(ii) \emph{Bi-Grid Voxelization (BGV)} that reasons in Cartesian and polar/cylindrical voxel spaces to reduce discretization bias and better match panoramic geometry;
(iii) a lightweight decoder with \emph{Hierarchical AMoE-3D} for dynamic multi-scale 3D fusion; and
(iv) a plug-and-play \emph{Gait Displacement Compensation (GDC)} that learns feature-level jitter correction without extra sensors.

\begin{figure*}[!t]
\centering
\includegraphics[width=1.0\linewidth]{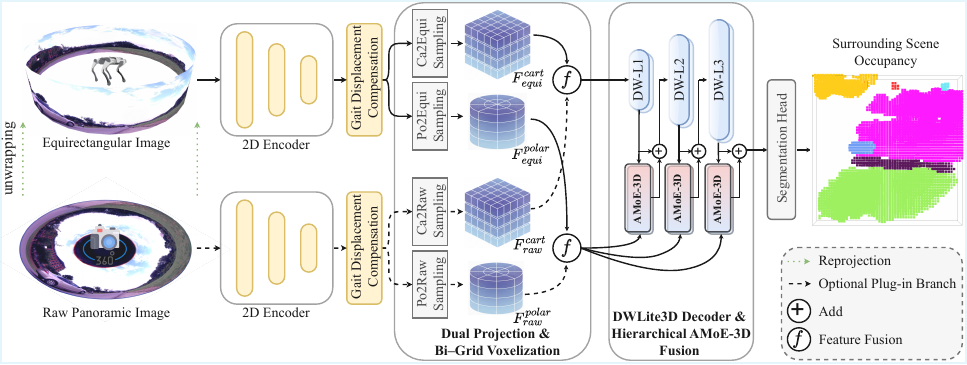}
\vskip -2ex
\caption{\textbf{OneOcc pipeline.}
Calibrated \emph{unwrapping} maps the raw panorama to equirectangular.
Two 2D encoders (\emph{DP-ER}) feed \emph{GDC} for feature-level gait compensation (optional on H3O).
At three scales, \emph{BGV} builds Cartesian and polar/cylindrical voxel grids; voxel-centroid features are sampled to each view (Ca/Po \(\rightarrow\) Equi/Raw) and fused.
A lightweight 3D decoder with \emph{Hierarchical AMoE-3D} aggregates multi-scale context, followed by a head predicting full-surround occupancy. 
\emph{Abbrev.:} Ca/Po = Cartesian/Polar; Equi = equirectangular.}
\vskip -2ex
\label{fig:oneocc_pipeline}
\end{figure*}

\noindent\textbf{Benchmarks: QuadOcc and Human360Occ (H3O).}
To advance panoramic SSC on legged/humanoid robots, we release two complementary benchmarks.
\emph{QuadOcc} is a real first-person \(360^{\circ}\) dataset on a quadruped within a campus domain, with standardized day/dusk/night coverage and semi-automatic ground truth (multi-frame LiDAR aggregation, Grounded-SAM~\cite{ren2024grounded} initialization, targeted manual fixes). 
It contains \emph{10 scenes} and \emph{24K frames}, uses \emph{6 semantic categories} for training.
The occupancy uses \emph{0.4\,m} voxels on a \emph{\(64\!\times\!64\!\times\!8\)} grid—reflecting embodied compute and lower speed than cars.
\emph{Human360Occ (H3O)} is CARLA-based human-ego \(360^{\circ}\) with simulated gait, \emph{160 sequences / 8K frames} over \emph{16 maps} and multiple weathers/times; each frame provides \emph{RGB}, \emph{metric depth}, \emph{semantic occupancy} at \emph{64\(\times\)64\(\times\)8} and \emph{128\(\times\)128\(\times\)16}, and \emph{pose}. 
It supports \emph{within-city} (per-map \(8{:}2\) train/val) and \emph{cross-city} (default \(12{/}4\) maps) splits. See details on the construction pipeline and dataset statistics in the supplementary material.

\noindent\textbf{Results in brief.}
On \emph{QuadOcc}, OneOcc reaches 20.56 mIoU, surpassing the best LiDAR baseline (LMSCNet, \(18.44\); \(+2.12\) mIoU, \(+11.5\%\)) and the best vision baseline (MonoScene, \(19.19\); \(+1.37\) mIoU). 
On \emph{H3O}, it attains 37.29/32.23 mIoU under within-/cross-city, improving over the best vision baselines (\(33.46/24.15\)) by \(+3.83\) and \(+8.08\) mIoU (up to \(+33.5\%\) relative); see Fig.~\ref{fig:teaser}{(b)}.

\noindent\textbf{Contributions.}
\begin{compactitem}
\item \textbf{Method.} 
\emph{OneOcc} integrates \emph{DP-ER}, \emph{BGV}, \emph{Hierarchical AMoE-3D}, and \emph{GDC} to deliver jitter-robust, full-surround SSC with lightweight computation.
\item \textbf{Datasets.} 
Two panoramic occupancy benchmarks for legged robots: \emph{QuadOcc} (real quadruped, \(360^{\circ}\), \(10\) scenes/\(24\)K frames, \(6\) categories, \(0.4\)m voxels on \(64\!\times\!64\!\times\!8\)) and \emph{H3O} (CARLA human-ego \(360^{\circ}\), \(160\) seq/\(8\)K frames on \(16\) maps, per-frame RGB/Depth/voxels/pose, two voxel resolutions).
\item \textbf{Evaluation.} 
Extensive benchmarking shows vision-only \emph{OneOcc} rivals or surpasses  classic LiDAR-based methods under realistic legged/humanoid settings.
\end{compactitem}

\section{Related Work}
\label{sec:related_works}

\noindent\textbf{Semantic scene completion.}
Semantic Scene Completion (SSC) predicts a complete 3D voxel scene with semantics from partial observations.
\textit{LiDAR-based SSC.} From SSCNet~\cite{song2017semantic} to LMSCNet~\cite{roldao2020lmscnet}, LiDAR pipelines evolved from dense 3D CNNs to efficient BEV-like decoders; later works (\textit{e.g.}, JS3C-Net~\cite{yan2021sparse}, SCPNet~\cite{xia2023scpnet}) enhance long-range context and 3D aggregation. Non-Cartesian discretizations mitigate quantization bias (PointOcc~\cite{zuo2023pointocc}, inspired by Cylinder3D~\cite{zhou2020cylinder3d}); recent efforts also improve supervision efficiency and robustness~\cite{wang2024not_voxels_equal,zhao2024lowrankocc}. Yet LiDAR stacks are often heavy and power-hungry for constrained platforms~\cite{ma2025licrocc,wang2024occrwkv,yang2021semantic,zou2021up,mei2023ssc}.
\textit{Vision-based SSC.} The ill-posed 2D$\rightarrow$3D lifting has progressed rapidly: MonoScene~\cite{cao2022monoscene} set a strong baseline; transformer models (VoxFormer~\cite{li2023voxformer}, OccFormer~\cite{zhang2023occformer}) improve global aggregation. Follow-ups refine supervision/projection (RenderOcc~\cite{pan2024renderocc}, InverseMatrixVT3D~\cite{ming2024inversematrixvt3d}, SfMOcc~\cite{marcuzzi2025sfmocc}) and pursue efficiency via sparsity or streamlined designs~\cite{tang2024sparseocc,xue2024bi_ssc,jevtic2025feed_forward}. Other lines study context disentangling/causality~\cite{liu2025disentangling,chen2025semantic_causality_occ} and scene-adaptive decoders or occlusion-aware projections~\cite{lee2025soap,lu2025vishall3d}; for deployment, OccFiner~\cite{shi2025offboard} and TALOS~\cite{jang2024talos} target offboard refinement and test-time adaptation.
\textit{Gaussian/weak-supervision SSC.}
GaussianOcc-style methods~\cite{gan2024gaussianocc,boeder2025gaussianflowocc} reduce dense 3D labeling via Gaussian/flow priors, and GaussianFormer variants~\cite{huang2024gaussianformer,huang2025gaussianformer_2} model scenes as Gaussian fields for efficient camera-based occupancy.
\textit{Panoramic SSC and multi-modality.}
While most camera-based SSC methods assume forward pinhole views, emerging works leverage omnidirectional sensing: panoramic depth for occupancy~\cite{wu2025omnidirectional}, real-time fisheye occupancy without semantics~\cite{pan2024generocc}, cylindrical fusion~\cite{ming2025occcylindrical}, event cameras for HDR/blur~\cite{guo2025event}, and radar-camera fusion for joint tasks~\cite{zheng2025doracamom}. 
Multimodal pipelines~\cite{ma2025licrocc,pan2024co_occ,li2025occmamba,yang2025metaocc} highlight complementary cues but increase system complexity. In contrast, we target a \emph{single panoramic camera} with \emph{full-360\textdegree\ continuity} under \emph{gait-induced jitter}, a setting under-explored by prior literature.

\noindent\textbf{Semantic scene understanding for legged robots.}
Legged autonomy motivates perception tailored to agile motion and strict payload budgets. Prior efforts include vision-based terrain reconstruction~\cite{zhang2022vision}, large-scale datasets~\cite{patel2025tartanground}, and traversability learning~\cite{deng2021vision,aegidius2025watch,oh2024trip}; complementary modules cover robust tracking~\cite{cao2025siamese,xin2024robotic} and instruction grounding~\cite{mei2024quadrupedgpt,han2025space}. With increasing evidence that limited FoV hinders agility~\cite{li2024move,cheng2024quadruped,wang2025omni}, panoramic sensors are adopted for omnidirectional tracking~\cite{luo2025omnidirectional}, multimodal odometry~\cite{li2025limo}, and even satellite communication on the move~\cite{liu2025starlink}. Some works study probabilistic semantic mapping~\cite{chen2025particle}, and humanoid systems explore panoramic/occupancy perception~\cite{cui2025humanoid,zhang2025humanoidpano,zhang2025roboocc,zhang2025occupancy_world_models_robots}. However, \emph{dense 3D semantic occupancy} from a \emph{single panoramic camera} on dynamic legged platforms is still largely open. We address this by releasing \emph{QuadOcc} and \emph{Human360Occ} and proposing \emph{OneOcc}, a lightweight panoramic occupancy framework built around dual-projection fusion, bi-grid voxelization, hierarchical 3D MoE fusion, and gait displacement compensation.

\begin{figure}[!t]
\centering
\includegraphics[width=1.0\linewidth]{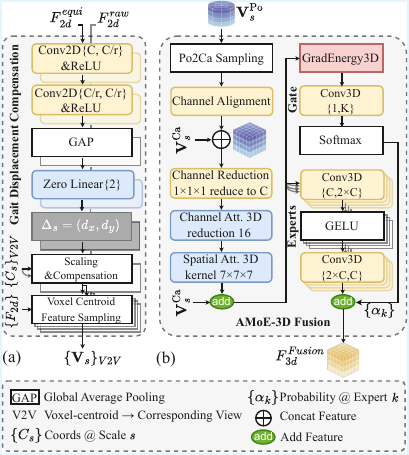}
\vskip -2ex
\caption{\textbf{GDC and AMoE-3D Fusion Module.}
(a) GDC regresses \(\Delta=(dx,dy)\) from 2D features via GAP+Linear (zero–init), applies scale/shift to multi–scale view coordinates \(\{C_s\}\) for voxel–centroid sampling, and yields \(\{F_{3d}\}_{V2V}\).
(b) AMoE-3D fuses Po2Ca-sampled polar features with Cartesian ones via channel alignment, concatenation, and reduction, applies 3D channel- and spatial attention, and performs MoE–Fuse3D with \(K\) experts and GradEnergy3D–Softmax gating:
\(y=\sum_k P_k\odot E_k(x)\).
The fused \(F^{\mathrm{Fusion}}_{3d}\) feeds the 3D decoder.
}
\vskip -2ex
\label{fig:gdc_amoe3d}
\end{figure}

\section{Method}
\label{sec:method}

\subsection{Problem Formulation}
Given a panoramic RGB image $\mathbf{I}\!\in\!\mathbb{R}^{H\times W\times 3}$ and calibrated intrinsics/extrinsics, we aim to predict a dense 3D semantic occupancy $\mathbf{S}\!\in\!\{0,\ldots,C\}^{X\times Y\times Z}$. We produce per-voxel logits $\mathbf{Z}\!\in\!\mathbb{R}^{X\times Y\times Z\times C}$, and the final label is $\hat{s}_{x,y,z}=\arg\max_{c}\mathbf{Z}_{x,y,z,c}$.

\subsection{Overview}
\textbf{Motivation.}
(1) A \(360^\circ\) panorama exhibits \emph{ring continuity} and strong projection distortion: a single projection cannot balance resolution and receptive fields near the poles vs.\ the equator;
(2) Legged locomotion induces \emph{gait jitters} (impulsive foot strikes, small roll/pitch) that hurt the feature$\to$voxel lifting step more than the image$\to$feature step;
(3) Surrounding SSC must reconcile \emph{near-field contact geometry} and \emph{far-field ring context}.
\textbf{Principles.} We therefore combine \emph{dual-projection encoders (DP-ER)}, \emph{bi-grid voxelization (BGV)}, a lightweight \emph{Gait Displacement Compensation (GDC)} \emph{before} lifting, and a \emph{Hierarchical AMoE-3D} decoder for scale-aware, anisotropic fusion.

\subsection{Unwrapping and Dual-Projection Encoders}
\paragraph{Motivation.}
Equirectangular views preserve \emph{azimuthal continuity} that is convolution-friendly; the raw panoramic annulus retains \emph{native geometry and fine textures}. Running both in parallel compensates the equator–pole trade-off.

\paragraph{Method.}
We calibrate the omnidirectional panoramic annular lens (PAL) camera with the Taylor polynomial model~\cite{scaramuzza2006toolbox}. The unwrapping samples the raw annulus at coordinates given by the calibrated inverse mapping:
\begin{equation}
\mathbf{I}^{\mathrm{equi}}(u,v)\;=\;\mathbf{I}^{\mathrm{raw}}\!\left(\,\mathcal{U}^{-1}(u,v;\boldsymbol{\kappa})\,\right),
\end{equation}
where $\boldsymbol{\kappa}=\{a_i,\,u_0,\,v_0,\,\mathbf{A}\}$ are the Taylor coefficients and principal point plus a $2{\times}2$ affine matrix (scale/skew), and $(u,v)\!\in\![0,W)\!\times\![0,H)$. For the equirectangular ray with spherical angles $(\phi,\theta)$:
\begin{align}
\phi \;&=\;\tfrac{2\pi}{W}\,u-\pi,\qquad
\theta \;=\;\tfrac{\pi}{2}-\tfrac{\pi}{H}\,v,\\
r(\theta) \;&=\;\sum_{i=0}^{N} a_i\,\theta^{\,i},\qquad
\begin{bmatrix}u_{\text{raw}}\\[2pt] v_{\text{raw}}\end{bmatrix}
=\begin{bmatrix}u_0\\[2pt] v_0\end{bmatrix}
+\mathbf{A}\,r(\theta)\begin{bmatrix}\cos\phi\\[2pt]\sin\phi\end{bmatrix}.
\end{align}
Two encoders then produce multi-scale features:
\begin{equation}
\{\mathbf{F}^{\mathrm{equi}}_{1/s}\}_{s\in\{1,4,8,16\}},\qquad
\{\mathbf{F}^{\mathrm{raw}}_{1/s}\}_{s\in\{1,4,8,16\}},
\end{equation}
with GroupNorm and $1{\times}1$ bottleneck alignment to standardize channels.

\paragraph{Why it helps.}
For legged robots, the equatorial ring dominates motion priors while polar areas are highly distorted. Using the Taylor unwrapping respects PAL optics, preserving \emph{ring continuity} in ER space and \emph{local textures} in the raw annulus, providing more stable cues for lifting.

%

\begin{table*}[t]
\centering
\setlength{\tabcolsep}{0.01\linewidth}
\begin{threeparttable}
\caption{\textbf{Semantic scene completion results on the QuadOcc validation set.}
All models are (re)trained by us on QuadOcc.
\textbf{Bold} = best, \underline{underline} = second best (within the LiDAR-based block; Vision-based block marked separately.)}
\label{table:quadocc_comp}

\begingroup
\small     

\begin{tabular}{l|l|c c c c c c|cc|cc}
\toprule
Method & Input
& \rotatebox{90}{\textcolor{colorVehicle}{$\blacksquare$} \makecell[l]{vehicle \\ (\quadoccfreq{vehicle}\%)}}
& \rotatebox{90}{\textcolor{colorPedestrian}{$\blacksquare$} \makecell[l]{pedestrian \\ (\quadoccfreq{pedestrian}\%)}}
& \rotatebox{90}{\textcolor{colorRoad}{$\blacksquare$} \makecell[l]{road \\ (\quadoccfreq{road}\%)}}
& \rotatebox{90}{\textcolor{colorBuilding}{$\blacksquare$} \makecell[l]{building \\ (\quadoccfreq{building}\%)}}
& \rotatebox{90}{\textcolor{colorVegetation}{$\blacksquare$} \makecell[l]{vegetation \\ (\quadoccfreq{vegetation}\%)}}
& \rotatebox{90}{\textcolor{colorTerrain}{$\blacksquare$} \makecell[l]{terrain \\ (\quadoccfreq{terrain}\%)}}
& Precision $\uparrow$ & Recall $\uparrow$ & IoU $\uparrow$ & \textbf{mIoU} $\uparrow$ \\
\midrule\midrule

\rowcolor{gray!20}
\multicolumn{12}{c}{\textit{LiDAR-based}} \\
\midrule
SSCNet~\cite{song2017semantic} & L & 0.00 & 0.04 & 44.34 & \underline{16.06} & 22.41 & 4.77 & 63.42 & \textbf{65.82} & 47.71 & 14.60 \\
SSCNet-full~\cite{song2017semantic} & L & 0.71 & 0.04 & \underline{55.14} & 14.93 & 19.66 & 7.76 & \underline{78.81} & 60.03 & \underline{51.69} & 16.38 \\
LMSCNet~\cite{roldao2020lmscnet} & L & \underline{0.88} & \underline{0.20} & \textbf{57.02} & \textbf{16.45} & \underline{24.39} & \textbf{11.70} & \textbf{79.70} & \underline{64.49} & \textbf{55.40} & \textbf{18.44} \\
OccRWKV~\cite{wang2024occrwkv} & L & \textbf{2.05} & \textbf{0.35} & 54.17 & 10.29 & \textbf{24.92} & \underline{8.38} & 71.98 & 62.80 & 50.47 & \underline{16.69} \\
\midrule

\rowcolor{gray!20}
\multicolumn{12}{c}{\textit{Vision-based}} \\
\midrule
VoxFormer-S~\cite{li2023voxformer} & P+D & 0.31 & 1.67 & 36.83 & 6.48 & 7.09 & 3.81 & 45.96 & 69.95 & 38.38 & 9.36 \\
SGN~\cite{mei2023camera} & P+D & 7.39 & 1.88 & 53.35 & 9.06 & 21.95 & 9.39 & 59.23 & 67.41 & 46.05 & 17.17 \\
\midrule
VoxFormer-S$^{\dagger}$~\cite{li2023voxformer} & P+L & 0.24 & 1.66 & 38.51 & 10.90 & 11.88 & 4.45 & 51.29 & 75.84 & 44.08 & 11.27 \\
VoxFormer-T$^{\dagger}$~\cite{li2023voxformer} & P+L & 0.30 & 1.65 & 40.21 & 14.48 & 15.86 & 4.64 & 56.94 & \textbf{77.10} & 48.70 & 12.86 \\
SGN$^{\dagger}$~\cite{mei2023camera} & P+L & \underline{11.62} & \underline{2.47} & 53.06 & \underline{15.60} & \underline{25.67} & 9.91 & \textbf{73.70} & 59.62 & \textbf{49.16} & \underline{19.72} \\
\midrule
OccFormer~\cite{zhang2023occformer} & P & 0.29 & 0.37 & 49.46 & 10.36 & 15.00 & 2.64 & 45.79 & \underline{76.99} & 40.28 & 13.02 \\
MonoScene~\cite{cao2022monoscene} & P & 8.15 & 1.59 & \textbf{55.66} & 12.88 & \textbf{26.10} & \underline{10.78} & 62.10 & 69.28 & 48.69 & 19.19 \\
\rowcolor{gray!20}OneOcc (Ours) & P & \textbf{12.16} & \textbf{2.86} & \underline{54.41} & \textbf{16.03} & 24.91 & \textbf{13.01} & \underline{66.69} & 64.74 & \underline{48.92} & \textbf{20.56} \\

\bottomrule
\end{tabular}

\endgroup

\begin{tablenotes}[flushleft]\footnotesize
\item[$\dagger$] Uses a different sensor modality from the original paper (here: \emph{adding LiDAR}). 
\item \textbf{Input legend:} L = LiDAR; P = \emph{Panoramic} camera; D = monocular Depth (pred.). 
\end{tablenotes}
\vskip -2ex
\end{threeparttable}
\end{table*}

\begin{table*}[t]
    \centering
    \setlength{\tabcolsep}{0.006\linewidth}
    \caption{\textbf{Semantic scene completion using Panoramas on H3O (Human360Occ).}
    Two vertically-stacked panels report results on \textbf{HOMO (within-city)} and \textbf{HETER (cross-city)} splits.
    Header frequencies are computed on \emph{non-empty} voxels.
    \textbf{Input legend:} P = Panoramic camera; D$_{pred.}$ = predicted monocular depth; D$_{gt}$ = ground-truth depth.}

    {\small
    \begin{tabular}{l|l|cccccccccc|cc|cc}
        \toprule
        \rowcolor{gray!20}
        \multicolumn{16}{c}{\textbf{HOMO (within-city)}}\\
        \midrule
        Method & Input
        & \ColHeadHomo{hthreeoRoad}{road}{road}
        & \ColHeadHomo{hthreeoSidewalk}{sidewalk}{sidewalk}
        & \ColHeadHomo{hthreeoBuilding}{building}{building}
        & \ColHeadHomo{hthreeoVegetation}{vegetation}{vegetation}
        & \ColHeadHomo{hthreeoCar}{car}{car}
        & \ColHeadHomo{hthreeoTruck}{truck}{truck}
        & \ColHeadHomo{hthreeoBus}{bus}{bus}
        & \ColHeadHomo{hthreeoTwoWheeler}{two\_wheeler}{twoWheeler}
        & \ColHeadHomo{hthreeoPerson}{person}{person}
        & \ColHeadHomo{hthreeoPole}{pole}{pole}
        & Precision & Recall & IoU & \textbf{mIoU} \\
        \midrule
        VoxFormer\text{-}S~\cite{li2023voxformer} & P{+}D$_{pred.}$ & 22.27 & 17.44 & 2.86 & 5.65 & 0.87 & 0.14 & 0.00 & 0.00 & 61.46 & 0.18 & 40.21 & 59.49 & 31.57 & 11.09 \\
        VoxFormer\text{-}S~\cite{li2023voxformer} & P{+}D$_{gt}$ & 35.84 & 25.86 & 12.32 & 9.90 & 3.73 & 0.70 & 0.19 & 0.17 & 64.79 & 0.93 & 59.91 & 71.12 & 48.19 & 15.44 \\
        
        SGN\text{-}T~\cite{mei2023camera} & P{+}D$_{pred.}$ & 54.55 & 44.61 & 15.51 & 35.16 & 23.91 & 7.68 & 4.52 & 15.77 & 65.24 & 19.46 & 61.82 & 64.36 & 46.05 & 28.64 \\
        SGN\text{-}S~\cite{mei2023camera} & P{+}D$_{pred.}$ & 54.81 & 47.44 & 15.64 & 33.22 & 24.23 & 8.53 & 2.62 & 13.51 & 65.88 & 19.72 & 62.28 & 63.72 & 45.98 & 28.56 \\
        OccFormer~\cite{zhang2023occformer} & P & 54.11 & 46.41 & 15.14 & 32.84 & 17.57 & 4.30 & 0.00 & 1.65 & 65.08 & 11.72 & 57.30 & 69.94 & 45.98 & 24.88 \\
        MonoScene~\cite{cao2022monoscene} & P & 62.45 & 57.18 & 19.97 & 38.91 & 29.76 & 13.77 & 5.03 & 16.76 & 70.15 & 20.67 & 68.78 & 71.66 & 54.08 & 33.46 \\
        \rowcolor{gray!20}OneOcc (Ours) & P & 63.82 & 62.92 & 22.86 & 41.03 & 33.74 & 17.79 & 6.47 & 26.58 & 71.13 & 26.55 & 69.14 & 74.18 & 55.73 & \textbf{37.29} \\
        \bottomrule
    \end{tabular}
    }

    \vspace{8pt}

    {\small
    \begin{tabular}{l|l|cccccccccc|cc|cc}
        \toprule
        \rowcolor{gray!20}
        \multicolumn{16}{c}{\textbf{HETER (cross-city)}}\\
        \midrule
        Method & Input
        & \ColHeadHeter{hthreeoRoad}{road}{road}
        & \ColHeadHeter{hthreeoSidewalk}{sidewalk}{sidewalk}
        & \ColHeadHeter{hthreeoBuilding}{building}{building}
        & \ColHeadHeter{hthreeoVegetation}{vegetation}{vegetation}
        & \ColHeadHeter{hthreeoCar}{car}{car}
        & \ColHeadHeter{hthreeoTruck}{truck}{truck}
        & \ColHeadHeter{hthreeoBus}{bus}{bus}
        & \ColHeadHeter{hthreeoTwoWheeler}{two\_wheeler}{twoWheeler}
        & \ColHeadHeter{hthreeoPerson}{person}{person}
        & \ColHeadHeter{hthreeoPole}{pole}{pole}
        & Precision & Recall & IoU & \textbf{mIoU} \\
        \midrule
        VoxFormer\text{-}S~\cite{li2023voxformer} & P{+}D$_{pred.}$ & 19.02 & 15.82 & 2.84 & 5.86 & 0.95 & 0.10 & 0.00 & 0.01 & 61.62 & 0.10 & 36.68 & 52.00 & 27.40 & 10.63 \\
        VoxFormer\text{-}S~\cite{li2023voxformer} & P{+}D$_{gt}$ & 32.55 & 25.95 & 8.98 & 9.93 & 4.68 & 0.44 & 0.00 & 0.00 & 65.45 & 0.26 & 59.08 & 61.21 & 42.99 & 14.82 \\
        SGN\text{-}T~\cite{mei2023camera} & P{+}D$_{pred.}$ & 42.57 & 23.94 & 9.42 & 15.84 & 18.22 & 10.21 & 1.55 & 3.11 & 63.44 & 11.93 & 45.84 & 47.11 & 30.26 & 20.02 \\
        SGN\text{-}S~\cite{mei2023camera} & P{+}D$_{pred.}$ & 43.47 & 24.70 & 9.19 & 20.27 & 17.21 & 8.21 & 1.85 & 5.08 & 61.10 & 14.89 & 47.82 & 50.98 & 32.76 & 20.60 \\
        OccFormer~\cite{zhang2023occformer} & P & 47.00 & 39.01 & 10.14 & 24.85 & 11.84 & 1.52 & 0.05 & 0.77 & 66.60 & 6.93 & 53.12 & 63.57 & 40.73 & 20.87 \\
        MonoScene~\cite{cao2022monoscene} & P & 49.68 & 41.41 & 11.04 & 20.57 & 20.34 & 10.36 & 1.98 & 2.79 & 69.68 & 13.69 & 67.39 & 55.00 & 43.44 & 24.15 \\
        \rowcolor{gray!20}OneOcc (Ours) & P & 58.60 & 48.28 & 16.34 & 25.98 & 30.68 & 19.35 & 12.87 & 14.16 & 72.89 & 23.11 & 67.89 & 64.77 & 49.58 & \textbf{32.23} \\
        \bottomrule
    \end{tabular}
    }
    \vskip -2ex
    
    \label{table:h3o_homo_heter_stacked}
\end{table*}

\subsection{Bi-Grid Voxelization and View2View Sampling}
\paragraph{Motivation.}
Cartesian voxels are accurate for \emph{near-field contact geometry} (footholds, obstacles), whereas polar/cylindrical voxels preserve \emph{azimuthal continuity} that matches panoramic imaging. In a \(360^\circ\) camera, the equirectangular horizontal axis is linearly related to azimuth \(\phi\); cylindrical voxels \((r,\varphi,z)\) therefore form a \emph{natural, camera-aligned} parameterization where \(\varphi\!\approx\!\phi\) and \(r\) correlates with depth, yielding uniform sampling along the ring and reduced far-field aliasing. Combining Cartesian and cylindrical grids reduces quantization bias and balances near/far evidence under legged motion.

\paragraph{Method.}
We define voxel centroids
\begin{equation}
\left\{
\begin{aligned}
\mathbf{c}^{\mathrm{Ca}}_{ijk} \;&=\; (x_i,y_j,z_k),\\
\mathbf{c}^{\mathrm{Po}}_{pqk} \;&=\; \big(r_p\cos\varphi_q,\; r_p\sin\varphi_q,\; z_k\big).
\end{aligned}
\right.
\end{equation}
For view $v\!\in\!\{\mathrm{equi},\mathrm{raw}\}$ and scale $s$, we lift features by bilinear sampling at the projected pixel $\pi_v(\cdot)$:
\begin{equation}
\left\{
\begin{aligned}
\mathbf{V}^{\mathrm{Ca}}_{s}(\mathbf{c}) \;&=\; \mathrm{bilinear}\!\left(\mathbf{F}^{v}_{1/s},\,\pi_v(\mathbf{c};\boldsymbol{\kappa})\right),\\
\mathbf{V}^{\mathrm{Po}}_{s}(\mathbf{c}) \;&=\; \mathrm{bilinear}\!\left(\mathbf{F}^{v}_{1/s},\,\pi_v(\mathbf{c};\boldsymbol{\kappa})\right).
\end{aligned}
\right.
\end{equation}
We fuse scales with per-voxel convex weights:
\begin{equation}
\left\{
\begin{aligned}
\mathbf{V}^{\mathrm{Ca}} \;&=\; \sum_{s}\alpha^{\mathrm{Ca}}_{s}\odot \mathbf{V}^{\mathrm{Ca}}_{s},\qquad \sum_{s}\alpha^{\mathrm{Ca}}_{s}=1,\\
\mathbf{V}^{\mathrm{Po}} \;&=\; \sum_{s}\alpha^{\mathrm{Po}}_{s}\odot \mathbf{V}^{\mathrm{Po}}_{s},\qquad \sum_{s}\alpha^{\mathrm{Po}}_{s}=1.
\end{aligned}
\right.
\end{equation}
To inject polar context into the Cartesian stream, we pre-compute cross-grid indices $\{\mathcal{J}_\ell\}_{\ell\in\{1,2,4\}}$ and form
\begin{equation}
\widetilde{\mathbf{V}}^{\mathrm{Ca}}_{\ell}
\;=\;
\mathrm{Align}_{1\times 1\times 1}\!\big(\,\mathbf{V}^{\mathrm{Po}}_{\ell}[\mathcal{J}_\ell]\,\big)
\;\Vert\;
\mathbf{V}^{\mathrm{Ca}}_{\ell}.
\end{equation}

\paragraph{Why it helps.}
On legged robots, near-field foothold safety leverages Cartesian precision while far-field loop/context exploits cylindrical efficiency. The camera-aligned polar stream improves \emph{azimuthal coherence} and reduces spherical-distortion artifacts; the fused bi-grid volume balances near/far evidence under gait jitters.

\subsection{Gait Displacement Compensation}
\paragraph{Motivation.}
Gait shocks induce phase errors; correcting them \emph{after} lifting acts on already voxel-quantized evidence. Compensating \emph{before} lifting---at sampling coordinates---avoids quantization and is cheaper.

\paragraph{Method.}
For each scale and projection, we regress a 2D displacement $\Delta_s\!=\!(d_x,d_y)$ with a zero-initialized head:
\begin{equation}
\Delta_s \;=\; \mathrm{Linear}_{0}\!\Big(\mathrm{GAP}(\mathbf{F}^{v}_{1/s})\Big),
\end{equation}
where $\mathrm{Linear}_{0}$ has all weights and bias set to zero so the initial warp is identity (cf.~zero-conv~\cite{zhang2023adding}). Before lifting we correct and resample:
\begin{align}
\widehat{\mathbf{p}} \;&=\; \pi_v(\mathbf{c};\boldsymbol{\kappa}) \;+\; \Delta_s,\\
\mathbf{V}^{(\cdot)}_{s}(\mathbf{c}) \;&=\; \mathrm{bilinear}\!\Big(\mathbf{F}^{v}_{1/s},\,\widehat{\mathbf{p}}\Big).
\end{align}
In practice, this also upgrades the lifting operator from integer-index gather~\cite{cao2022monoscene} to bilinear sampling, which reduces projection aliasing and further stabilizes GDC.
Here, $v\!\in\!\{\text{unwrapped},\text{spherical}\}$ indexes the projection path; $\pi_v(\mathbf{c};\boldsymbol{\kappa})$ projects a 3D voxel centroid $\mathbf{c}$ to image pixels using the calibrated Taylor camera with parameters $\boldsymbol{\kappa}$. $\widehat{\mathbf{p}}$ is the corrected sampling pixel; $\mathrm{bilinear}(\cdot)$ samples the 2D feature map $\mathbf{F}^{v}_{1/s}$, and $\mathbf{V}^{(\cdot)}_{s}$ denotes the lifted 3D feature.

\paragraph{Why it helps.}
GDC routes the \emph{phase error} back to 2D, preventing mis-attributed voxels. The zero-init head stabilizes early training and leaves features unperturbed when motion is tiny ($\Delta_s\!\approx\!0$). We additionally provide qualitative analysis of GDC in the supplementary material.

\subsection{Hierarchical AMoE-3D}
\paragraph{Motivation.}
Panoramic scenes are \emph{anisotropic}: strong azimuthal variation vs.\ vertical structure, and large near/far scale disparity. We thus (i) inject polar cues hierarchically (coarse-to-fine), and (ii) apply \emph{voxel-wise expert selection} to prevent over-smoothing in flat regions while enhancing edges/contacts critical for locomotion.

\noindent\textbf{Method.}
The 3D decoder is a depthwise-separable UNet with trilinear upsampling at three levels (L1/L2/L3). At each level we fuse $\widetilde{\mathbf{V}}^{\mathrm{Ca}}$ using a \emph{dual-path volumetric saliency} and a \emph{Mixture-of-Experts} (MoE) fuse; we term this module \emph{AMoE-3D}, where \emph{A} denotes attention implemented via the dual-path (channel \& spatial) saliency gates.

\textit{Dual-path volumetric saliency (channel \& spatial gates):}
\begin{equation}
\left\{
\begin{aligned}
\mathbf{A}_c \;&=\; \sigma\!\Big(\mathrm{MLP}(\mathrm{GAP}(\mathbf{X}))+\mathrm{MLP}(\mathrm{GMP}(\mathbf{X}))\Big),\\
\mathbf{A}_s \;&=\; \sigma\!\Big(g^{7\times 7\times 7}\!\left([\mathrm{Avg}(\mathbf{X});\,\mathrm{Max}(\mathbf{X})]\right)\Big),\\
\mathbf{Y}   \;&=\; \mathbf{X}\odot \mathbf{A}_c \odot \mathbf{A}_s.
\end{aligned}
\right.
\end{equation}

\textit{MoE-Fuse3D with gradient-energy gating:}
\begin{equation}
\left\{
\begin{aligned}
&\mathrm{GradEnergy3D}(\mathbf{Y}) \;=\; \sum_{a\in\{x,y,z\}}\big\|\nabla_a \mathbf{Y}\big\|_2^2,\\
&\boldsymbol{\alpha} \;=\; \operatorname{softmax}\!\big(W_g * \mathrm{GradEnergy3D}(\mathbf{Y})\big),\\
&\widetilde{\mathbf{Y}} \;=\; \mathbf{Y} + \sum_{k=1}^{K}\alpha_k\,E_k(\mathbf{Y}),
\end{aligned}
\right.
\end{equation}
where $E_k$ are $1{\times}1{\times}1$ Conv–GELU–Conv experts. 
Here, $\mathbf{X}\!\in\!\mathbb{R}^{C\times D\times H\times W}$ 
is the level-$\ell$ Cartesian feature and $\widetilde{\mathbf{V}}^{\mathrm{Ca}}$ is the polar volume resampled onto the same Cartesian grid; $\mathbf{A}_c\!\in\!\mathbb{R}^{C\times1\times1\times1}$ and $\mathbf{A}_s\!\in\!\mathbb{R}^{1\times D\times H\times W}$ are the channel and spatial gates. $\nabla_a$ denotes the discrete 3D gradient along axis $a\!\in\!\{x,y,z\}$, $*$ is 3D convolution, and $[\,;\,]$ concatenates along channels.

\paragraph{Why it helps.}
Hierarchical polar injections aggregate \emph{far-field azimuthal cues} at coarse levels—\textit{e.g.}, ring-consistent \emph{road/building/vegetation} context —and refine \emph{near-field contact geometry} at fine levels. Gradient-energy gating emphasizes high-contrast structures at category transitions (vehicles, poles) while suppressing overfitting on large ground flats (road), improving stability of foothold decisions.

\begin{table*}[t]
    \centering
    \caption{\textbf{Ablations on QuadOcc (Panorama only).}
    Core columns show if a component is enabled (\cmark) or disabled (\xmark).}
    \vskip -2.25ex
    \setlength{\tabcolsep}{0.02\linewidth} 
    \renewcommand{\arraystretch}{1.12}
    \resizebox{0.9\textwidth}{!}{
    \begin{tabular}{l|cccc|cccccc|cccc}
        \toprule
        \multirow{2}{*}{Variant} &
        \multicolumn{4}{c|}{\textbf{Core Designs}} &
        \multicolumn{6}{c|}{\textbf{Per-class IoU (QuadOcc)}} &
        \multicolumn{4}{c}{\textbf{Overall}} \\
        \cmidrule(lr){2-5}\cmidrule(lr){6-11}\cmidrule(lr){12-15}
        & \rotatebox{90}{\makecell[l]{\textbf{GDC}\\\scriptsize Gait Disp.\\\scriptsize Compensation}}
        & \rotatebox{90}{\makecell[l]{\textbf{DP\textendash ER}\\\scriptsize Dual Projection\\\scriptsize (Equi. + Raw)}}
        & \rotatebox{90}{\makecell[l]{\textbf{BGV}\\\scriptsize Bi\textendash Grid\\\scriptsize Voxelization}}
        & \rotatebox{90}{\makecell[l]{\textbf{AMoE\textendash 3D}\\\scriptsize Attention\textendash MoE\\\scriptsize 3D Fusion}}
        & \rotatebox{90}{\textcolor{colorVehicle}{$\blacksquare$} \makecell[l]{vehicle \\ (\quadoccfreq{vehicle}\%)}}     
        & \rotatebox{90}{\textcolor{colorPedestrian}{$\blacksquare$} \makecell[l]{pedestrian \\ (\quadoccfreq{pedestrian}\%)}}
        & \rotatebox{90}{\textcolor{colorRoad}{$\blacksquare$} \makecell[l]{road \\ (\quadoccfreq{road}\%)}} 
        & \rotatebox{90}{\textcolor{colorBuilding}{$\blacksquare$} \makecell[l]{building \\ (\quadoccfreq{building}\%)}} 
        & \rotatebox{90}{\textcolor{colorVegetation}{$\blacksquare$} \makecell[l]{vegetation \\ (\quadoccfreq{vegetation}\%)}} 
        & \rotatebox{90}{\textcolor{colorTerrain}{$\blacksquare$} \makecell[l]{terrain \\ (\quadoccfreq{terrain}\%)}}
        & \rotatebox{90}{\makecell[l]{SC\\Precision}}
        & \rotatebox{90}{\makecell[l]{SC\\Recall}}
        & \rotatebox{90}{\makecell[l]{SC\\IoU}}
        & \rotatebox{90}{\makecell[l]{\textbf{SSC}\\\textbf{mIoU}}} \\
        \midrule
        (Q0) baseline & \xmark & \xmark & \xmark & \xmark
        & 8.15 & 1.59 & \textbf{55.66} & 12.88 & \textbf{26.10} & 10.78 
        & 62.10 & \textbf{69.28} & 48.69 & 19.19 \\
        (Q1) + GDC                            & \cmark & \xmark & \xmark & \xmark
        & 11.76 & 2.48 & 53.73 & 15.07 & 22.95 & 11.51 
        & 64.05 & \underline{66.08} & \underline{48.78} & 19.58 \\
        (Q2) + DP\textendash ER               & \cmark & \cmark & \xmark & \xmark
        & 11.82 & 2.51 & 54.16 & 15.39 & 23.40 & 12.08 
        & 65.42 & 65.01 & 48.38 & 19.89 \\
        (Q3) + BGV                            & \cmark & \cmark & \cmark & \xmark
        & \underline{12.07} & \underline{2.77} & 54.29 & \underline{15.76} & 24.19 & \underline{12.74} 
        & \underline{66.56} & 64.70 & 48.68 & \underline{20.30} \\
        \rowcolor{gray!10}
        (Q4) + AMoE\textendash 3D              & \cmark & \cmark & \cmark & \cmark
        & \textbf{12.16} & \textbf{2.86} & \underline{54.41} & \textbf{16.03} & \underline{24.91} & \textbf{13.01} 
        & \textbf{66.69} & 64.74 & \textbf{48.92} & \textbf{20.56} \\
        \bottomrule
    \end{tabular}
    }
    \vskip -3.75ex
    \label{tab:quadssc_ablation_core}
\end{table*}

\subsection{Segmentation Head and Loss}
\paragraph{Method.}
A $1{\times}1{\times}1$ head outputs per-voxel logits $\mathbf{Z}$. We supervise only \emph{valid} voxels (from lifting visibility) with deep supervision at strides $\{1,2,4\}$.

\paragraph{Loss.}
We follow MonoScene~\cite{cao2022monoscene} for most losses: (i) standard cross-entropy on valid voxels, (ii) the scene-class affinity (SCAL) terms for semantics and geometry, and (iii) the frustums proportion (FP) loss. We do \emph{not} use the relation loss, as we observed it encourages co-occurrence priors that over-smooth azimuthal boundaries and suppress small near-field classes under panoramic legged setting, yielding no gain and occasional degradation on our data. The final objective is:
\begin{equation}
\mathcal{L}_{\mathrm{total}}
\;=\;
\mathcal{L}_{\mathrm{CE}}
\;+\;
\mathcal{L}_{\mathrm{SCAL}}^{\mathrm{sem}}
\;+\;
\mathcal{L}_{\mathrm{SCAL}}^{\mathrm{geo}}
\;+\;
\mathcal{L}_{\mathrm{FP}}.
\end{equation}
Class re-weighting for $\mathcal{L}_{\mathrm{CE}}$ follows standard SSC practice.

\section{Experiments}
\label{sec:exp}

%
\begin{table}[!t]
    \centering
    \caption{\textbf{Performance comparison on the QuadOcc validation set across different lighting conditions.}
    Results are shown for day, dusk, and night scenarios. P  (Precision), R (Recall), IoU and mIoU are reported.
    Methods marked with ${\dagger}$ use a sensor modality different from the original paper.
    The best performance for each input modality is highlighted in \textbf{bold}.}
    \vskip -2ex
    \label{tab:quadocc_lighting_comparison}
    \setlength{\tabcolsep}{2pt}
    \resizebox{0.5\textwidth}{!}{%
    \begin{tabular}{l | c c c c | c c c c | c c c c}
        \toprule
        \multirow{2}{*}{\textbf{Method}} & \multicolumn{4}{c|}{\textbf{Day}~\sun} & \multicolumn{4}{c|}{\textbf{Dusk}~\dusk} & \multicolumn{4}{c}{\textbf{Night}~\night} \\
        \cmidrule(lr){2-5} \cmidrule(lr){6-9} \cmidrule(lr){10-13}
        & \small P & \small R & \small IoU & \small \textbf{mIoU} & \small P & \small R & \small IoU & \small \textbf{mIoU} & \small P & \small R & \small IoU & \small \textbf{mIoU} \\
        \midrule\midrule
        
        \multicolumn{13}{c}{\textit{LiDAR-based}} \\ 
        \midrule 
        SSCNet~\cite{song2017semantic} & 67.12 & 62.95 & 48.12 & 14.03 & 52.97 & \textbf{76.34} & 45.50 & 14.88 & 57.81 & \textbf{75.30} & 48.60 & 9.51 \\ 
        SSCNet-full~\cite{song2017semantic} & 80.56 & 58.52 & 51.28 & 16.04 & 73.79 & 64.34 & 52.37 & 17.14 & 73.64 & 67.31 & 54.24 & 11.56 \\ 
        LMSCNet~\cite{roldao2020lmscnet} & \textbf{81.18} & 62.50 & \textbf{54.59} & \textbf{17.33} & \textbf{75.34} & 70.92 & \textbf{57.56} & \textbf{18.91} & \textbf{75.68} & 72.67 & \textbf{58.91} & \textbf{13.40} \\
        OccRWKV~\cite{wang2024occrwkv} & 72.84 & \textbf{63.83} & 51.56 & 16.25 & 66.43 & 59.30 & 45.63 & 15.42 & 74.10 & 58.31 & 48.44 & 12.61 \\
        
        \midrule
        \multicolumn{13}{c}{\textit{Vision-based}} \\ 
        \midrule 
        OccFormer~\cite{zhang2023occformer} & 46.23 & \textbf{77.08} & 40.64 & 12.82 & 41.79 & 79.45 & 37.72 & 14.46 & 49.49 & 72.18 & 41.56 & 10.21 \\
        VoxFormer-S${\dagger}$~\cite{li2023voxformer} & 55.80 & 73.96 & 46.64 & 11.44 & 37.42 & 82.81 & 34.73 & 8.95 & 47.10 & 82.41 & 42.80 & 11.84 \\
        VoxFormer-T${\dagger}$~\cite{li2023voxformer} & 61.77 & 75.02 & \textbf{51.24} & 12.91 & 42.25 & \textbf{84.68} & 39.25 & 11.06 & 51.93 & \textbf{84.62} & \textbf{47.45} & 12.88 \\
        SGN~\cite{mei2023camera} & 61.54 & 68.71 & 48.06 & 17.94 & 54.07 & 69.06 & 43.53 & 15.19 & 46.83 & 52.29 & 32.81 & 10.58 \\
        MonoScene~\cite{cao2022monoscene} & 65.47 & 69.94 & 51.09 & 18.58 & 47.62 & 77.10 & 41.72 & 15.14 & 51.35 & 66.47 & 40.78 & \textbf{14.20} \\ 
        \rowcolor{gray!20}OneOcc (Ours) & \textbf{69.10} & 64.66 & 50.15 & \textbf{21.15} & \textbf{57.46} & 71.98 & \textbf{46.96} & \textbf{19.86} & \textbf{63.56} & 53.75 & 41.09 & 13.50 \\ 
                        
        \bottomrule
    \end{tabular}%
    }
    \vskip -3ex
\end{table}


\subsection{Setup}
\noindent \textbf{Datasets.}
We evaluate on two panoramic occupancy benchmarks for legged/humanoid platforms.
\textbf{QuadOcc}: real first-person $360^{\circ}$ data on a quadruped within a campus; day/dusk/night, sequence-heterogeneous split; $24$K \emph{consecutive} frames with \emph{stride-5} training; $6$ classes on a $64{\times}64{\times}8$ grid (0.4\,m).
\textbf{Human360Occ (H3O)}: CARLA-based human-ego $360^{\circ}$ with simulated gait over $16$ maps and diverse weather/lighting; RGB, depth, occupancy (two resolutions), and poses. 
We report \emph{within-city} (H3O-Homo) and \emph{cross-city} (H3O-Heter). 
See supplementary for the data collection pipelines and statistics of QuadOcc and H3O.

\noindent \textbf{Metrics.}
Following semantic scene completion (SSC) practice, we report per-class IoU and mean IoU (mIoU) on non-empty voxels, plus overall precision (P) and recall (R).

\noindent \textbf{Baselines.}
We re-train popular \underline{LiDAR} SSC systems (SSCNet~\cite{song2017semantic}, LMSCNet~\cite{roldao2020lmscnet}, OccRWKV~\cite{wang2024occrwkv}) and adapt representative \underline{vision} SSC methods (OccFormer~\cite{zhang2023occformer}, VoxFormer~\cite{li2023voxformer}, SGN~\cite{mei2023camera}, MonoScene~\cite{cao2022monoscene}) to panoramas via calibrated unwrapping to equirectangular. For \emph{P+L} entries in Table~\ref{table:quadocc_comp}, we add LiDAR to the corresponding vision model per the table footnote. Details in the supplementary.

\noindent \textbf{Implementation.}
Unless noted, training follows official configs with minimal dataset-specific changes; schedules/image sizes/voxelization match for fairness. On H3O we disable the raw-projection branch (native equirectangular), keeping components as in ablations. Training uses a single RTX-4090 GPU. Details in the supplementary.

\begin{figure*}[!t]
\centering
\includegraphics[width=0.95\linewidth]{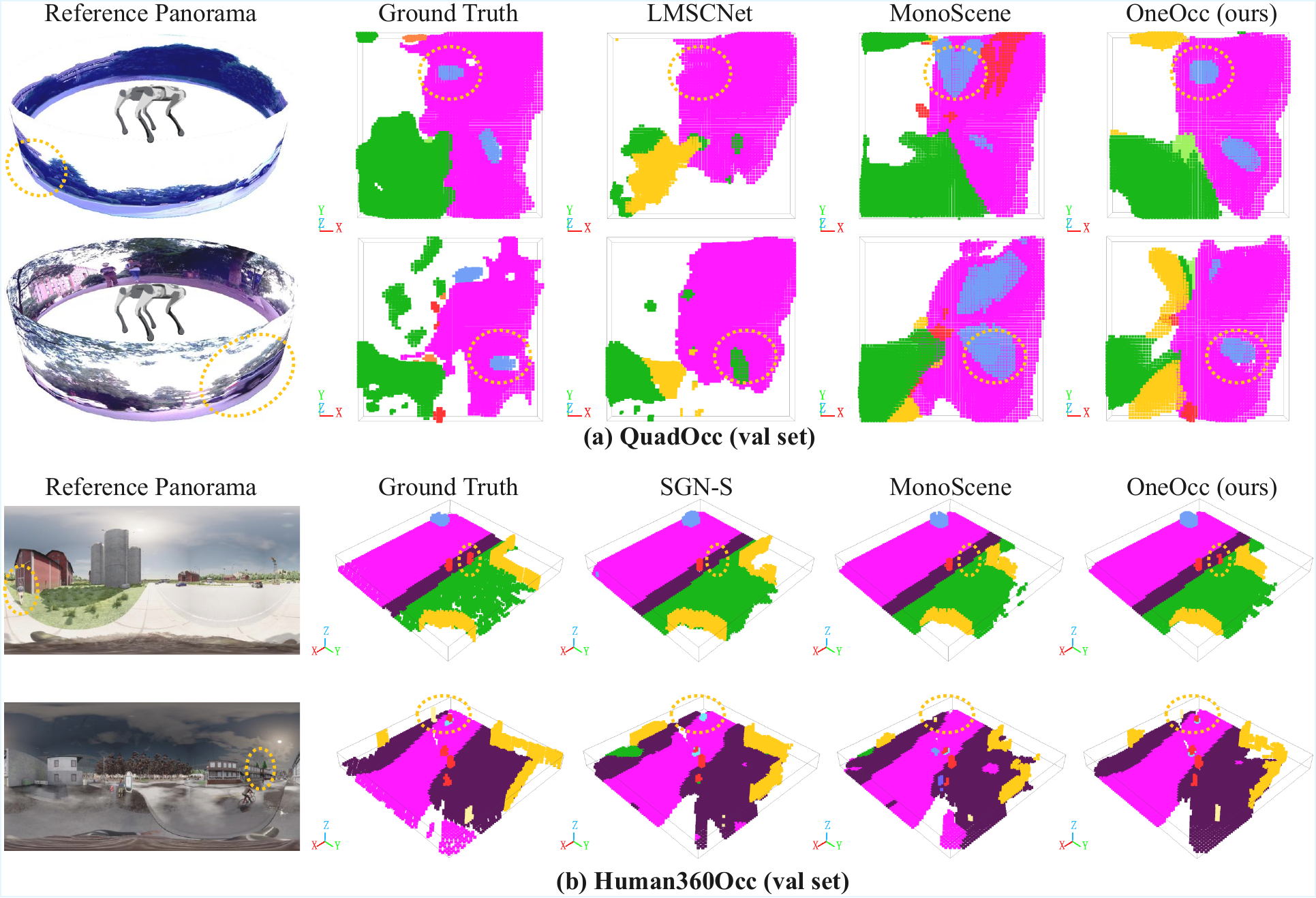}
\vskip -2ex
\caption{\textbf{Qualitative comparisons on (a) \emph{QuadOcc} and (b) \emph{Human360Occ} (val sets).}
Left to right: Reference panorama, ground-truth occupancy, baselines, and OneOcc (ours).
Across both real (QuadOcc) and simulated (H3O) panoramic settings, OneOcc better preserves the global layout and $360^{\circ}$ continuity, produces cleaner class boundaries, and recovers occluded structures.
Yellow dotted circles highlight typical improvements over LMSCNet/SGN-S and MonoScene (\textit{e.g.}, foreground-background disambiguation and far-range layout).}
\vskip -3ex
\label{fig:qualitative_comp}
\end{figure*}

\subsection{Quantitatively Comparisons}
\textbf{QuadOcc.}
Table~\ref{table:quadocc_comp} reports \emph{QuadOcc (val)}. OneOcc attains 20.56 mIoU, surpassing the best LiDAR baseline LMSCNet (\underline{18.44}) and the strongest vision baseline MonoScene (19.19). 
SGN$^{\dagger}$ (P+L) achieves the highest geometry IoU (49.16) among vision entries by leveraging LiDAR, yet camera-only OneOcc delivers the best \emph{mIoU} (20.56) among vision methods. Task-aligned panoramic fusion (DP-ER, BGV, AMoE-3D, with GDC on legged data) enables a camera-only pipeline to rival and even surpass popular LiDAR stacks at this range/resolution. 

\noindent \textbf{Human360Occ (H3O).} Table~\ref{table:h3o_homo_heter_stacked} summarizes generalization: On H3O-Homo OneOcc reaches 37.29 mIoU (+3.83 vs. MonoScene), while on H3O-Heter it attains 32.23 mIoU (+8.08). The larger margin under \emph{heterogeneous} cross-city shifts highlights OneOcc's strong distribution robustness: DP-ER and AMoE-3D transfer across maps and lighting/weather, and their distortion-aware priors mitigate panoramic aliasing, yielding state-of-the-art camera-only mIoU under distribution shift.

\noindent \textbf{Robustness to Lighting Conditions.}
Table~\ref{tab:quadocc_lighting_comparison} reports day/dusk/night.
OneOcc leads the best vision baseline in day (21.15 vs.\ 18.58) and dusk (19.86 vs.\ 15.14); at night its mIoU (13.50) trails MonoScene (14.20) but shows \emph{higher precision}, likely from frustum-artifact suppression (cf. Fig.~\ref{fig:qualitative_comp}). 
\emph{LiDAR trend.} Dusk often exceeds day due to weaker solar background in near-IR, while night drops with altered return statistics and sparser echoes on low-albedo/specular surfaces, hurting long-range completion.

\subsection{Ablations}
We ablate GDC, DP-ER, BGV, AMoE-3D (Table~\ref{tab:quadssc_ablation_core}).
Starting from Q0 (ER-only, no GDC, single-grid, no AMoE-3D; 19.19 mIoU), we add:
+GDC \(\rightarrow\) 19.58 (+0.39),
+DP-ER \(\rightarrow\) 19.89 (+0.31),
+BGV \(\rightarrow\) 20.30 (+0.41),
+AMoE-3D (full) \(\rightarrow\) 20.56 (+0.26). Gains are additive: GDC stabilizes sampling; DP-ER provides complementary cues; BGV reduces discretization bias; AMoE-3D sharpens edges/contacts without over-smoothing flats. We further provide more detailed per-module ablations, calibration-robustness analysis, temporal aggregation baselines, and resolution scaling results in the supplementary material.

\subsection{Qualitative Comparisons}
Fig.~\ref{fig:qualitative_comp} shows typical cases.
On QuadOcc, MonoScene exhibits frustum-direction artifacts (ghost elongation), while OneOcc suppresses them via \emph{DP-ER} (cross-view corroboration), \emph{GDC} (phase-stable sampling), and \emph{BGV} (near/far balance).
On Human360Occ, OneOcc recovers pedestrians close in color to background that competing vision methods miss.
Overall, OneOcc preserves global layout and $360^{\circ}$ continuity and better handles occlusions.

\subsection{Efficiency}
Inference on an RTX\,4090 at $608{\times}1216$, batch size $1$.
OneOcc averages 69.93\,ms (FPS $\approx$ 14.30).
The model has 101.76M parameters and peaks at 1.82\,GB CUDA allocation (2.58\,GB reserved) in FP32.
With mixed precision (FP32+FP16), latency drops to 52.84\,ms (FPS $\approx$ 18.92), and peak allocation is 1.49\,GB.
The footprint remains moderate, making OneOcc suitable for onboard legged perception on a single GPU.
We additionally report measured on-device runtime and a module-wise latency breakdown on NVIDIA Jetson AGX Orin in the supplementary material.

\section{Conclusion}
\label{sec:conclusion}

We introduced \emph{OneOcc}, a panoramic \emph{camera-only} framework for semantic occupancy on legged and humanoid platforms. OneOcc stabilizes feature phases under periodic motion via \emph{Gait Displacement Compensation} (GDC), fuses raw annular and equirectangular streams with \emph{Dual-Projection} (DP--ER), balances near--far geometry through \emph{Bi-Grid Voxelization} (BGV), and aggregates multi-scale evidence using a hierarchical \emph{AMoE-3D} decoder. Together, these components turn monocular panoramas into a geometry-aware 3D grid. We also contribute \emph{QuadOcc} (real $360^{\circ}$ quadruped) and report cross-domain evaluation on \emph{Human360Occ}, facilitating systematic study across robot morphologies and environments. 

\noindent\textbf{Limitations \& Outlook.}
OneOcc assumes accurate calibration with bounded drift. A possible remedy is online extrinsic self-calibration, optionally regularized by lightweight odometry priors. We propose \emph{robotic occupancy} as an intermediate layer for world models and vision-language-action models: tokens for language control, and occupancy-sequence pretraining for cross-robot transfer.

\clearpage

\section*{Acknowledgment}

This research was funded by the Natural Science Foundation of Zhejiang Province (Grant No. LZ24F050003), the National Natural Science Foundation of China (Grant Nos. 12174341 and 62473139), the Hunan Provincial Research and Development Project (Grant No. 2025QK3019), the MOE AcRF Tier 1 SSHR-TG Incubator Grant FY24 (Grant No. RSTG7/24), and the opening project of the State Key Laboratory of Autonomous Intelligent Unmanned Systems (Grant No. ZZKF2025-2-10). Hao Shi was also supported by the China Scholarship Council (CSC) for joint Ph.D. training at Nanyang Technological University (NTU).
{
    \small
    \bibliographystyle{ieeenat_fullname}
    \bibliography{main}

@String(CVPR= {IEEE Conf. Comput. Vis. Pattern Recog.})

@String(ICCV= {Int. Conf. Comput. Vis.})

@String(ECCV= {Eur. Conf. Comput. Vis.})

@String(TOG= {ACM Trans. Graph.})

@String(ICLR = {Int. Conf. Learn. Represent.})

@String(IJCAI = {IJCAI})

@String(AAAI = {AAAI})

@String(CVPR  = {CVPR})

@String(ICCV  = {ICCV})

@String(ECCV  = {ECCV})

@String(TOG   = {ACM TOG})

@String(ICLR  = {ICLR})

@inproceedings{li2023voxformer,
  title={{VoxFormer:} {Sparse} Voxel Transformer for Camera-Based {3D} Semantic Scene Completion},
  author={Yiming Li and
                  Zhiding Yu and
                  Christopher B. Choy and
                  Chaowei Xiao and
                  Jos{\'{e}} M. {\'{A}}lvarez and
                  Sanja Fidler and
                  Chen Feng and
                  Anima Anandkumar},
  booktitle={CVPR},
  year={2023}
}

@article{marcuzzi2025sfmocc,
  title={{SfmOcc:} {Vision-based} {3D} Semantic Occupancy Prediction in Urban Environments},
  author={Rodrigo Marcuzzi and
                  Lucas Nunes and
                  Elias Marks and
                  Louis Wiesmann and
                  Thomas L{\"{a}}be and
                  Jens Behley and
                  Cyrill Stachniss},
  journal={IEEE Robotics and Automation Letters},
  year={2025},
  publisher={IEEE}
}

@article{lai2024rtonet,
  title={{RTONet:} {Real-time} Occupancy Network for Semantic Scene Completion},
  author={Lai, Quan and Zheng, Haifeng and Feng, Xinxin and Zheng, Mingkui and Chen, Huacong and Chen, Wenqiang},
  journal={IEEE Robotics and Automation Letters},
  year={2024},
  publisher={IEEE}
}

@article{ma2025licrocc,
  title={{LiCROcc:} {Teach} Radar for Accurate Semantic Occupancy Prediction Using {LiDAR} and Camera},
  author={Yukai Ma and
                  Jianbiao Mei and
                  Xuemeng Yang and
                  Licheng Wen and
                  Weihua Xu and
                  Jiangning Zhang and
                  Xingxing Zuo and
                  Botian Shi and
                  Yong Liu},
  journal={IEEE Robotics and Automation Letters},
  year={2025},
  publisher={IEEE}
}

@article{wang2024omega,
  title={{OMEGA:} {Efficient} Occlusion-Aware Navigation for Air-Ground Robots in Dynamic Environments via State Space Model},
  author={Junming Wang and
                  Xiuxian Guan and
                  Zekai Sun and
                  Tianxiang Shen and
                  Dong Huang and
                  Fangming Liu and
                  Heming Cui},
  journal={IEEE Robotics and Automation Letters},
  year={2025},
  publisher={IEEE}
}

@article{zhai2024wildocc,
  title={{WildOcc:} {A} Benchmark for Off-Road {3D} Semantic Occupancy Prediction},
  author={Zhai, Heng and Mei, Jilin and Min, Chen and Chen, Liang and Zhao, Fangzhou and Hu, Yu},
  journal={arXiv preprint arXiv:2410.15792},
  year={2024}
}

@article{hoeller2022neural,
  title={Neural scene representation for locomotion on structured terrain},
  author={Hoeller, David and Rudin, Nikita and Choy, Christopher and Anandkumar, Animashree and Hutter, Marco},
  journal={IEEE Robotics and Automation Letters},
  year={2022},
  publisher={IEEE}
}

@article{chen2025particle,
  title={Particle-based Instance-aware Semantic Occupancy Mapping in Dynamic Environments},
  author={Chen, Gang and Wang, Zhaoying and Dong, Wei and Alonso-Mora, Javier},
  journal={IEEE Transactions on Robotics},
  year={2025},
  publisher={IEEE}
}

@article{oh2024trip,
  title={{TRIP:} {Terrain} Traversability Mapping With Risk-Aware Prediction for Enhanced Online Quadrupedal Robot Navigation},
  author={Minho Oh and
                  Byeongho Yu and
                  I Made Aswin Nahrendra and
                  Seoyeon Jang and
                  Hyeonwoo Lee and
                  Dongkyu Lee and
                  Seungjae Lee and
                  Yeeun Kim and
                  Kevin Christiansen Marsim and
                  Hyungtae Lim and
                  Hyun Myung},
  journal={arXiv preprint arXiv:2411.17134},
  year={2024}
}

@article{li2025limo,
  title={{LiMo-Calib:} {On-site} Fast {LiDAR}-Motor Calibration for Quadruped Robot-Based Panoramic {3D} Sensing System},
  author={Jianping Li and
                  Zhongyuan Liu and
                  Xinhang Xu and
                  Jinxin Liu and
                  Shenghai Yuan and
                  Fang Xu and
                  Lihua Xie},
  journal={arXiv preprint arXiv:2502.12655},
  year={2025}
}

@article{zhang2025roboocc,
  title={{RoboOcc:} {Enhancing} the Geometric and Semantic Scene Understanding for Robots},
  author={Zhang Zhang and
                  Qiang Zhang and
                  Wei Cui and
                  Shuai Shi and
                  Yijie Guo and
                  Gang Han and
                  Wen Zhao and
                  Hengle Ren and
                  Renjing Xu and
                  Jian Tang},
  journal={arXiv preprint arXiv:2504.14604},
  year={2025}
}

@article{patel2025tartanground,
  title={{TartanGround:} {A} Large-Scale Dataset for Ground Robot Perception and Navigation},
  author={Manthan Patel and
                  Fan Yang and
                  Yuheng Qiu and
                  Cesar Cadena and
                  Sebastian A. Scherer and
                  Marco Hutter and
                  Wenshan Wang},
  journal={arXiv preprint arXiv:2505.10696},
  year={2025}
}

@article{zuo2023pointocc,
  title={{PointOcc:} {Cylindrical} Tri-Perspective View for Point-based {3D} Semantic Occupancy Prediction},
  author={Zuo, Sicheng and Zheng, Wenzhao and Huang, Yuanhui and Zhou, Jie and Lu, Jiwen},
  journal={arXiv preprint arXiv:2308.16896},
  year={2023}
}

@inproceedings{cao2024slcf_sequential,
  title={{SLCF-Net:} {Sequential} {LiDAR-camera} fusion for semantic scene completion using a {3D} recurrent {U-Net}},
  author={Helin Cao and
                  Sven Behnke},
  booktitle={ICRA},
  year={2024}
}

@inproceedings{yan2024pointssc,
  title={{PointSSC:} {A} Cooperative Vehicle-Infrastructure Point Cloud Benchmark for Semantic Scene Completion},
  author={Yan, Yuxiang and Liu, Boda and Ai, Jianfei and Li, Qinbu and Wan, Ru and Pu, Jian},
  booktitle={ICRA},
  year={2024}
}

@article{yang2025metaocc,
  title={{MetaOcc:} {Spatio-temporal} Fusion of Surround-View {4D} Radar and Camera for {3D} Occupancy Prediction with Dual Training Strategies},
  author={Long Yang and
                  Lianqing Zheng and
                  Wenjin Ai and
                  Minghao Liu and
                  Sen Li and
                  Qunshu Lin and
                  Shengyu Yan and
                  Jie Bai and
                  Zhixiong Ma and
                  Xichan Zhu},
  journal={arXiv preprint arXiv:2501.15384},
  year={2025}
}

@article{ming2025occcylindrical,
  title={{OccCylindrical:} {Multi-modal} Fusion with Cylindrical Representation for {3D} Semantic Occupancy Prediction},
  author={Zhenxing Ming and
                  Julie Stephany Berrio and
                  Mao Shan and
                  Yaoqi Huang and
                  Hongyu Lyu and
                  Nguyen Hoang Khoi Tran and
                  Tzu{-}Yun Tseng and
                  Stewart Worrall},
  journal={arXiv preprint arXiv:2505.03284},
  year={2025}
}

@inproceedings{pan2024generocc,
  title={{GenerOcc:} {Self-supervised} Framework of Real-time {3D} Occupancy Prediction for Monocular Generic Cameras},
  author={Xianghui Pan and
                  Jiayuan Du and
                  Shuai Su and
                  Wenhao Zong and
                  Xiao Wang and
                  Chengju Liu and
                  Qijun Chen},
  booktitle={IROS},
  year={2024}
}

@inproceedings{pan2024renderocc,
  title={{RenderOcc:} {Vision-centric} {3D} Occupancy Prediction with {2D} Rendering Supervision},
  author={Mingjie Pan and
                  Jiaming Liu and
                  Renrui Zhang and
                  Peixiang Huang and
                  Xiaoqi Li and
                  Hongwei Xie and
                  Bing Wang and
                  Li Liu and
                  Shanghang Zhang},
  booktitle={ICRA},
  year={2024}
}

@article{cao2025siamese,
  title={Siamese Adaptive Network-Based Accurate and Robust Visual Object Tracking Algorithm for Quadrupedal Robots},
  author={Cao, Zhengcai and Li, Junnian and Shao, Shibo and Zhang, Dong and Zhou, MengChu},
  journal={IEEE Transactions on Cybernetics},
  year={2025},
  publisher={IEEE}
}

@article{li2024move,
  title={{MOVE:} {Multi-skill} Omnidirectional Legged Locomotion with Limited View in {3D} Environments},
  author={Li, Songbo and Luo, Shixin and Wu, Jun and Zhu, Qiuguo},
  journal={arXiv preprint arXiv:2412.03353},
  year={2024}
}

@inproceedings{li2024stereonavnet,
  title={{StereoNavNet:} {Learning} to Navigate using Stereo Cameras with Auxiliary Occupancy Voxels},
  author={Li, Hongyu and Pad{\i}r, Ta{\c{s}}k{\i}n and Jiang, Huaizu},
  booktitle={IROS},
  year={2024}
}

@inproceedings{cheng2024quadruped,
  title={Quadruped robot traversing {3D} complex environments with limited perception},
  author={Cheng, Yi and Liu, Hang and Pan, Guoping and Liu, Houde and Ye, Linqi},
  booktitle={IROS},
  year={2024}
}

@inproceedings{ming2024inversematrixvt3d,
  title={{InverseMatrixVT3D:} {An} Efficient Projection Matrix-Based Approach for {3D} Occupancy Prediction},
  author={Ming, Zhenxing and Berrio, Julie Stephany and Shan, Mao and Worrall, Stewart},
  booktitle={IROS},
  year={2024}
}

@article{dong2025marg,
  title={{MARG:} {Mastering} Risky Gap Terrains for Legged Robots with Elevation Mapping},
  author={Dong, Yinzhao and Ma, Ji and Zhao, Liu and Li, Wanyue and Lu, Peng},
  journal={IEEE Transactions on Robotics},
  year={2025},
  publisher={IEEE}
}

@article{niijima2025realtime_multi_plane,
  title={Real-time Multi-Plane Segmentation Based on {GPU} Accelerated High-Resolution {3D} Voxel Mapping for Legged Robot Locomotion},
  author={Niijima, Shun and Tsuzaki, Ryoichi and Takasugi, Noriaki and Kinoshita, Masaya},
  journal={arXiv preprint arXiv:2510.01592},
  year={2025}
}

@inproceedings{li2024sscbench,
  title={{SSCBench:} {A} Large-Scale {3D} Semantic Scene Completion Benchmark for Autonomous Driving},
  author={Yiming Li and
                  Sihang Li and
                  Xinhao Liu and
                  Moonjun Gong and
                  Kenan Li and
                  Nuo Chen and
                  Zijun Wang and
                  Zhiheng Li and
                  Tao Jiang and
                  Fisher Yu and
                  Yue Wang and
                  Hang Zhao and
                  Zhiding Yu and
                  Chen Feng},
  booktitle={IROS},
  year={2024}
}

@inproceedings{yang2021semantic,
  title={Semantic Segmentation-assisted Scene Completion for {LiDAR} Point Clouds},
  author={Xuemeng Yang and
                  Hao Zou and
                  Xin Kong and
                  Tianxin Huang and
                  Yong Liu and
                  Wanlong Li and
                  Feng Wen and
                  Hongbo Zhang},
  booktitle={IROS},
  year={2021}
}

@inproceedings{zou2021up,
  title={Up-to-down network: Fusing multi-scale context for {3D} semantic scene completion},
  author={Hao Zou and
                  Xuemeng Yang and
                  Tianxin Huang and
                  Chujuan Zhang and
                  Yong Liu and
                  Wanlong Li and
                  Feng Wen and
                  Hongbo Zhang},
  booktitle={IROS},
  year={2021}
}

@inproceedings{mei2023ssc,
  title={{SSC-RS:} {Elevate} {LiDAR} semantic scene completion with representation separation and {BEV} fusion},
  author={Mei, Jianbiao and Yang, Yu and Wang, Mengmeng and Huang, Tianxin and Yang, Xuemeng and Liu, Yong},
  booktitle={IROS},
  year={2023}
}

@inproceedings{luo2025omnidirectional,
  title={Omnidirectional Multi-Object Tracking},
  author={Kai Luo and
                  Hao Shi and
                  Sheng Wu and
                  Fei Teng and
                  Mengfei Duan and
                  Chang Huang and
                  Yuhang Wang and
                  Kaiwei Wang and
                  Kailun Yang},
  booktitle={CVPR},
  year={2025}
}

@inproceedings{xin2024robotic,
  title={A Robotic-centric Paradigm for {3D} Human Tracking Under Complex Environments Using Multi-modal Adaptation},
  author={Shuo Xin and
                  Zhen Zhang and
                  Liang Liu and
                  Xiaojun Hou and
                  Deye Zhu and
                  Mengmeng Wang and
                  Yong Liu},
  booktitle={IROS},
  year={2024}
}

@inproceedings{aegidius2025watch,
  title={Watch Your {STEPP:} {Semantic} Traversability Estimation Using Pose Projected Features},
  author={Sebastian {\AE}gidius and
                  Dennis Hadjivelichkov and
                  Jianhao Jiao and
                  Jonathan Embley{-}Riches and
                  Dimitrios Kanoulas},
  booktitle={ICRA},
  year={2025}
}

@article{han2025space,
  title={Space-Aware Instruction Tuning: Dataset and Benchmark for Guide Dog Robots Assisting the Visually Impaired},
  author={Han, ByungOk and Yun, Woo-han and Seo, Beom-Su and Kim, Jaehong},
  journal={arXiv preprint arXiv:2502.07183},
  year={2025}
}

@article{zheng2025doracamom,
  title={Doracamom: Joint {3D} Detection and Occupancy Prediction with Multi-view {4D} Radars and Cameras for Omnidirectional Perception},
  author={Lianqing Zheng and
                  Jianan Liu and
                  Runwei Guan and
                  Long Yang and
                  Shouyi Lu and
                  Yuanzhe Li and
                  Xiaokai Bai and
                  Jie Bai and
                  Zhixiong Ma and
                  Hui{-}Liang Shen and
                  Xichan Zhu},
  journal={arXiv preprint arXiv:2501.15394},
  year={2025}
}

@inproceedings{deng2021vision,
  title={Vision-based navigation for a small-scale quadruped robot Pegasus-Mini},
  author={Ganyu Deng and
                  Jianwen Luo and
                  Caiming Sun and
                  Dongwei Pan and
                  Longyao Peng and
                  Ning Ding and
                  Aidong Zhang},
  booktitle={ROBIO},
  year={2021}
}

@article{mei2024quadrupedgpt,
  title={{QuadrupedGPT:} {Towards} a Versatile Quadruped Agent in Open-ended Worlds},
  author={Mei, Yuting and Wang, Ye and Zheng, Sipeng and Jin, Qin},
  journal={arXiv preprint arXiv:2406.16578},
  year={2024}
}

@article{wang2024he_nav,
  title={{HE-Nav:} {A} High-Performance and Efficient Navigation System for Aerial-Ground Robots in Cluttered Environments},
  author={Junming Wang and
                  Zekai Sun and
                  Xiuxian Guan and
                  Tianxiang Shen and
                  Dong Huang and
                  Zongyuan Zhang and
                  Tianyang Duan and
                  Fangming Liu and
                  Heming Cui},
  journal={IEEE Robotics and Automation Letters},
  year={2024},
  publisher={IEEE}
}

@inproceedings{wang2024occrwkv,
  title={{OccRWKV:} {Rethinking} Efficient {3D} Semantic Occupancy Prediction with Linear Complexity},
  author={Junming Wang and
                  Wei Yin and
                  Xiaoxiao Long and
                  Xingyu Zhang and
                  Zebin Xing and
                  Xiaoyang Guo and
                  Qian Zhang},
  booktitle={ICRA},
  year={2025}
}

@inproceedings{zhang2022vision,
  title={Vision-assisted localization and terrain reconstruction with quadruped robots},
  author={Zhang, Chengyang and Zhang, Jiashi and Wu, Jun and Zhu, Qiuguo},
  booktitle={IROS},
  year={2022}
}

@inproceedings{zhao2024lowrankocc,
  title={{LowRankOcc:} {Tensor} decomposition and low-rank recovery for vision-based {3D} semantic occupancy prediction},
  author={Zhao, Linqing and Xu, Xiuwei and Wang, Ziwei and Zhang, Yunpeng and Zhang, Borui and Zheng, Wenzhao and Du, Dalong and Zhou, Jie and Lu, Jiwen},
  booktitle={CVPR},
  year={2024}
}

@inproceedings{song2017semantic,
  title={Semantic scene completion from a single depth image},
  author={Shuran Song and
                  Fisher Yu and
                  Andy Zeng and
                  Angel X. Chang and
                  Manolis Savva and
                  Thomas A. Funkhouser},
  booktitle={CVPR},
  year={2017}
}

@inproceedings{tang2024sparseocc,
  title={{SparseOcc:} {Rethinking} Sparse Latent Representation for Vision-Based Semantic Occupancy Prediction},
  author={Tang, Pin and Wang, Zhongdao and Wang, Guoqing and Zheng, Jilai and Ren, Xiangxuan and Feng, Bailan and Ma, Chao},
  booktitle={CVPR},
  year={2024}
}

@inproceedings{roldao2020lmscnet,
  title={{LMSCNet:} {Lightweight} Multiscale {3D} Semantic Completion},
  author={Roldao, Luis and de Charette, Raoul and Verroust-Blondet, Anne},
  booktitle={3DV},
  year={2020}
}

@inproceedings{yan2021sparse,
  title={Sparse Single Sweep {LiDAR} Point Cloud Segmentation via Learning Contextual Shape Priors from Scene Completion},
  author={Xu Yan and
                  Jiantao Gao and
                  Jie Li and
                  Ruimao Zhang and
                  Zhen Li and
                  Rui Huang and
                  Shuguang Cui},
  booktitle={AAAI},
  year={2021}
}

@inproceedings{xia2023scpnet,
  title={{SCPNet:} {Semantic} Scene Completion on Point Cloud},
  author={Zhaoyang Xia and
                  Youquan Liu and
                  Xin Li and
                  Xinge Zhu and
                  Yuexin Ma and
                  Yikang Li and
                  Yuenan Hou and
                  Yu Qiao},
  booktitle={CVPR},
  year={2023}
}

@inproceedings{cao2022monoscene,
  title={{MonoScene:} {Monocular} {3D} semantic scene completion},
  author={Cao, Anh-Quan and de Charette, Raoul},
  booktitle={CVPR},
  year={2022}
}

@inproceedings{xue2024bi_ssc,
  title={{Bi-SSC:} {Geometric-semantic} Bidirectional Fusion for Camera-based {3D} Semantic Scene Completion},
  author={Xue, Yujie and Li, Ruihui and Wu, Fan and Tang, Zhuo and Li, Kenli and Duan, Mingxing},
  booktitle={CVPR},
  year={2024}
}

@inproceedings{zhang2023occformer,
  title={{OccFormer:} {Dual-path} Transformer for Vision-based {3D} Semantic Occupancy Prediction},
  author={Zhang, Yunpeng and Zhu, Zheng and Du, Dalong},
  booktitle={ICCV},
  year={2023}
}

@article{mei2023camera,
  title={Camera-Based {3D} Semantic Scene Completion With Sparse Guidance Network},
  author={Jianbiao Mei and
                  Yu Yang and
                  Mengmeng Wang and
                  Junyu Zhu and
                  Jongwon Ra and
                  Yukai Ma and
                  Laijian Li and
                  Yong Liu},
  journal={IEEE Transactions on Image Processing},
  year={2024},
  publisher={IEEE}
}

@article{shi2025offboard,
  title={Offboard occupancy refinement with hybrid propagation for autonomous driving},
  author={Shi, Hao and Wang, Song and Zhang, Jiaming and Yin, Xiaoting and Wang, Guangming and Zhu, Jianke and Yang, Kailun and Wang, Kaiwei},
  journal={IEEE Transactions on Intelligent Transportation Systems},
  year={2025},
  publisher={IEEE}
}

@inproceedings{jang2024talos,
  title={{TALoS:} {Enhancing} Semantic Scene Completion via Test-time Adaptation on the Line of Sight},
  author={Jang, Hyun-Kurl and Kim, Jihun and Kweon, Hyeokjun and Yoon, Kuk-Jin},
  booktitle={NeurIPS},
  year={2024}
}

@article{gan2024gaussianocc,
  title={{GaussianOcc:} {Fully} Self-supervised and Efficient {3D} Occupancy Estimation with Gaussian Splatting},
  author={Gan, Wanshui and Liu, Fang and Xu, Hongbin and Mo, Ningkai and Yokoya, Naoto},
  journal={arXiv preprint arXiv:2408.11447},
  year={2024}
}

@article{boeder2025gaussianflowocc,
  title={{GaussianFlowOcc:} {Sparse} and Weakly Supervised Occupancy Estimation using Gaussian Splatting and Temporal Flow},
  author={Boeder, Simon and Gigengack, Fabian and Risse, Benjamin},
  journal={arXiv preprint arXiv:2502.17288},
  year={2025}
}

@article{lin2025one_flight,
  title={One flight over the gap: A survey from perspective to panoramic vision},
  author={Xin Lin and
                  Xian Ge and
                  Dizhe Zhang and
                  Zhaoliang Wan and
                  Xianshun Wang and
                  Xiangtai Li and
                  Wenjie Jiang and
                  Bo Du and
                  Dacheng Tao and
                  Ming{-}Hsuan Yang and
                  Lu Qi},
  journal={arXiv preprint arXiv:2509.04444},
  year={2025}
}

@article{cui2025humanoid,
  title={Humanoid Occupancy: Enabling A Generalized Multimodal Occupancy Perception System on Humanoid Robots},
  author={Wei Cui and
                  Haoyu Wang and
                  Wenkang Qin and
                  Yijie Guo and
                  Gang Han and
                  Wen Zhao and
                  Jiahang Cao and
                  Zhang Zhang and
                  Jiaru Zhong and
                  Jingkai Sun and
                  Pihai Sun and
                  Shuai Shi and
                  Botuo Jiang and
                  Jiahao Ma and
                  Jiaxu Wang and
                  Hao Cheng and
                  Zhichao Liu and
                  Yang Wang and
                  Zheng Zhu and
                  Guan Huang and
                  Jian Tang and
                  Qiang Zhang},
  journal={arXiv preprint arXiv:2507.20217},
  year={2025}
}

@article{zhang2025humanoidpano,
  title={{HumanoidPano:} {Hybrid} Spherical {panoramic-LiDAR} Cross-Modal Perception for Humanoid Robots},
  author={Qiang Zhang and
                  Zhang Zhang and
                  Wei Cui and
                  Jingkai Sun and
                  Jiahang Cao and
                  Yijie Guo and
                  Gang Han and
                  Wen Zhao and
                  Jiaxu Wang and
                  Chenghao Sun and
                  Lingfeng Zhang and
                  Hao Cheng and
                  Yujie Chen and
                  Lin Wang and
                  Jian Tang and
                  Renjing Xu},
  journal={arXiv preprint arXiv:2503.09010},
  year={2025}
}

@article{wu2025omnidirectional,
  title={{OmniOcc:} {Cylindrical} Voxel-Based Semantic Occupancy Prediction for Omnidirectional Vision Systems},
  author={Wu, Chaofan and Li, Jiaheng and Cao, Jinghao and Li, Ming and Du, Sidan and Li, Yang},
  journal={IEEE Access},
  year={2025},
  publisher={IEEE}
}

@article{zhang2025occupancy_world_models_robots,
  title={Occupancy World Model for Robots},
  author={Zhang Zhang and
                  Qiang Zhang and
                  Wei Cui and
                  Shuai Shi and
                  Yijie Guo and
                  Gang Han and
                  Wen Zhao and
                  Jingkai Sun and
                  Jiahang Cao and
                  Jiaxu Wang and
                  Hao Cheng and
                  Xiaozhu Ju and
                  Zhengping Che and
                  Renjing Xu and
                  Jian Tang},
  journal={arXiv preprint arXiv:2505.05512},
  year={2025}
}

@article{zhou2020cylinder3d,
  title={{Cylinder3D:} {An} Effective {3D} Framework for Driving-scene {LiDAR} Semantic Segmentation},
  author={Hui Zhou and
                  Xinge Zhu and
                  Xiao Song and
                  Yuexin Ma and
                  Zhe Wang and
                  Hongsheng Li and
                  Dahua Lin},
  journal={arXiv preprint arXiv:2008.01550},
  year={2020}
}

@article{pan2024co_occ,
  title={{Co-Occ:} {Coupling} Explicit Feature Fusion With Volume Rendering Regularization for Multi-Modal {3D} Semantic Occupancy Prediction},
  author={Pan, Jingyi and Wang, Zipeng and Wang, Lin},
  journal={IEEE Robotics and Automation Letters},
  year={2024},
  publisher={IEEE}
}

@article{guo2025event,
  title={Event-aided Semantic Scene Completion},
  author={Guo, Shangwei and Shi, Hao and Wang, Song and Yin, Xiaoting and Yang, Kailun and Wang, Kaiwei},
  journal={arXiv preprint arXiv:2502.02334},
  year={2025}
}

@inproceedings{li2025occmamba,
  title={{OccMamba:} {Semantic} occupancy prediction with state space models},
  author={Li, Heng and Hou, Yuenan and Xing, Xiaohan and Ma, Yuexin and Sun, Xiao and Zhang, Yanyong},
  booktitle={CVPR},
  year={2025}
}

@inproceedings{wu2025embodiedocc,
  title={{EmbodiedOcc:} {Embodied} {3D} Occupancy Prediction for Vision-based Online Scene Understanding},
  author={Wu, Yuqi and Zheng, Wenzhao and Zuo, Sicheng and Huang, Yuanhui and Zhou, Jie and Lu, Jiwen},
  booktitle={ICCV},
  year={2025}
}

@inproceedings{wang2025embodiedocc++,
  title={{EmbodiedOcc++:} {Boosting} Embodied {3D} Occupancy Prediction with Plane Regularization and Uncertainty Sampler},
  author={Wang, Hao and Wei, Xiaobao and Zhang, Xiaoan and Li, Jianing and Bai, Chengyu and Li, Ying and Lu, Ming and Zheng, Wenzhao and Zhang, Shanghang},
  booktitle={MM},
  year={2025}
}

@inproceedings{liu2024revisit_human_scene_interaction,
  title={Revisit human-scene interaction via space occupancy},
  author={Liu, Xinpeng and Hou, Haowen and Yang, Yanchao and Li, Yong-Lu and Lu, Cewu},
  booktitle={ECCV},
  year={2024}
}

@inproceedings{lu2025vishall3d,
  title={{VisHall3D:} {Monocular} Semantic Scene Completion from Reconstructing the Visible Regions to Hallucinating the Invisible Regions},
  author={Lu, Haoang and Su, Yuanqi and Zhang, Xiaoning and Gao, Longjun and Xue, Yu and Wang, Le},
  booktitle={ICCV},
  year={2025}
}

@inproceedings{chen2025semantic_causality_occ,
  title={Semantic Causality-Aware Vision-Based {3D} Occupancy Prediction},
  author={Chen, Dubing and Zheng, Huan and Zhou, Yucheng and Li, Xianfei and Liao, Wenlong and He, Tao and Peng, Pai and Shen, Jianbing},
  booktitle={ICCV},
  year={2025}
}

@inproceedings{liu2025disentangling,
  title={Disentangling Instance and Scene Contexts for {3D} Semantic Scene Completion},
  author={Liu, Enyu and Yu, En and Chen, Sijia and Tao, Wenbing},
  booktitle={ICCV},
  year={2025}
}

@inproceedings{jevtic2025feed_forward,
  title={Feed-Forward {SceneDINO} for Unsupervised Semantic Scene Completion},
  author={Jevti{\'c}, Aleksandar and Reich, Christoph and Wimbauer, Felix and Hahn, Oliver and Rupprecht, Christian and Roth, Stefan and Cremers, Daniel},
  booktitle={ICCV},
  year={2025}
}

@article{min2025advancing_orad,
  title={Advancing Off-Road Autonomous Driving: The Large-Scale {ORAD-3D} Dataset and Comprehensive Benchmarks},
  author={Min, Chen and Mei, Jilin and Zhai, Heng and Wang, Shuai and Sun, Tong and Kong, Fanjie and Li, Haoyang and Mao, Fangyuan and Liu, Fuyang and Wang, Shuo and Nie, Yiming and Zhu, Qi and Xiao, Liang and Zhao, Dawei and Hu, Yu},
  journal={arXiv preprint arXiv:2510.16500},
  year={2025}
}

@article{wu2025synthetic_v2x,
  title={A Synthetic Benchmark for Collaborative {3D} Semantic Occupancy Prediction in {V2X} Autonomous Driving},
  author={Wu, Hanlin and Lin, Pengfei and Javanmardi, Ehsan and Bao, Naren and Qian, Bo and Si, Hao and Tsukada, Manabu},
  journal={arXiv preprint arXiv:2506.17004},
  year={2025}
}

@inproceedings{huang2025gaussianformer_2,
  title={{GaussianFormer-2:} {Probabilistic} Gaussian Superposition for Efficient {3D} Occupancy Prediction},
  author={Huang, Yuanhui and Thammatadatrakoon, Amonnut and Zheng, Wenzhao and Zhang, Yunpeng and Du, Dalong and Lu, Jiwen},
  booktitle={CVPR},
  year={2025}
}

@inproceedings{huang2024gaussianformer,
  title={{GaussianFormer:} {Scene} as Gaussians for Vision-Based {3D} Semantic Occupancy Prediction},
  author={Huang, Yuanhui and Zheng, Wenzhao and Zhang, Yunpeng and Zhou, Jie and Lu, Jiwen},
  booktitle={ECCV},
  year={2024}
}

@inproceedings{wang2025omni,
  title={{Omni-Perception:} {Omnidirectional} Collision Avoidance for Legged Locomotion in Dynamic Environments},
  author={Zifan Wang and
                  Teli Ma and
                  Yufei Jia and
                  Xun Yang and
                  Jiaming Zhou and
                  Wenlong Ouyang and
                  Qiang Zhang and
                  Junwei Liang},
  booktitle={CoRL},
  year={2025}
}

@inproceedings{semantickitti,
  title={{SemanticKITTI:} {A} Dataset for Semantic Scene Understanding of {LiDAR} Sequences},
  author={Jens Behley and
                  Martin Garbade and
                  Andres Milioto and
                  Jan Quenzel and
                  Sven Behnke and
                  Cyrill Stachniss and
                  J{\"{u}}rgen Gall},
  booktitle={ICCV},
  year={2019}
}

@inproceedings{lee2025soap,
  title={{SOAP:} {Vision-centric} {3D} Semantic Scene Completion with Scene-Adaptive Decoder and Occluded Region-Aware View Projection},
  author={Lee, Hyo-Jun and Koh, Yeong Jun and Kim, Hanul and Kim, Hyunseop and Lee, Yonguk and Lee, Jinu},
  booktitle={CVPR},
  year={2025}
}

@inproceedings{semanticposs,
  title={{SemanticPOSS:} {A} Point Cloud Dataset with Large Quantity of Dynamic Instances},
  author={Pan, Yancheng and Gao, Biao and Mei, Jilin and Geng, Sibo and Li, Chengkun and Zhao, Huijing},
  booktitle={IV},
  year={2020},
}

@INPROCEEDINGS{kitti,
  author = {Andreas Geiger and Philip Lenz and Raquel Urtasun},
  title = {Are we ready for Autonomous Driving? {The} {KITTI} Vision Benchmark Suite},
  booktitle = {CVPR},
  year = {2012}
}

@inproceedings{nyu,
  title={Indoor Segmentation and Support Inference from {RGBD} Images},
  author={Nathan Silberman and
                  Derek Hoiem and
                  Pushmeet Kohli and
                  Rob Fergus},
  booktitle={ECCV},
  year={2012}
}

@inproceedings{wang2024not_voxels_equal,
  title={Not all voxels are equal: Hardness-aware semantic scene completion with self-distillation},
  author={Wang, Song and Yu, Jiawei and Li, Wentong and Liu, Wenyu and Liu, Xiaolu and Chen, Junbo and Zhu, Jianke},
  booktitle={CVPR},
  year={2024}
}

@inproceedings{scannet,
  title={{ScanNet:} {Richly-annotated} {3D} Reconstructions of Indoor Scenes},
  author={Angela Dai and
                  Angel X. Chang and
                  Manolis Savva and
                  Maciej Halber and
                  Thomas A. Funkhouser and
                  Matthias Nie{\ss}ner},
  booktitle={CVPR},
  year={2017}
}

@inproceedings{wang2023openoccupancy,
  title={{OpenOccupancy:} {A} Large Scale Benchmark for Surrounding Semantic Occupancy Perception},
  author={Xiaofeng Wang and
                  Zheng Zhu and
                  Wenbo Xu and
                  Yunpeng Zhang and
                  Yi Wei and
                  Xu Chi and
                  Yun Ye and
                  Dalong Du and
                  Jiwen Lu and
                  Xingang Wang},
  booktitle={ICCV},
  year={2023}
}

@inproceedings{tian2024occ3d,
  title={{Occ3D:} {A} Large-Scale {3D} Occupancy Prediction Benchmark for Autonomous Driving},
  author={Xiaoyu Tian and
                  Tao Jiang and
                  Longfei Yun and
                  Yucheng Mao and
                  Huitong Yang and
                  Yue Wang and
                  Yilun Wang and
                  Hang Zhao},
  booktitle={NeurIPS},
  year={2023}
}

@inproceedings{wang2024label_efficient,
  title={Label-efficient Semantic Scene Completion with Scribble Annotations},
  author={Song Wang and
                  Jiawei Yu and
                  Wentong Li and
                  Hao Shi and
                  Kailun Yang and
                  Junbo Chen and
                  Jianke Zhu},
  booktitle={IJCAI},
  year={2024}
}

@article{zheng2024omnihd,
  title={{OmniHD-Scenes:} {A} next-generation multimodal dataset for autonomous driving},
  author={Lianqing Zheng and
                  Long Yang and
                  Qunshu Lin and
                  Wenjin Ai and
                  Minghao Liu and
                  Shouyi Lu and
                  Jianan Liu and
                  Hongze Ren and
                  Jingyue Mo and
                  Xiaokai Bai and
                  Jie Bai and
                  Zhixiong Ma and
                  Xichan Zhu},
  journal={arXiv preprint arXiv:2412.10734},
  year={2024}
}

@inproceedings{wang2024agrnav,
  title={{AGRNav:} {Efficient} and Energy-Saving Autonomous Navigation for Air-Ground Robots in Occlusion-Prone Environments},
  author={Junming Wang and
                  Zekai Sun and
                  Xiuxian Guan and
                  Tianxiang Shen and
                  Zongyuan Zhang and
                  Tianyang Duan and
                  Dong Huang and
                  Shixiong Zhao and
                  Heming Cui},
  booktitle={ICRA},
  year={2024}
}

@article{patel2025hierarchical,
  title={A Hierarchical Graph-Based Terrain-Aware Autonomous Navigation Approach for Complementary Multimodal Ground-Aerial Exploration},
  author={Akash Patel and
                  Mario Alberto Valdes Saucedo and
                  Nikolaos Stathoulopoulos and
                  Viswa Narayanan Sankaranarayanan and
                  Ilias Tevetzidis and
                  Christoforos Kanellakis and
                  George Nikolakopoulos},
  journal={arXiv preprint arXiv:2505.14859},
  year={2025}
}

@article{wang2025odyssey,
  title={{ODYSSEY:} {Open-world} Quadrupeds Exploration and Manipulation for Long-Horizon Tasks},
  author={Wang, Kaijun and Lu, Liqin and Liu, Mingyu and Jiang, Jianuo and Li, Zeju and Zhang, Bolin and Zheng, Wancai and Yu, Xinyi and Chen, Hao and Shen, Chunhua},
  journal={arXiv preprint arXiv:2508.08240},
  year={2025}
}

@inproceedings{kartmann2021semantic,
  title={Semantic scene manipulation based on {3D} spatial object relations and language instructions},
  author={Kartmann, Rainer and Liu, Danqing and Asfour, Tamim},
  booktitle={Humanoids},
  year={2021}
}

@article{chen2025learning_traversal,
  title={Learning Autonomous and Safe Quadruped Traversal of Complex Terrains Using Multi-Layer Elevation Maps},
  author={Chen, Yeke and Ma, Ji and Luo, Zeren and Han, Yimin and Dong, Yinzhao and Xu, Bowen and Lu, Peng},
  journal={IEEE Robotics and Automation Letters},
  year={2025},
  publisher={IEEE}
}

@article{escontrela2025gaussgym,
  title={{GaussGym:} {An} open-source real-to-sim framework for learning locomotion from pixels},
  author={Escontrela, Alejandro and Kerr, Justin and Allshire, Arthur and Frey, Jonas and Duan, Rocky and Sferrazza, Carmelo and Abbeel, Pieter},
  journal={arXiv preprint arXiv:2510.15352},
  year={2025}
}

@article{li2025manipdreamer3d,
  title={{ManipDreamer3D} : {Synthesizing} Plausible Robotic Manipulation Video with Occupancy-aware {3D} Trajectory},
  author={Li, Ying and Wei, Xiaobao and Chi, Xiaowei and Li, Yuming and Zhao, Zhongyu and Wang, Hao and Ma, Ningning and Lu, Ming and Zhang, Shanghang},
  journal={arXiv preprint arXiv:2509.05314},
  year={2025}
}

@inproceedings{dengler2025efficient_manipulation,
  title={Efficient Manipulation-Enhanced Semantic Mapping With Uncertainty-Informed Action Selection},
  author={Dengler, Nils and M{\"u}cke, Jesper and Menon, Rohit and Bennewitz, Maren},
  booktitle={Humanoids},
  year={2025}
}

@article{liu2025starlink,
  title={The Starlink Robot: A Platform and Dataset for Mobile Satellite Communication},
  author={Liu, Boyi and Zhang, Qianyi and Yang, Qiang and Jiao, Jianhao and Chauhan, Jagmohan and Kanoulas, Dimitrios},
  journal={arXiv preprint arXiv:2506.19781},
  year={2025}
}

@article{ren2024grounded,
  title={Grounded {SAM}: {Assembling} open-world models for diverse visual tasks},
  author={Tianhe Ren and
                  Shilong Liu and
                  Ailing Zeng and
                  Jing Lin and
                  Kunchang Li and
                  He Cao and
                  Jiayu Chen and
                  Xinyu Huang and
                  Yukang Chen and
                  Feng Yan and
                  Zhaoyang Zeng and
                  Hao Zhang and
                  Feng Li and
                  Jie Yang and
                  Hongyang Li and
                  Qing Jiang and
                  Lei Zhang},
  journal={arXiv preprint arXiv:2401.14159},
  year={2024}
}

@inproceedings{scaramuzza2006toolbox,
  title={A toolbox for easily calibrating omnidirectional cameras},
  author={Scaramuzza, Davide and Martinelli, Agostino and Siegwart, Roland},
  booktitle={IROS},
  year={2006}
}

@inproceedings{zhang2023adding,
  title={Adding conditional control to text-to-image diffusion models},
  author={Zhang, Lvmin and Rao, Anyi and Agrawala, Maneesh},
  booktitle={ICCV},
  year={2023}
}

@article{xu2022fast,
  title={{FAST-LIO2:} {Fast} Direct {LiDAR}-Inertial Odometry},
  author={Xu, Wei and Cai, Yixi and He, Dongjiao and Lin, Jiarong and Zhang, Fu},
  journal={IEEE Transactions on Robotics},
  year={2022},
  publisher={IEEE}
}

@inproceedings{dosovitskiy2017carla,
  title={{CARLA:} {An} open urban driving simulator},
  author={Dosovitskiy, Alexey and Ros, German and Codevilla, Felipe and Lopez, Antonio and Koltun, Vladlen},
  booktitle={CoRL},
  year={2017}
}

@inproceedings{cordts2016cityscapes,
  title={The cityscapes dataset for semantic urban scene understanding},
  author={Cordts, Marius and Omran, Mohamed and Ramos, Sebastian and Rehfeld, Timo and Enzweiler, Markus and Benenson, Rodrigo and Franke, Uwe and Roth, Stefan and Schiele, Bernt},
  booktitle={CVPR},
  year={2016}
}

@article{gao2022review,
  title={Review on panoramic imaging and its applications in scene understanding},
  author={Gao, Shaohua and Yang, Kailun and Shi, Hao and Wang, Kaiwei and Bai, Jian},
  journal={IEEE Transactions on Instrumentation and Measurement},
  year={2022},
  publisher={IEEE}
}

@article{chen2025point,
  title={{Point-MoE:} {Towards} Cross-Domain Generalization in {3D} Semantic Segmentation via Mixture-of-Experts},
  author={Chen, Xuweiyi and Zhou, Wentao and RoyChowdhury, Aruni and Cheng, Zezhou},
  journal={arXiv preprint arXiv:2505.23926},
  year={2025}
}

@inproceedings{zha2024point,
  title={Point Cloud Mixture-of-Domain-Experts Model for {3D} Self-supervised Learning},
  author={Zha, Yaohua and Dai, Tao and Guo, Hang and Wang, Yanzi and Chen, Bin and Chen, Ke and Xia, Shu-Tao},
  booktitle={IJCAI},
  year={2025}
}

@inproceedings{AdamW,
  title={Decoupled weight decay regularization},
  author={Loshchilov, Ilya and Hutter, Frank},
  booktitle={ICLR},
  year={2019}
}

@inproceedings{MonoDepth2,
  title={Digging into self-supervised monocular depth estimation},
  author={Cl{\'{e}}ment Godard and
                  Oisin Mac Aodha and
                  Michael Firman and
                  Gabriel J. Brostow},
  booktitle={CVPR},
  year={2019}
}

@inproceedings{LSS,
  title={Lift, splat, shoot: Encoding images from arbitrary camera rigs by implicitly unprojecting to {3D}},
  author={Philion, Jonah and Fidler, Sanja},
  booktitle={ECCV},
  year={2020}
}

@inproceedings{Mask2Former,
  title={Masked-attention mask transformer for universal image segmentation},
  author={Cheng, Bowen and Misra, Ishan and Schwing, Alexander G and Kirillov, Alexander and Girdhar, Rohit},
  booktitle={CVPR},
  year={2022}
}

@inproceedings{DETR,
  title={Deformable {DETR}: {Deformable} Transformers for End-to-End Object Detection},
  author={Zhu, Xizhou and Su, Weijie and Lu, Lewei and Li, Bin and Wang, Xiaogang and Dai, Jifeng},
  booktitle={ICLR},
  year={2021}
}

@inproceedings{EfficientNet,
  title={Efficientnet: Rethinking model scaling for convolutional neural networks},
  author={Mingxing Tan and
                  Quoc V. Le},
  booktitle={ICML},
  year={2019}
}

@article{Adam,
  title={Adam: A method for stochastic optimization},
  author={Diederik P. Kingma and
                  Jimmy Ba},
  journal={ICLR},
  year={2015}
}

@article{3dgs,
  title={{3D} Gaussian Splatting for Real-Time Radiance Field Rendering},
  author={Bernhard Kerbl and
                  Georgios Kopanas and
                  Thomas Leimk{\"{u}}hler and
                  George Drettakis},
  journal={ACM Transactions on Graphics (TOG)},
  year={2023},
  publisher={ACM New York, NY, USA}
}

@article{shi2024beyond,
  title={Beyond the field-of-view: Enhancing scene visibility and perception with clip-recurrent transformer},
  author={Shi, Hao and Jiang, Qi and Yang, Kailun and Yin, Xiaoting and Ni, Huajian and Wang, Kaiwei},
  journal={IEEE Transactions on Intelligent Vehicles},
  year={2025},
  publisher={IEEE}
}

@inproceedings{zheng2024occworld,
  title={{OccWorld:} {Learning} a {3D} Occupancy World Model for Autonomous Driving},
  author={Zheng, Wenzhao and Chen, Weiliang and Huang, Yuanhui and Zhang, Borui and Duan, Yueqi and Lu, Jiwen},
  booktitle={ECCV},
  year={2024}
}

@inproceedings{liu2024let,
  title={Let {Occ} Flow: Self-Supervised {3D} Occupancy Flow Prediction},
  author={Liu, Yili and Mou, Linzhan and Yu, Xuan and Han, Chenrui and Mao, Sitong and Xiong, Rong and Wang, Yue},
  booktitle={CoRL},
  year={2024}
}

@article{tan2023ovo,
  title={{OVO:} {Open-vocabulary} Occupancy},
  author={Tan, Zhiyu and Dong, Zichao and Zhang, Cheng and Zhang, Weikun and Ji, Hang and Li, Hao},
  journal={arXiv preprint arXiv:2305.16133},
  year={2023}
}

@inproceedings{zuo2025gaussianworld,
  title={{GaussianWorld:} {Gaussian} World Model for Streaming {3D} Occupancy Prediction},
  author={Zuo, Sicheng and Zheng, Wenzhao and Huang, Yuanhui and Zhou, Jie and Lu, Jiwen},
  booktitle={CVPR},
  year={2025}
}

@inproceedings{feng2025gaussian,
  title={Gaussian-based World Model: Gaussian Priors for Voxel-Based Occupancy Prediction and Future Motion Prediction},
  author={Feng, Tuo and Wang, Wenguan and Yang, Yi},
  booktitle={ICCV},
  year={2025}
}

@inproceedings{chen2025alocc,
  title={{ALOcc:} {Adaptive} Lifting-Based {3D} Semantic Occupancy and Cost Volume-Based Flow Predictions},
  author={Chen, Dubing and Fang, Jin and Han, Wencheng and Cheng, Xinjing and Yin, Junbo and Xu, Chengzhong and Khan, Fahad Shahbaz and Shen, Jianbing},
  booktitle={ICCV},
  year={2025}
}
}

\clearpage
%



\section{More Ablations}

\noindent\textbf{Setup and notation.}
Unless stated otherwise, we evaluate on H3O-Heter with the native equirectangular view and report P/R/IoU/mIoU on non-empty voxels. 
We additionally include QuadOcc only where the raw annulus is required (DP-ER). Grid bounds and taxonomy follow the main paper. For H3O we study both native $64{\times}64{\times}8$ and $128{\times}128{\times}16$ resolutions. Timing is measured with the batch size $1$ on a single NVIDIA RTX-4090 GPU (FP32). We report Params (M), FPS, and Mem (GB).

\subsection{Hierarchical AMoE-3D: Number of Experts $K$}
\label{sec:ablation_moe}
\begin{bluequestion}
\noindent\textit{Question.} How does the number of 3D experts affect accuracy?
\end{bluequestion}

\begin{table}[h]
\centering
\caption{A1: AMoE-3D experts on H3O-Heter. $K{=}1$ reduces to a fused bottleneck. Gate is 3D Gradient-Energy by default unless noted. We report precision (P), recall (R), IoU, and mIoU on non-empty voxels under the official H3O-Heter protocol.}
\label{tab:ablation_moe}
\resizebox{\columnwidth}{!}{
\begin{tabular}{ll|cccc}
\midrule
\rowcolor{gray!10}
$K$ & \textbf{Gate}        & \textbf{P} & \textbf{R} & \textbf{IoU} & \textbf{mIoU} \\
\midrule
\midrule
1   & --          & 65.38 & 61.02 & 46.12 & 29.68 \\
2   & GradEnergy3D  & 66.79 & 63.69 & 48.37 & 31.25 \\
\rowcolor{gray!10}
4 & GradEnergy3D & 67.89 & \textbf{64.77} & \textbf{49.58} & \textbf{32.23} \\
8   & GradEnergy3D  & \textbf{68.08} & 64.39 & 49.46 & 32.03 \\
4   & Uniform     & 66.21 & 63.04 & 47.70 & 31.13 \\
4   & Top-$k$     & 67.18 & 64.06 & 48.79 & 31.84 \\
\midrule
\end{tabular}}
\end{table}

\noindent\textit{Findings.}
Increasing $K$ from $1$ to $4$ consistently improves robustness on H3O-Heter. Moving to $K{=}8$ slightly increases precision but reduces recall, leading to marginal declines in IoU/mIoU due to expert fragmentation and noisier routing. 3D Gradient-Energy gating (Sec.~3.6) prioritizes high-gradient regions (class boundaries and thin structures), yielding better voxel-wise expert assignment than uniform or naive Top-$k$ routing~\cite{chen2025point}. $K{=}4$ achieves the best accuracy–efficiency trade-off and is our default. We also note related progress in point-cloud domain adaptation: Point-MoDE~\cite{zha2024point} employs a mixture-of-domain-experts to enhance cross-domain generalization, reaching conclusions consistent with ours.

\vspace{0.6em}
\subsection{Horizontal Field-of-View: from $90^\circ$ to $360^\circ$}
\label{sec:ablation_fov}
\begin{bluequestion}
\noindent\textit{Question.} How much does full $360^\circ$ surround help, and how do methods degrade when evaluated under narrower FoVs?
\end{bluequestion}

\paragraph{Protocol (train-once, test-with-crops).}
Unless otherwise stated, all methods are trained with full $360^\circ$ ER panoramas. At evaluation time, we reduce the horizontal FoV by cropping the panorama \emph{without finetuning} the models. We use forward-centered cropping so the input spatial extent truly shrinks as the FoV narrows (thereby reflecting realistic compute/memory savings).

\begin{table}[h]
\centering
\caption{A2: FoV ablation on H3O-Heter (ER panorama; horizontal crop at eval only; no retraining). We report overall metrics on non-empty voxels. Both OneOcc and MonoScene~\cite{cao2022monoscene} are \emph{trained once at $360^\circ$}, and \emph{evaluated} at narrower FoVs via forward-centered cropping.}

\label{tab:ablation_fov}
\resizebox{\columnwidth}{!}{
\begin{tabular}{cclccccccc}
\midrule
\rowcolor{gray!10}
\textbf{FoV} & \textbf{Visible} & \textbf{Method} & \textbf{\#Params} & \textbf{FPS} $\uparrow$ & \textbf{Mem} $\downarrow$ & \textbf{P} $\uparrow$ & \textbf{R} $\uparrow$ & \textbf{IoU} $\uparrow$ & \textbf{mIoU} $\uparrow$ \\
\midrule
\midrule
\multirow{2}{*}{$90^\circ$}  & \multirow{2}{*}{$25\%$}  & MonoScene~\cite{cao2022monoscene} & 146.13M & 13.07 & 1.72GB & 56.14 & 13.59 & 12.28 & 4.65 \\
                            &                            & OneOcc (ours) & 101.76M & 24.19 & 1.65GB & 58.65 & 15.43 & 13.92 & 7.35 \\
\midrule
\multirow{2}{*}{$180^\circ$} & \multirow{2}{*}{$50\%$}  & MonoScene~\cite{cao2022monoscene} & 146.13M & 12.63 & 1.78GB & 63.40 & 29.64 & 25.31 & 13.74 \\
                            &                            & OneOcc (ours) & 101.76M & 19.24 & 1.70GB & 62.68 & 31.62 & 26.61 & 17.27 \\
\midrule
\multirow{2}{*}{$270^\circ$} & \multirow{2}{*}{$75\%$}  & MonoScene~\cite{cao2022monoscene} & 146.13M & 10.08 & 2.09GB & 65.46 & 41.19 & 33.84 & 18.20 \\
                            &                            & OneOcc (ours) & 101.76M & 16.13 & 1.76GB & 65.79 & 47.53 & 38.11 & 24.28 \\
\midrule
\multirow{2}{*}{$360^\circ$} & \multirow{2}{*}{$100\%$} & MonoScene~\cite{cao2022monoscene} & 146.13M & 8.29 & 2.60GB & 67.39 & 55.00 & 43.44 & 24.15 \\
& & \cellcolor{gray!10}OneOcc (ours) &
\cellcolor{gray!10}101.76M & \cellcolor{gray!10}14.30 &
\cellcolor{gray!10}1.82GB & \cellcolor{gray!10}67.89 &
\cellcolor{gray!10}64.77 & \cellcolor{gray!10}49.58 &
\cellcolor{gray!10}32.23 \\
\midrule
\end{tabular}}
\end{table}

\noindent\textit{Findings.}
\begin{compactitem}
\item \textbf{Train-once, test-with-crops.} For OneOcc, mIoU scales near-monotonically with FoV: $+9.92$ (90$\to$180), $+7.01$ (180$\to$270), and $+7.95$ (270$\to$360), totaling \textbf{$+24.88$} from $90^\circ$ to $360^\circ$ (\textbf{$+77.2\%$} rel.). MonoScene also improves but less (\textbf{$+19.50$} total). This indicates that \emph{surround cues translate into long-range context}, with stronger returns when the azimuthal ring is closed.

\item \textbf{Closing-the-ring effect.} The last increment (270$\to$360) is sizable for both methods, and larger for OneOcc (\textbf{$+7.95$} vs. \textbf{$+5.95$} mIoU), evidencing a \emph{nonlinear benefit} when the $360^\circ$ ring continuity becomes complete.

\item \textbf{Precision parity, Recall advantage.} OneOcc and MonoScene have similar precision at a given FoV (\textit{e.g.}, $67.89$ vs. $67.39$ at $360^\circ$), while the \emph{recall gap} dominates OneOcc’s advantage: \textbf{$+9.77$} recall points at $360^\circ$ (and \textbf{$+6.34$} at $270^\circ$). This matches our design goal: cylindrical alignment and dual-projection preserve azimuthal continuity, yielding \emph{more complete} occupancy recovery as FoV expands.

\item \textbf{Absolute margins grow with FoV.} OneOcc surpasses MonoScene by $+2.70$ / $+3.53$ / $+6.08$ / $+8.08$ mIoU at \{$90^\circ,180^\circ,270^\circ,360^\circ$\}, respectively. Relative gains are \textbf{$+58.1\%$}, \textbf{$+25.7\%$}, \textbf{$+33.4\%$}, \textbf{$+33.5\%$}.

\item \textbf{Accuracy--throughput Pareto (AET).} We define
$\text{AET}=\text{mIoU}\times\text{FPS}$ to summarize accuracy at a given runtime budget.
OneOcc vs.\ MonoScene AET is:
\vspace{-0.2em}
\[
\setlength{\jot}{0pt}
\begin{aligned}
90^\circ  &:\; 177.8~\text{vs}~60.8~(\mathbf{2.93}\times) \\
180^\circ &:\; 332.3~\text{vs}~173.5~(\mathbf{1.91}\times) \\
270^\circ &:\; 391.6~\text{vs}~183.5~(\mathbf{2.13}\times) \\
360^\circ &:\; 460.9~\text{vs}~200.2~(\mathbf{2.30}\times)
\end{aligned}
\]

Across all FoVs, OneOcc is \emph{Pareto-superior} in accuracy--throughput.

\item \textbf{Memory/parameter efficiency.} OneOcc uses 30.3\% fewer parameters (101.76M vs. 146.13M). Its mIoU per GB is consistently higher:
$4.45$ vs. $2.70$ ($90^\circ$),
$10.16$ vs. $7.72$ ($180^\circ$),
$13.80$ vs. $8.71$ ($270^\circ$),
$17.71$ vs. $9.29$ ($360^\circ$),
\textit{i.e.}, \textbf{$1.31$--$1.91\times$} memory-normalized gains.

\item \textbf{Practical iso-accuracy choices.} If targeting $\sim\!24$ mIoU, \emph{OneOcc@270$^\circ$} achieves 24.28 mIoU at 16.13 FPS and 1.76 GB, matching/exceeding \emph{MonoScene@360$^\circ$} (24.15 mIoU at 8.29 FPS, 2.60 GB). Thus, for similar accuracy, OneOcc needs \textbf{$\sim$2$\times$} throughput and $-0.84$ GB memory. Likewise, \emph{OneOcc@180$^\circ$} (17.27 mIoU, 19.24 FPS) rivals \emph{MonoScene@270$^\circ$} (18.20 mIoU, 10.08 FPS) at roughly \textbf{$\sim$1.9$\times$} throughput.
\end{compactitem}

\noindent\textit{Takeaways.} Full-surround cues are critical; methods that explicitly encode ring continuity not only achieve higher absolute accuracy, but also deliver better \emph{accuracy-per-compute} and \emph{accuracy-per-memory}. In embodied settings, these advantages persist even when FoV must shrink to meet runtime constraints.

\paragraph{Why an ego-centric $360^\circ$ symmetric occupancy grid?}
Unlike automotive datasets (\textit{e.g.}, front-view grids such as SemanticKITTI~\cite{semantickitti}) that emphasize a forward driving cone, embodied agents must reason about objects and free space \emph{all around} the ego: turning-in-place, backtracking, and manipulations behind/aside the agent are common. A \emph{symmetrically} centered $360^\circ$ grid (front/back/left/right balanced) aligns the spatial prior with such behaviors, reduces coordinate bias, and enables globally consistent occlusion reasoning and memory beyond the forward frustum. Empirically, this grid pairs naturally with ring-continuous features and improves cross-directional consistency in long-range context.

\begin{figure}[!t]
\centering
\includegraphics[width=1.0\linewidth]{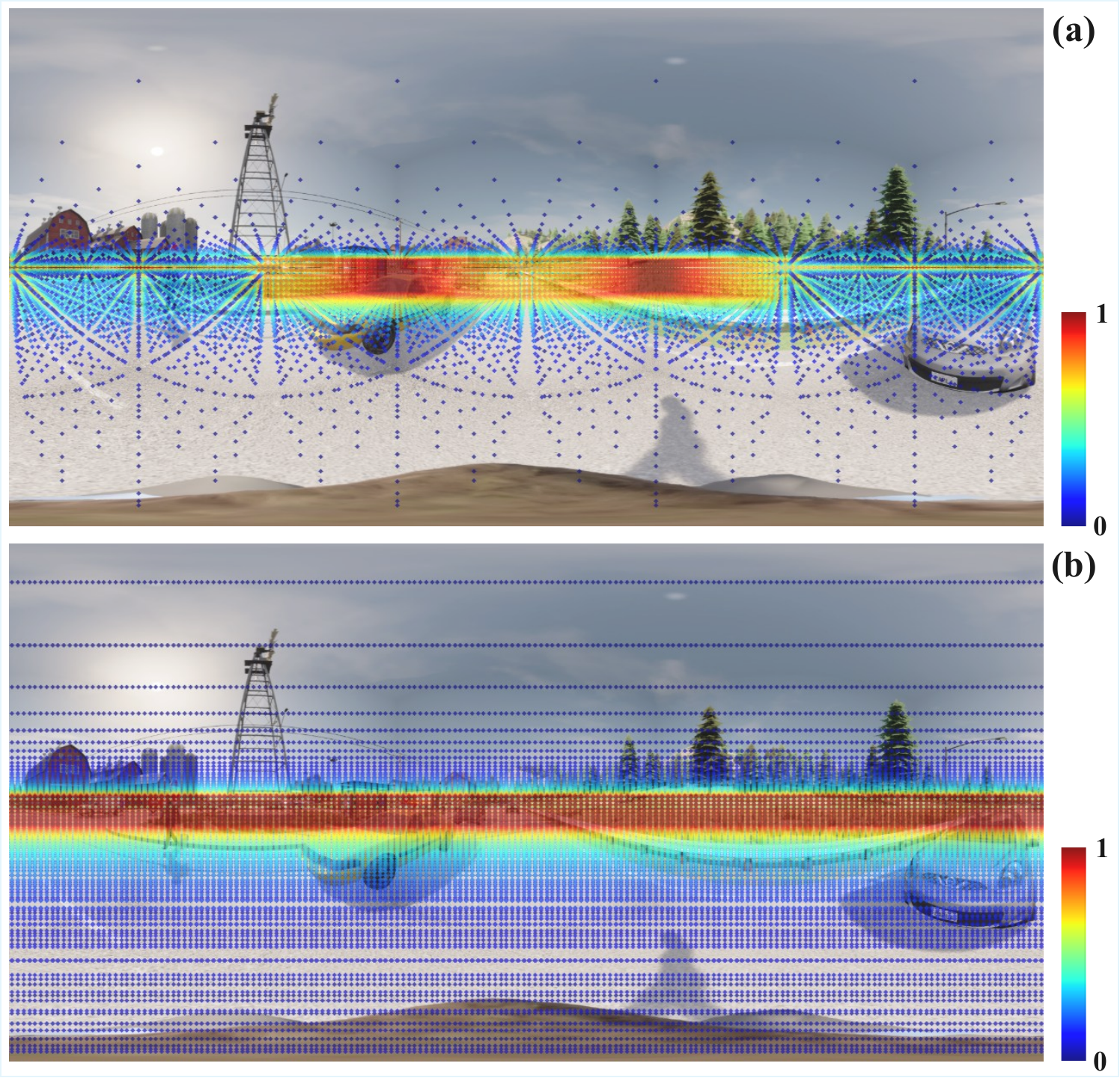}
\caption{\textbf{Cartesian vs. cylindrical voxel projections on a panoramic view.}
  We visualize how different voxel parameterizations project onto an equirectangular panorama in Human360Occ.
  Dots denote voxel centroids projected to the image; the colored band (0--1) encodes the normalized depth distribution of occupied voxels along each azimuthal ray.
  (a) A conventional axis-aligned \emph{Cartesian} grid in world coordinates produces fan-shaped footprints when projected to the panorama: voxels at different heights but similar depth map to tilted rays, and far-range structures such as roads, snow banks, and building façades are squeezed into a narrow band around the equator.
  This favors metrically uniform sampling in 3D but breaks the angular regularity of the panoramic image.
  (b) In contrast, an omnidirectional camera-centric \emph{cylindrical} grid uses equal steps in azimuth and height, so the projected voxel centroids form nearly horizontal sampling rows that follow the equirectangular parameterization and maintain ring continuity for distant structures.
  This complementarity motivates combining both grids in our Bi-Grid voxelization to balance near-field metric fidelity and far-field azimuthal coherence.}
\label{fig:cart_polar_porj}
\end{figure}

\vspace{0.6em}
\subsection{Bi-Grid Voxelization: Cartesian vs.\ Cylinder}
\label{sec:ablation_bgv}
\begin{bluequestion}
\noindent\textit{Question.} Does the camera-aligned cylindrical grid help beyond a standard Cartesian grid?
\end{bluequestion}

\begin{table}[h]
\centering
\caption{A3: Single-grid vs.\ Bi-Grid on H3O-Heter. We stratify \textit{near} using an XY-centered crop (center ratio $0.5$, \textit{i.e.}, front/back/left/right $\pm6.4$\,m) and \textit{far} using the full XY footprint ($\pm12.8$\,m); the vertical range is fixed to $z\in[-2,\,1.2]$\,m.}
\label{tab:ablation_bigrid}
\resizebox{\columnwidth}{!}{
\begin{tabular}{l|ccc|ccccc}
\midrule
\rowcolor{gray!10}
Voxelization & \#Params & FPS $\uparrow$ & Mem $\downarrow$ & P $\uparrow$ & R $\uparrow$ & IoU $\uparrow$ & Near mIoU $\uparrow$ & Far mIoU $\uparrow$ \\
\midrule
\midrule
Cartesian-only       & 101.73M & 14.32 & 1.70GB & 62.54 & 63.18 & 45.84 & 36.42 & 30.56 \\
Cylindrical-only     & 101.73M & 14.33 & 1.78GB & 63.65 & \textbf{65.30} & 47.57 & 34.15 & 31.00 \\
\rowcolor{gray!10}
Bi-Grid & 101.76M & 14.30 & 1.82GB & \textbf{67.89} & 64.77 & \textbf{49.58} & \textbf{37.35} & \textbf{32.23} \\
\midrule
\end{tabular}}
\end{table}

\noindent\textit{Setup.}
Unless otherwise stated, we predict occupancy on a Cartesian grid and evaluate within a fixed vertical range $z\!\in\![-2,\,1.2]$\,m. For range-wise analysis, we follow a panoramic-camera protocol:
\emph{near} uses an XY-centered crop with a center ratio of $0.5$ (\textit{i.e.}, front/back/left/right $\pm6.4$\,m around the ego) and excludes far voxels; 
\emph{far} uses the full XY footprint (front/back/left/right $\pm12.8$\,m).
The far split is intrinsically harder due to larger geometric errors, heavier occlusions, sparser voxels, and higher sampling noise, so many classes naturally obtain higher Recall/IoU in the near window, which in turn boosts the near mIoU.

\noindent\textit{Findings.}
\begin{compactitem}
\item \textbf{Complementary inductive biases.}
A cylindrical grid (aligned with panoramic camera rays and azimuth) preserves ring continuity and equal-angle sampling, favoring \emph{far-field} layout regularities (roads/sidewalks bands, building facades). Compared with Cartesian-only, Cylindrical-only raises Recall by {+2.12} (65.30 vs.\ 63.18) and IoU by {+1.73}, but degrades \emph{near} mIoU by {$-2.27$} (34.15 vs.\ 36.42), while slightly improving \emph{far} mIoU by {+0.44} (31.00 vs.\ 30.56), echoing Cylinder3D-style observations on cylindrical parameterizations for long-range structure~\cite{zhou2020cylinder3d}.
\item \textbf{Bi-Grid synergy with negligible overhead.}
Our Bi-Grid fuses both discretizations and delivers the best of both worlds: over Cartesian-only it gains {+5.35} Precision (67.89 vs.\ 62.54), {+1.59} Recall, and {+3.74} IoU, while consistently lifting \emph{both} ranges (near: +0.93, far: +1.67).
Against Cylindrical-only, Bi-Grid trades a small Recall drop ({$-0.53$}) for a much larger Precision gain (+4.24), resulting in the highest IoU/mIoU overall.
The parameter and runtime overhead are minimal (+0.03\,M params; $\sim$14.3 FPS maintained; +0.12\,GB peak memory).
\item \textbf{Why the overhead is tiny (implementation).}
The voxel centroids for both cartesian and cylindrical grids are precomputed and projected to the images in the \texttt{dataloader}; at inference we only bilinearly sample 2D features at these fixed locations.
We remove the offset-MLP that predicts per-voxel 2D sampling displacements, so the parameter budget barely changes (observed $\Delta\!\approx\!+0.03$\,M) and throughput remains $\sim$14.3 FPS.
The main extra cost is memory: Bi-Grid performs two sets of feature samplings and maintains slightly larger activation buffers, causing a modest peak-memory increase (about +0.12\,GB).
\item \textbf{Why Cartesian still matters near the ego.}
Evaluation and downstream planners operate on a Euclidean (Cartesian) grid; close-range contact geometry (ground, curbs, thin structures, small objects) benefits from uniform metric spacing and axis-aligned neighborhoods. This explains why Cartesian-only is stronger than Cylindrical-only on \emph{near} mIoU (36.42 vs.\ 34.15).
\item \textbf{Panoramic geometry alignment.}
With equirectangular/panoramic imaging, cylindrical bins match the camera’s spherical parameterization and mitigate azimuthal aliasing, improving \emph{far}-field coherence; Cartesian bins better preserve metric fidelity and local topology critical for \emph{near}-field planning. Bi-Grid inherits both advantages.
\item \textbf{Context from SSC/OCC.}
Prior occupancy works (\eg, VoxFormer~\cite{li2023voxformer}, OccFiner~\cite{shi2025offboard}) typically report higher short-range fidelity due to denser observations and stronger priors, while accuracy decays with distance. Our \emph{near}/\emph{far} split makes this explicit: the near window naturally boosts Recall/IoU for many classes, and cylindrical alignment helps counteract the far-range drop; Bi-Grid couples the two.
\end{compactitem}

\noindent\textit{Takeaways.}
(i) Cylindrical improves far-field ring coherence but may underfit metric-accurate contact geometry nearby; 
(ii) Cartesian preserves near-field fidelity but under-exploits azimuthal continuity; 
(iii) For panoramic input with Cartesian output, Bi-Grid is a near-free, robust default that simultaneously improves both \emph{near} and \emph{far} regimes.

\begin{figure*}[!t]
\centering
\includegraphics[width=1.0\textwidth]{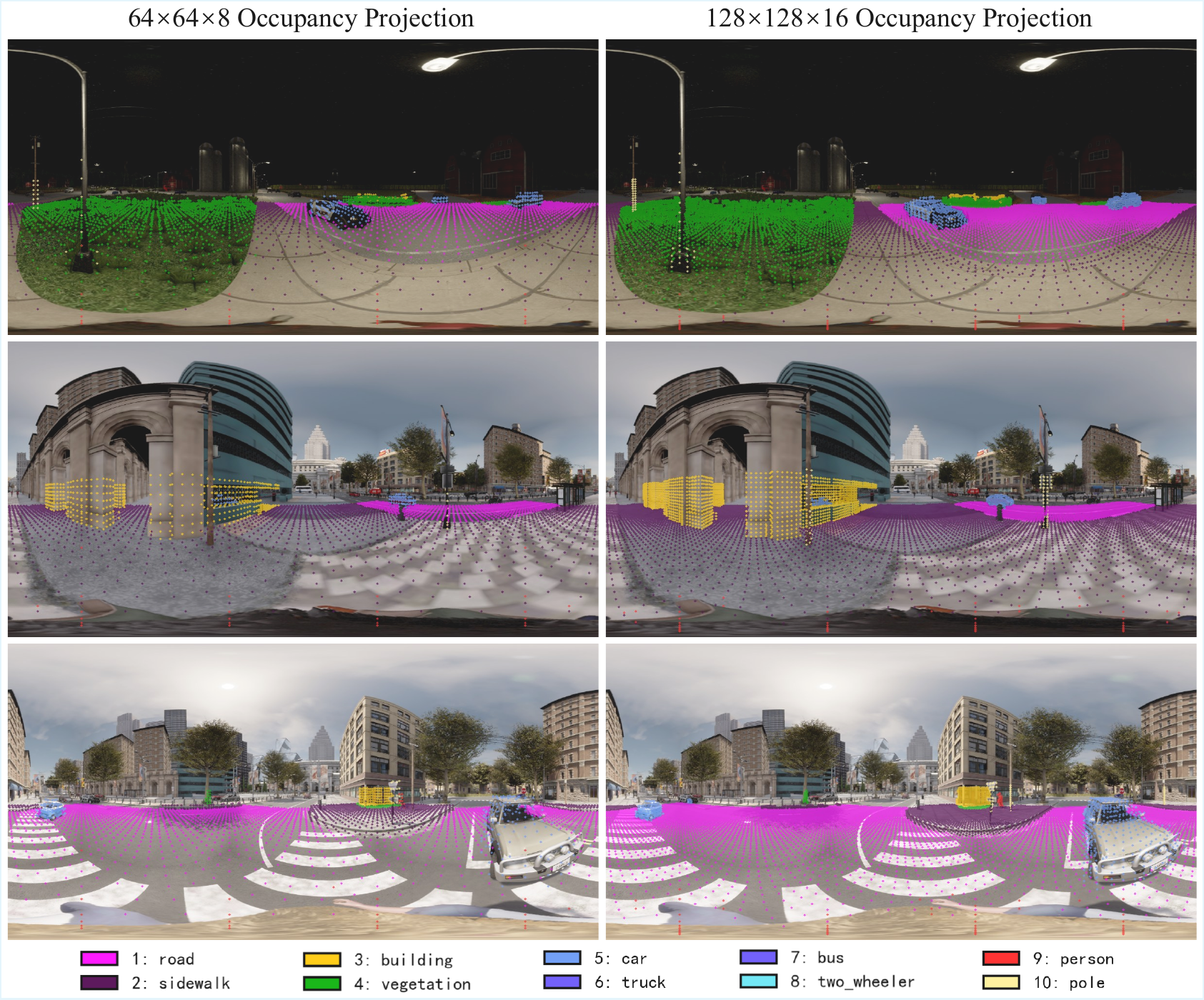}
\caption{\textbf{Semantic ground-truth occupancy projections at different voxel resolutions.}
  We visualize the projection of \emph{ground-truth} semantic voxels from grids of size $64\times64\times8$ (left) and $128\times128\times16$ (right) onto the panoramic image.
  Colored dots follow the legend at the bottom and indicate voxel centers of different semantic classes relevant to legged locomotion, such as road, sidewalk, buildings, vegetation, vehicles, pedestrians, and poles.
  The higher-resolution $128\times128\times16$ grid produces visually denser and more detailed occupancy patterns, with sharper boundaries and more finely sampled thin structures.
  However, the coarser $64\times64\times8$ grid still provides sufficient spatial coverage and granularity for a legged robot to perceive traversable surfaces and nearby obstacles in $360^\circ$ around the body.
  Considering the computation and payload constraints of embodied agents, we adopt $64\times64\times8$ as the default resolution for main results on both datasets.}
\label{fig:vox_res_proj}
\end{figure*}

\vspace{0.6em}
\subsection{Dual-Projection: ER-only vs. Raw-only}
\label{sec:ablation_dper}
\begin{bluequestion}
\noindent\textit{Question.} How much does processing both the equirectangular (ER) panorama \emph{and} the raw annulus help?
\end{bluequestion}

\noindent\textbf{Setup.}
All variants share the same backbone/decoder and training schedule; only the input projection path differs. 
For \emph{ER-only} we unwrap the PAL annulus to equirectangular and feed a single ER encoder. 
For \emph{Raw-only} we process the native annulus with a single raw-path encoder. 
\emph{Dual (ER+Raw)} instantiates both encoders and fuses features before 3D lifting. 
Unless otherwise stated, inference is measured with batch size~1, FP32; ER-based paths use $352\times1216$, and Raw-only uses $512\times512$. 
We report wall-clock throughput (FPS) and peak CUDA memory alongside accuracy (Precision $P$, Recall $R$, IoU, mIoU). 
On H3O (native ER panoramas) we adopt ER-only by default; on QuadOcc (true PAL optics) we compare all three paths.

\begin{table}[h]
\centering
\caption{A4: Projection path ablation. On H3O we use ER-only by default (native ER); on QuadOcc we compare ER-only, Raw-only, and Dual (ER+Raw).}
\label{tab:ablation_projection}
\resizebox{\columnwidth}{!}{
\begin{tabular}{l|ccc|cccccc}
\midrule
\rowcolor{gray!10}
Projection Path & \#Params & FPS $\uparrow$ & Mem $\downarrow$ & P $\uparrow$ & R $\uparrow$ & IoU $\uparrow$ & mIoU $\uparrow$ \\
\midrule
\midrule
ER-only       & 101.76M & 18.15 & 1.71GB & 65.77    & \textbf{64.81}    & 48.47    & 20.03    \\
Raw-only      & 101.97M & 25.18 & 1.55GB & 66.42    & 62.86    & 47.70    & 19.76    \\
\rowcolor{gray!10}
Dual (ER+Raw) & 189.83M & 15.46 & 2.14GB & \textbf{66.69} & 64.74 & \textbf{48.92} & \textbf{20.56} \\
\midrule
\end{tabular}}
\end{table}

\noindent\emph{Findings.}
\begin{compactitem}
\item \textbf{Accuracy on QuadOcc.} Dual achieves the best overall occupancy quality, improving over ER-only by \,+0.45\,IoU and \,+0.53\,mIoU (48.92/20.56 vs.\ 48.47/20.03), with higher precision (66.69 vs.\ 65.77) at essentially unchanged recall (64.74 vs.\ 64.81). 
Raw-only is weaker on IoU/mIoU (47.70/19.76), indicating that using the raw annulus \emph{alone} is insufficient for best voxel quality. 

\item \textbf{Cost/efficiency.} Relative to ER-only, Dual increases parameters by \,+86.5\% (189.83M vs.\ 101.76M) and peak memory by \,+25.1\% (2.14\,GB vs.\ 1.71\,GB), while throughput drops by \,14.8\% (15.46 vs.\ 18.15 FPS). 
Raw-only is the fastest (25.18 FPS; +38.7\% vs.\ ER-only) and most memory-frugal (1.55\,GB), but sacrifices recall ($-1.95$) and IoU ($-0.77$). 

\item \textbf{H3O (native ER).} When inputs are already native ER panoramas, the \emph{incremental} benefit of adding a raw path is marginal relative to its compute/memory overhead; ER-only is the practical setting.
\end{compactitem}

\noindent\textbf{Why Dual helps on QuadOcc (true PAL).}
A PAL camera forms a circular annulus on the sensor. Unwrapping to ER is convolution-friendly and keeps azimuthal continuity, but induces latitude-dependent distortion and can smooth out high-frequency structures near the inner/outer rings. 
The \emph{raw annulus} stream preserves native PAL geometry and local texture statistics; fusing it with the ER stream lets the network resolve ER-induced distortions and seam/edge artifacts while retaining the global continuity provided by ER. 
This complementary coverage improves precision (fewer false positives around thin/elongated objects) without hurting recall, thereby nudging IoU/mIoU upward.

\noindent\emph{Takeaway.}
For true PAL deployments (QuadOcc-like), Dual (ER+Raw) is a robust choice when accuracy is prioritized over a modest efficiency loss. 
For native ER inputs (H3O-like), ER-only offers the best accuracy--efficiency trade-off; enable Dual only when the last bit of precision is critical.

\begin{figure}[!t]
\centering
\includegraphics[width=1.0\linewidth]{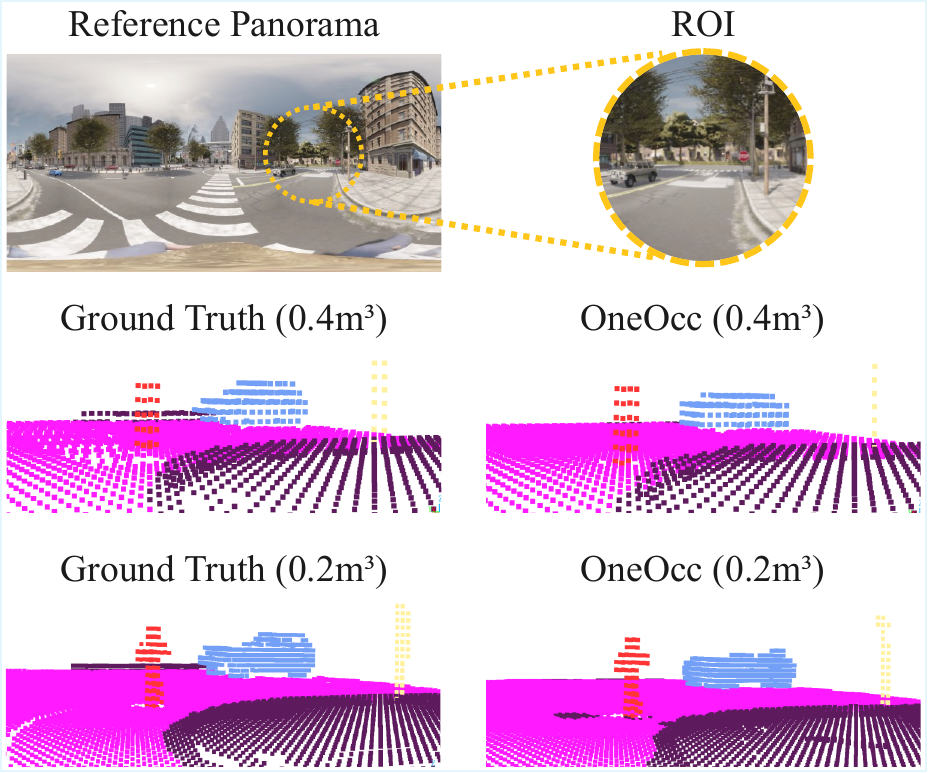}
\vskip -2ex
\caption{\textbf{Voxel grid resolution vs. throughput/accuracy on \emph{H3O-Heter}}.
We keep identical 3D bounds and voxel size; only the grid resolution changes:
\emph{BEV} increases from $64{\times}64$ to $128{\times}128$ while the number of vertical bins increases from $8$ to $16$.
Timings are measured on a single RTX\,4090 with batch size $1$ (FP32).
Raising resolution adds +3.41\,M parameters (101.76$\rightarrow$105.17\,M), reduces FPS (14.30$\rightarrow$5.01), and inflates peak memory (1.82$\rightarrow$10.71\,GB), with a concomitant drop in IoU/mIoU (49.58/32.23$\rightarrow$29.98/21.16).
}
\vskip -2ex
\label{fig:vox_res}
\end{figure}

\vspace{0.6em}
\subsection{Voxel Resolution Scaling}
\label{sec:ablation_resolution}
\begin{bluequestion}
\noindent\textit{Question.} How does performance scale when increasing the voxel grid from $64{\times}64{\times}8$ to $128{\times}128{\times}16$ under identical spatial bounds and training/inference settings?
\end{bluequestion}

\noindent\textit{Setup.}
We isolate the effect of resolution by fixing the spatial range $R$, evaluation protocol (\emph{H3O-Heter}), and all model/hyperparameters; only the voxel grid resolution is varied.
We report precision (P), recall (R), IoU, and mIoU on non-empty voxels, together with \#Params/FPS/Memory measured on a single RTX\,4090 (FP32, batch size 1).

\begin{table}[h]
\centering
\caption{A5: Resolution scaling on H3O-Heter. Same spatial bounds; only the grid resolution changes.}
\label{tab:ablation_resolution}
\resizebox{\columnwidth}{!}{
\begin{tabular}{l|lll|ccccc}
\midrule
\rowcolor{gray!10}
Resolution & \#Params & FPS $\uparrow$ & Mem $\downarrow$ & P $\uparrow$ & R $\uparrow$ & IoU $\uparrow$ & mIoU $\uparrow$ \\
\midrule
\midrule
$64{\times}64{\times}8$    & 101.76M & 14.30 & 1.82GB & 67.89 & 64.77 & 49.58 & 32.23 \\
$128{\times}128{\times}16$ & 105.17M & 5.01 & 10.71GB & 44.69 & 47.66 & 29.98 & 21.16 \\
\midrule
\end{tabular}}
\end{table}

\noindent\textit{Findings.}
\begin{compactitem} 
\item \textbf{Throughput \& footprint.}
Increasing BEV resolution by $2{\times}$ in $X$/$Y$ and doubling vertical bins ($8{\to}16$) leaves the backbone almost unchanged in parameter count (${+}3.41M$) but cuts throughput by ${\sim}3{\times}$ and raises memory by ${\sim}5.9{\times}$, as denser 3D feature maps amplify intermediate activations.

\item \textbf{Accuracy at higher resolution.}
Finer-grained occupancy is intrinsically harder: evaluating and \emph{learning} at a denser discretization exacerbates class imbalance and boundary sensitivity, and demands longer-range completion for thin, rare structures.
Empirically, our $128{\times}128{\times}16$ setting yields lower aggregate IoU/mIoU than $64{\times}64{\times}8$ despite the finer grid.
This trend aligns with multiscale SSC literature (\textit{e.g.}, LMSCNet~\cite{roldao2020lmscnet}), where coarse-scale heads are designed for efficiency and are consistently easier to optimize/infer than their fine-scale counterparts\footnote{See LMSCNet’s multiscale completion design and coarser-head efficiency.}.

\item \textbf{Embodied perspective.}
Our target use case is \emph{embodied intelligence} with tight on-board compute and memory budgets.
Given the ${\sim}3{\times}$ FPS gain and ${\sim}6{\times}$ lower memory at $64{\times}64{\times}8$, we adopt this resolution as the \emph{main} setting in the paper: it strikes the best accuracy–efficiency trade-off for real-time, resource-constrained platforms, while $128{\times}128{\times}16$ is reserved for analyses prioritizing geometric detail with a substantial compute penalty.
\end{compactitem}

Therefore, under fixed bounds, $64{\times}64{\times}8$ is the preferred default for accuracy–efficiency in embodied settings; \textbf{$128{\times}128{\times}16$} offers crisper geometry on slender structures but at markedly worse throughput/memory and lower overall IoU/mIoU.

\vspace{0.6em}
\subsection{Calibration Robustness}
\label{sec:ablation_calibration}
\begin{bluequestion}
\noindent\textit{Question.} How robust is OneOcc to test-time intrinsic/extrinsic calibration perturbations, and can simple calibration-noise augmentation flatten the degradation curve?
\end{bluequestion}

\noindent\textit{Setup.}
We evaluate robustness on \emph{H3O-Heter} by injecting controlled calibration noise at \emph{data loading time} in the equirectangular projection.
Intrinsic perturbation is implemented as a global pixel scaling, while extrinsic perturbation is simulated as per-point pixel offsets.
For reproducibility, the perturbation is deterministic for each frame via a fixed seed and frame id, with clipping and FoV-mask update applied accordingly.
We report mIoU under three settings: joint intrinsic+extrinsic noise, intrinsic-only noise, and extrinsic-only noise, each with perturbation ratios $p \in \{1\%, 2\%, 5\%\}$.
We additionally include robustness-oriented variants retrained with $5\%$ joint calibration-noise augmentation.

\begin{table}[h]
\centering
\caption{A6: Robustness to intrinsic/extrinsic calibration noise on H3O-Heter. We report mIoU under clean and noisy projections.}
\label{tab:ablation_calibration}
\resizebox{\columnwidth}{!}{
\begin{tabular}{l|c|ccc|ccc|ccc}
\midrule
\rowcolor{gray!10}
 & Clean & \multicolumn{3}{c|}{Both (Intr.+Extr.)} & \multicolumn{3}{c|}{Intr. only} & \multicolumn{3}{c}{Extr. only} \\
\rowcolor{gray!10}
Method & 0\% & 1\% & 2\% & 5\% & 1\% & 2\% & 5\% & 1\% & 2\% & 5\% \\
\midrule
\midrule
MonoScene & 24.15 & 21.99 & 19.58 & 13.96 & 22.78 & 21.00 & 16.30 & 22.75 & 20.69 & 15.90 \\
OneOcc    & 32.23 & 27.26 & 23.67 & 16.75 & 27.88 & 24.80 & 19.15 & 29.42 & 25.28 & 18.95 \\
\midrule
MonoScene$^{\dagger}$ & 18.97 & 18.86 & 18.64 & 16.48 & 18.71 & 18.28 & 15.74 & 19.01 & 19.17 & 18.92 \\
OneOcc$^{\dagger}$    & 23.00 & 22.97 & 23.03 & 22.84 & 22.86 & 22.84 & 22.48 & 22.94 & 23.15 & 23.16 \\
\midrule
\end{tabular}}
{\scriptsize\noindent\textit{$^{\dagger}$ Retrained with $5\%$ joint intrinsic+extrinsic calibration-noise augmentation.}\par}
\end{table}

\noindent\textit{Findings.}
\begin{compactitem}
\item \textbf{Robustness under calibration perturbation.}
Under all corruption levels, OneOcc consistently remains above MonoScene.
With joint intrinsic+extrinsic noise, OneOcc achieves 27.26/23.67/16.75 mIoU at 1\%/2\%/5\%, outperforming MonoScene by +5.27/+4.09/+2.79, respectively.
This indicates that the proposed dual-projection lifting and 3D reasoning pipeline retains stronger geometric consistency under imperfect calibration.

\item \textbf{Intrinsic vs.\ extrinsic noise.}
Both perturbation types degrade performance, while their combination is the most harmful as expected.
Intrinsic-only noise is slightly more disruptive at low-to-moderate levels, likely because global projection-scale distortion affects all lifted features simultaneously, whereas extrinsic perturbation mainly introduces spatial misalignment.
Nevertheless, OneOcc preserves a clear margin over the baseline in all settings.

\item \textbf{Effect of calibration-noise augmentation.}
Retraining with simple $5\%$ calibration-noise augmentation substantially flattens the degradation curve.
For example, the robustness-oriented OneOcc$^{\dagger}$ variant stays nearly constant at 22.97/23.03/22.84 mIoU under 1\%/2\%/5\% joint noise.
This comes at the cost of lower clean performance (23.00 vs.\ 32.23), suggesting a standard robustness--accuracy trade-off:
the default model is preferable when calibration is reliable, whereas noise augmentation is attractive for long-term deployment under persistent drift.
\end{compactitem}

Therefore, OneOcc is reasonably robust to bounded calibration errors and consistently more resilient than MonoScene; when stronger calibration drift is expected, simple calibration-noise augmentation can further improve robustness at the expense of clean-set accuracy.

\vspace{0.6em}
\subsection{Temporal Aggregation}
\label{sec:ablation_temporal}
\begin{bluequestion}
\noindent\textit{Question.} Does lightweight temporal aggregation materially improve occupancy prediction, and how does it trade accuracy against latency and memory compared with the default single-frame design?
\end{bluequestion}

\noindent\textit{Setup.}
We compare the default \emph{single-frame} OneOcc with two simple 3-frame temporal variants under identical image resolution ($608{\times}1216$), voxel bounds, backbone, and evaluation protocol.
The first variant performs temporal feature averaging using \emph{ground-truth pose} for alignment, representing a minimal upper-bound baseline with reliable multi-frame registration.
The second variant adopts a lightweight BEVFormer-style temporal attention over three frames.
We report mIoU on \emph{QuadOcc-val} and \emph{H3O-Heter}, together with latency and memory measured on a single RTX\,4090 (FP32, batch size 1).

\begin{table}[h]
\centering
\caption{A7: Temporal aggregation vs.\ the default single-frame design. Temporal fusion improves accuracy, but also increases deployment cost.}
\label{tab:ablation_temporal}
\resizebox{\columnwidth}{!}{
\begin{tabular}{l|cc|cc}
\midrule
\rowcolor{gray!10}
Setting & QuadOcc-val mIoU $\uparrow$ & H3O-Heter mIoU $\uparrow$ & Latency (ms) $\downarrow$ & Memory (GB) $\downarrow$ \\
\midrule
\midrule
OneOcc (single-frame) & 20.56 & 32.23 & 69.93 & 1.82 \\
+ Temporal average by GT pose (3 frames) & 20.92 & 33.74 & 69.93 & 1.82 \\
+ Temporal BEVFormer-like attention (3 frames) & 21.18 & 34.25 & 78.60 & 2.35 \\
\midrule
\end{tabular}}
\end{table}

\noindent\textit{Findings.}
\begin{compactitem}
\item \textbf{Temporal aggregation is beneficial with reliable alignment.}
Both 3-frame variants improve over the default single-frame setting.
Simple temporal averaging already raises mIoU from 20.56 to 20.92 on QuadOcc-val and from 32.23 to 33.74 on H3O-Heter, confirming that multi-frame context helps suppress transient gait-induced jitter and recover partially occluded structures when alignment is accurate.

\item \textbf{Stronger temporal fusion yields larger gains but higher cost.}
The BEVFormer-style temporal attention further improves mIoU to 21.18 on QuadOcc-val and 34.25 on H3O-Heter, but increases latency from 69.93\,ms to 78.60\,ms and memory from 1.82\,GB to 2.35\,GB.
This shows that temporal fusion is effective, yet its benefit is not free: better multi-frame reasoning comes with additional state handling, feature fusion overhead, and higher deployment cost.

\item \textbf{Why single-frame remains the default.}
Our goal is an \emph{easy-to-deploy}, \emph{low-latency}, and \emph{odometry-free} embodied perception system.
The temporal-average variant relies on ground-truth pose for reliable frame alignment and therefore should be interpreted as an upper-bound reference rather than a plug-and-play deployment mode.
In contrast, the default single-frame OneOcc avoids temporal drift accumulation, requires no motion history, and maintains a cleaner inference path for resource-constrained onboard platforms.
\end{compactitem}

Therefore, temporal aggregation can further improve panoramic occupancy prediction when accurate multi-frame alignment is available, but the default {single-frame OneOcc} remains the preferred setting for real-time embodied deployment due to its simplicity, robustness to drift, and lower system complexity.

\begin{figure*}[!t]
\centering
\includegraphics[width=1.0\textwidth]{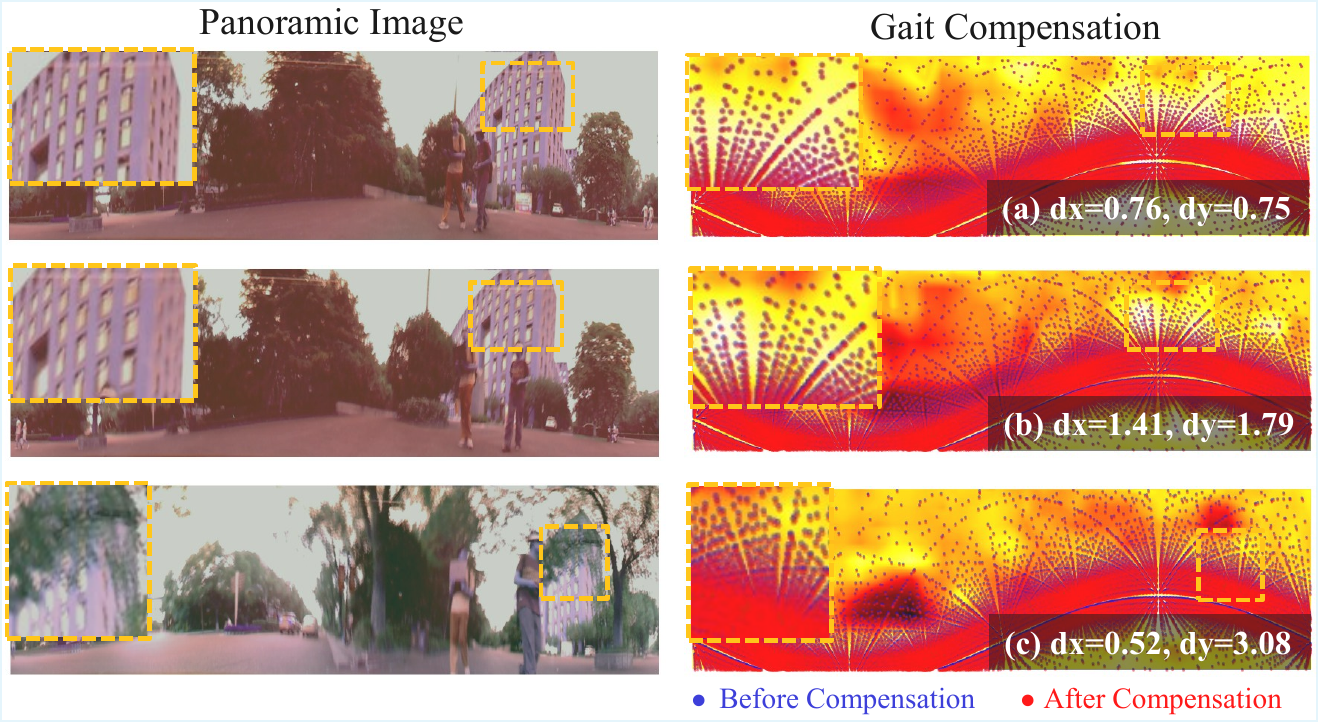}
\vskip -2ex
\caption{\textbf{Panoramic gait displacement compensation on real-world quadruped data.}
  We visualize the effect of Gait Displacement Compensation (GDC) on our QuadOcc dataset collected by a quadruped robot equipped with a panoramic camera.
  Each row shows one panorama captured at a different gait intensity: the left column is the input image with a zoom-in of a distant structure, and the right column overlays FLoSP~\cite{cao2022monoscene} sampling locations on the equirectangular feature map before (blue) and after (red) GDC.
  From top to bottom, the robot motion changes from mild to pronounced body oscillation, and the estimated vertical offsets $dy$ increase accordingly (see the $(dx,dy)$ values in (a)--(c)), while the horizontal offsets $dx$ remain comparatively small.
  This indicates that gait-induced jitter produces a characteristic \emph{vertical} motion-blur pattern during exposure, and GDC automatically adapts the strength of its correction mainly along the vertical axis.
  Although the predicted displacements are real-valued and sometimes fractional, we augmented FLoSP with bilinear interpolation, so sub-pixel shifts of the sampling grid are still meaningful and help recover sharper, less blur-contaminated features for downstream occupancy prediction in this single-frame model.}
\label{fig:gait}
\end{figure*}

\section{Qualitative Analysis of GDC}
\label{sec:qualitative_analysis_gdc}
Figure~\ref{fig:gait} provides a qualitative view of GDC on real-world quadruped runs.
The three rows correspond to panoramas captured under increasing gait intensity, ranging from mild to pronounced body oscillation.
Because the panoramic camera is mounted on a moving body, the gait induces vertical shake during the exposure time and creates a characteristic motion-blur pattern: distant objects and tree lines are smeared mainly along the vertical direction.
Without compensation, the fixed FLoSP~\cite{cao2022monoscene} sampling grid (blue) may query different parts of this blurred streak depending on the current gait phase, effectively mixing background and foreground evidence within a single frame.
GDC predicts per-frame offsets that correct this effect: from (a) to (c), the estimated $dy$ increases while $dx$ stays small, showing that the module automatically scales its response with the blur magnitude and focuses almost exclusively on the dominant vertical direction.
Note that OneOcc operates purely on single panoramas without temporal aggregation, so GDC is not used for temporal alignment; instead, it refines the instantaneous feature sampling locations to better match the underlying static geometry under gait-induced motion blur.
Since we implemented FLoSP with bilinear interpolation rather than the classic nearest-neighbor lookup~\cite{cao2022monoscene}, the real-valued (often fractional) $(dx,dy)$ still produce meaningful sub-pixel shifts of the sampling grid, which leads to cleaner voxel features and the consistent mIoU gains observed in our ablations.

\section{Lightweight Design Philosophy of OneOcc}
\label{sec:supp-params}

While Sec.~4.5 of the main paper reports the overall runtime
and memory footprint, here we decompose the parameter
counts of OneOcc and the MonoScene~\cite{cao2022monoscene} baseline into their
major components. We focus on the panoramic setting on
QuadOcc as well as the Human360Occ (H3O) variant.

\subsection{QuadOcc: dual-projection variant}
Tab.~\ref{tab:supp-params-quad} summarizes the parameter
breakdown when training on QuadOcc with the full
dual-projection encoders (DP--ER) and bi-grid voxelization.
MonoScene~\cite{cao2022monoscene} consists of a single 2D encoder operating on the
equirectangular panorama and a standard 3D UNet decoder,
resulting in $132\mathrm{M}$ parameters for the 2D encoder and
$16.9\mathrm{M}$ for the 3D decoder ($148.9\mathrm{M}$ total).

In contrast, OneOcc factorizes its capacity across (i) two
2D encoders for raw and equirectangular views (DP--ER),
(ii) a lightweight depthwise-separable 3D decoder with
Hierarchical AMoE-3D, and (iii) a set of geometry-aware
auxiliary modules (GDC, volumetric fusion, and tiny gating
heads). Each 2D encoder in DP--ER contains $87.8\mathrm{M}$ parameters, and the
DWLite3D decoder with AMoE-3D attention blocks (excluding the gating
heads) contains $12.3\mathrm{M}$. The remaining components---GDC heads,
3D SE/fusion layers, and the lightweight AMoE-3D gating MLPs---add only
about $0.27\mathrm{M}$ parameters in total (less than $0.2\%$ of the
overall capacity). For clarity, we count the AMoE-3D attention and expert
weights inside the ``3D decoder'' bucket, while the gating networks are
included in the ``GDC+fusion+gates'' bucket in
Tab.~\ref{tab:supp-params-quad} and Tab.~\ref{tab:supp-params-h3o}.

Although the total parameter count increases by
$\approx 26\%$ compared to MonoScene
($148.9\mathrm{M} \rightarrow 189.8\mathrm{M}$), the design is
\emph{structurally lightweight}: (1) the 3D decoder is
$27\%$ smaller ($16.9\mathrm{M} \rightarrow 12.3\mathrm{M}$) thanks
to the depthwise-separable DWLite3D design; (2) each single
DP--ER encoder is $33\%$ lighter than MonoScene's 2D
encoder ($132\mathrm{M} \rightarrow 87.8\mathrm{M}$); and (3)
the additional geometry-aware modules (GDC, BGV-based
fusion, AMoE-3D gates) are extremely cheap in parameters.
The modest increase in total parameters on QuadOcc therefore
comes almost entirely from using two lighter encoders in
parallel, which is amortized by GPU parallelism and justified
by the improved robustness under gait-induced jitter.

\begin{table}[t]
    \centering
    \small
    \setlength{\tabcolsep}{5pt}
     \caption{Parameter breakdown on QuadOcc (panorama-only). The additional geometry-aware modules in OneOcc (GDC, BGV-informed volumetric fusion, and the lightweight AMoE-3D gating heads) contribute less than $0.2\%$ of the total parameters. The AMoE-3D attention and expert weights are counted as part of the 3D decoder. Both the 3D decoder and each single 2D encoder are lighter than the MonoScene~\cite{cao2022monoscene} counterpart.}

    \resizebox{\linewidth}{!}{%
    \begin{tabular}{lcccc}
        \midrule
        \rowcolor{gray!10}
        \textbf{Method} & \textbf{2D encoder(s)} & \textbf{3D decoder} & \textbf{GDC+fusion+gates} & \textbf{Total} \\
        \midrule
        \midrule
        MonoScene (QuadOcc) &
        $1\times 132\mathrm{M}$ & $16.9\mathrm{M}$ & -- &
        $148.9\mathrm{M}$ \\
        \rowcolor{gray!10}
        OneOcc (QuadOcc) &
        $2\times 87.8\mathrm{M}$ & $12.3\mathrm{M}$ & $\approx 0.27\mathrm{M}$ &
        $189.8\mathrm{M}$ \\
        \midrule
    \end{tabular}%
    }
    \label{tab:supp-params-quad}
\end{table}

\subsection{H3O: single-projection variant}
On Human360Occ (H3O), the images are already native
equirectangular, so we follow the main paper and disable the
raw-annulus branch in DP--ER and the associated fusion
layers, while keeping the same DWLite3D decoder and GDC.
Tab.~\ref{tab:supp-params-h3o} reports the resulting
parameter statistics. MonoScene~\cite{cao2022monoscene} again uses a single 2D
encoder ($132\mathrm{M}$) and the standard 3D decoder
($16.9\mathrm{M}$), totaling $148.9\mathrm{M}$ parameters.

The H3O configuration of OneOcc uses a single 2D encoder
($87.8\mathrm{M}$), the same $12.3\mathrm{M}$ DWLite3D decoder,
and the small GDC and gating modules, for a total of
${\sim}100\mathrm{M}$ parameters. 
This corresponds to a reduction of
approximately $33\%$ in parameter count compared to
MonoScene~\cite{cao2022monoscene} ($148.9\mathrm{M} \rightarrow 101.7\mathrm{M}$), while retaining all the geometry-aware components that are shown in the main paper ablations to be crucial for far-field context and near-field contact geometry.

\begin{table}[t]
    \centering
    \small
    \setlength{\tabcolsep}{5pt}
    \caption{Parameter breakdown on H3O. When the raw panoramic branch is disabled, OneOcc becomes markedly lighter than MonoScene~\cite{cao2022monoscene} in terms of learned parameters, while preserving GDC, the lightweight AMoE-3D gating heads, and the DWLite3D attention blocks (whose weights are counted inside the 3D decoder).}

    \resizebox{\linewidth}{!}{%
    \begin{tabular}{lcccc}
        \midrule
        \rowcolor{gray!10}
        \textbf{Method} & \textbf{2D encoder(s)} & \textbf{3D decoder} & \textbf{GDC+gates} & \textbf{Total} \\
        \midrule
        \midrule
        MonoScene (H3O) &
        $1\times 132\mathrm{M}$ & $16.9\mathrm{M}$ & -- &
        $148.9\mathrm{M}$ \\
        \rowcolor{gray!10}
        OneOcc (H3O) &
        $1\times 87.8\mathrm{M}$ & $12.3\mathrm{M}$ & $\approx 0.27\mathrm{M}$ &
        $101.7\mathrm{M}$ \\
        \midrule
    \end{tabular}%
    }
    \label{tab:supp-params-h3o}
\end{table}

\subsection{Runtime on NVIDIA Jetson AGX Orin}
\label{sec:supp-orin-runtime}

To complement the parameter analysis above, we further report
\emph{measured} runtime on an NVIDIA Jetson AGX Orin 64GB
(MAX power mode), which provides a more deployment-realistic
efficiency reference than desktop-GPU throughput alone.
In particular, we compare OneOcc against MonoScene~\cite{cao2022monoscene}
under the QuadOcc panoramic setting using the native input size
$1{\times}3{\times}370{\times}1220$ and batch size $1$.
All numbers are averaged over $100$ runs after $5$ warm-up iterations.
Following a practical edge-deployment setup, the 2D UNet-style
image encoder/decoder is executed in INT8 using TensorRT,
while the remaining geometry-aware modules are run in FP16.
We decompose the latency into four stages:
(i) \emph{2D Enc. + 2D Dec.}, \textit{i.e.}, the image-space feature extraction
and decoding backbone;
(ii) \emph{Lift2Cart Samp.}, which lifts image features into the Cartesian voxel space;
(iii) \emph{Cart2Polar Samp.}, the additional Cartesian-to-polar resampling used only in OneOcc;
and (iv) \emph{3D Dec.}, the volumetric decoder operating on lifted features.

\begin{table}[t]
    \centering
    \small
    \setlength{\tabcolsep}{5.2pt}
    \caption{Measured runtime breakdown on NVIDIA Jetson AGX Orin 64GB (MAX power mode).
    Input: $1{\times}3{\times}370{\times}1220$ (QuadOcc panoramas), batch size $1$;
    $5$ warm-up iterations + $100$ measured runs.
    The 2D image encoder/decoder is executed in INT8 (TensorRT), while the remaining modules are run in FP16.
    Despite the extra Cartesian-to-polar resampling step, OneOcc remains faster overall due to its lighter 2D/3D computation.}
    \resizebox{\linewidth}{!}{%
    \begin{tabular}{l|cccc|cc}
        \midrule
        \rowcolor{gray!10}
        \textbf{Method} & \textbf{2D Enc. + 2D Dec.} & \textbf{Lift2Cart Samp.} & \textbf{Cart2Polar Samp.} & \textbf{3D Dec.} & \textbf{Total} & \textbf{FPS} \\
        \midrule
        \midrule
        MonoScene & 61.97 & 2.46 & \textit{n.a.} & 33.58 & 98.01 & 10.20 \\
        \rowcolor{gray!10}
        OneOcc    & 53.86 & 2.48 & 2.40 & 14.57 & 73.32 & 13.64 \\
        \midrule
    \end{tabular}%
    }
    \label{tab:orin_runtime}
\end{table}

As shown in Tab.~\ref{tab:orin_runtime}, OneOcc achieves
a total latency of $73.32$\,ms ($13.64$ FPS), compared with
$98.01$\,ms ($10.20$ FPS) for MonoScene~\cite{cao2022monoscene},
corresponding to an overall speedup of approximately $1.34{\times}$
on the embedded platform.
Notably, this gain is achieved \emph{despite} the additional
\emph{Cart2Polar Samp.} stage, which is absent in MonoScene.
The main reason is that OneOcc reduces computation in the
dominant stages.
In the 2D branch, OneOcc requires $53.86$\,ms versus
$61.97$\,ms for MonoScene, reflecting the lighter
dual-projection image backbone design discussed above.
The largest reduction appears in the 3D decoder:
OneOcc uses only $14.57$\,ms, whereas MonoScene requires
$33.58$\,ms.
This is consistent with the parameter analysis in
Tab.~\ref{tab:supp-params-quad}--\ref{tab:supp-params-h3o},
where the DWLite3D decoder is substantially lighter than the
standard dense 3D UNet decoder.
By contrast, the geometry-aware sampling operations are cheap:
\emph{Lift2Cart Samp.} is nearly identical for both methods
($2.48$\,ms vs.\ $2.46$\,ms), and the extra
\emph{Cart2Polar Samp.} in OneOcc adds only $2.40$\,ms.
These measurements support the lightweight positioning of OneOcc
from an edge-deployment perspective.
Rather than relying only on desktop-GPU throughput or theoretical
scaling estimates, the Jetson AGX Orin results show that the
proposed design remains practically efficient on an embedded
robotics platform while providing stronger panoramic semantic
occupancy prediction.

\subsection{Discussion}
Across both benchmarks, OneOcc follows a consistent
lightweight design philosophy: (i) concentrate capacity in
flexible 2D encoders that can be reconfigured (dual-projection
on QuadOcc vs.\ single-projection on H3O), (ii) employ a
depthwise-separable DWLite3D decoder with Hierarchical
AMoE-3D to reduce 3D convolutional cost, and (iii) ensure that
geometry-specific modules (GDC, BGV-based volumetric fusion,
and the AMoE-3D gating heads that select experts based on
GradEnergy) remain extremely parameter-efficient.
On QuadOcc, this yields a modest increase in parameters
but significantly better panoramic SSC under gait-induced
jitter; on H3O, the same architecture, with the redundant
projection path removed, becomes \emph{substantially} lighter
than MonoScene~\cite{cao2022monoscene} while still providing clear performance gains.

\begin{figure*}[t]
  \centering
  \tikzset{
    block/.style={draw, rounded corners, very thick, align=center,
                  inner sep=2pt, minimum width=3.05cm, minimum height=1.2cm},
    arr/.style={-Latex, very thick},
    tinyanno/.style={font=\scriptsize, inner sep=1pt}
  }
  \begin{tikzpicture}[node distance=0.9cm and 0.6cm]
    \node[block] (sens) {Sensor Suite \& Sync\\{\footnotesize Livox Mid-360 + PAL}\\{\scriptsize HW trigger + SW ts}};
    \node[block, right=of sens] (ocam) {OCam Unwarping\\{\footnotesize $I^{\mathrm{raw}}\!\to\! I^{\mathrm{equi}}$}\\{\scriptsize Taylor/OCam~\cite{scaramuzza2006toolbox}}};
    \node[block, right=of ocam] (seg) {Open-Vocabulary Seg.\\{\footnotesize Grounded-SAM}\\{\scriptsize $S_t,\ Q_t$}};
    \node[block, right=of seg] (proj) {LiDAR--Image Label Transfer\\{\footnotesize $\mathbf{u}=\pi(\mathbf{T}_{L\to C}\tilde{\mathbf{p}}_L)$}\\{\scriptsize $\ell(\mathbf{p}),\ w(\mathbf{p}){=}Q_t(\mathbf{u})$}};

    \node[block, below=1.2cm of sens] (dyn) {Semantic-Guided Dynamic Mapping\\{\footnotesize LiDAR-SLAM, tubes $\{\Gamma_k\}$}\\{\scriptsize boxes $\{\mathcal{B}_{k,t}\}$}};
    \node[block, right=of dyn] (vox) {Voxelization \& Voting\\{\footnotesize grid $\mathbf{V}:64{\times}64{\times}8@0.4$m}\\{\scriptsize Eq.~\ref{eq:vote} $\to L_i,\ \mathrm{conf}(V_i)$}};
    \node[block, right=of vox] (temp) {Temporal Aggregation\\{\footnotesize window $[t\!-\!n,\,t\!+\!n]$}\\{\scriptsize Eqs.~\ref{eq:temporal-transform}, \ref{eq:temporal-vote}}};
    \node[block, right=of temp] (mode) {Gated 3D Mode\\{\footnotesize mask $\mathcal{M}_q$ (Eq.~\ref{eq:quant-mask})}\\{\scriptsize $H_c(\mathbf{x})$, $\mathbf{V}_{\mathrm{ref}}[\mathbf{x}]$}};

    \draw[arr] (sens) -- (ocam);
    \draw[arr] (ocam) -- (seg);
    \draw[arr] (seg) -- (proj);

    \draw[arr] (proj.south) |- ++(0,-0.6) -| (dyn.north);

    \draw[arr] (dyn) -- (vox);
    \draw[arr] (vox) -- (temp);
    \draw[arr] (temp) -- (mode);

    \node[tinyanno, above=0.05cm of sens] {Real robot, time-synced LiDAR \& PAL};
    \node[tinyanno, above=0.05cm of mode] {Outputs: $\mathbf{V}_{\mathrm{ref}}$, conf, splits/resolution};
  \end{tikzpicture}
  \caption{\textbf{QuadOcc ground-truth construction pipeline.}
  From synchronized LiDAR \& PAL capture, OCam unwarping and open-vocabulary segmentation provide $S_t$ and $Q_t$.
  LiDAR--image label transfer assigns $(\ell,w)$ to points, followed by semantic-guided dynamic mapping.
  Voxelization and majority voting (Eq.~\ref{eq:vote}) produce $\mathbf{V}$; temporal aggregation
  (Eqs.~\ref{eq:temporal-transform}, \ref{eq:temporal-vote}) and a gated 3D mode with $\mathcal{M}_q$ (Eq.~\ref{eq:quant-mask})
  yield the refined volume $\mathbf{V}_{\mathrm{ref}}$.}
  \label{fig:quadocc_pipeline}
\end{figure*}

\section{QuadOcc: Dataset Construction}
\label{sec:supp-quadocc}

\paragraph{Goal and setting}
\textbf{QuadOcc} targets full-surround \emph{semantic occupancy} from a \emph{single} omnidirectional panorama on a quadruped robot.
Each frame couples an omnidirectional image captured via Panoramic Annular Lens (PAL)~\cite{gao2022review} with time-synchronized LiDAR for semi-automatic Ground Truth (GT) construction; outputs are $64{\times}64{\times}8$ semantic volumes at $0.4$\,m per voxel under a $6$-class taxonomy.
The dataset comprises $10$ scenes and $24K$ frames across day/dusk/night, reflecting realistic legged operation with moderate speeds and tight payload/power budgets.

\subsection{Sensor Suite, Time Sync, and Calibration}
We mount on a quadruped a Livox Mid-360 LiDAR (with built-in IMU) and an omnidirectional panoramic annular lens (PAL) camera, time-synchronized by hardware trigger and software timestamp alignment. Let $\mathcal{I} = \{I_t\}$ be panoramic RGB streams and $\mathcal{L}=\{\mathcal{P}_t\}$ be LiDAR scans.

\paragraph{PAL (Taylor/OCam) calibration and unwarping.}
We adopt the generic Taylor (OCam) model~\cite{scaramuzza2006toolbox} for PAL calibration. For an equirectangular pixel $(u,v)\in [0,W)\times[0,H)$ with spherical angles:
\begin{equation}
\phi = \tfrac{2\pi}{W}u - \pi,\quad \theta = \tfrac{\pi}{2} - \tfrac{\pi}{H}v,
\end{equation}
the corresponding polar radius on the annulus is $r(\theta)=\sum_{i=0}^{N} a_i \theta^i$, and the raw-annulus coordinates are:
\begin{equation}
\begin{bmatrix} u_{\mathrm{raw}} \\ v_{\mathrm{raw}} \end{bmatrix}
=
\begin{bmatrix} u_0 \\ v_0 \end{bmatrix}
+ A \, r(\theta) \begin{bmatrix} \cos\phi \\ \sin\phi \end{bmatrix},
\end{equation}
where $\{a_i\}$, $(u_0,v_0)$, and $A\!\in\!\mathbb{R}^{2\times 2}$ come from calibration. The unwarped equirectangular image $I^{\mathrm{equi}}$ is obtained by sampling the raw annulus $I^{\mathrm{raw}}$ at $(u_{\mathrm{raw}},v_{\mathrm{raw}})$.

\begin{figure*}[!t]
\centering
\includegraphics[width=1.0\linewidth]{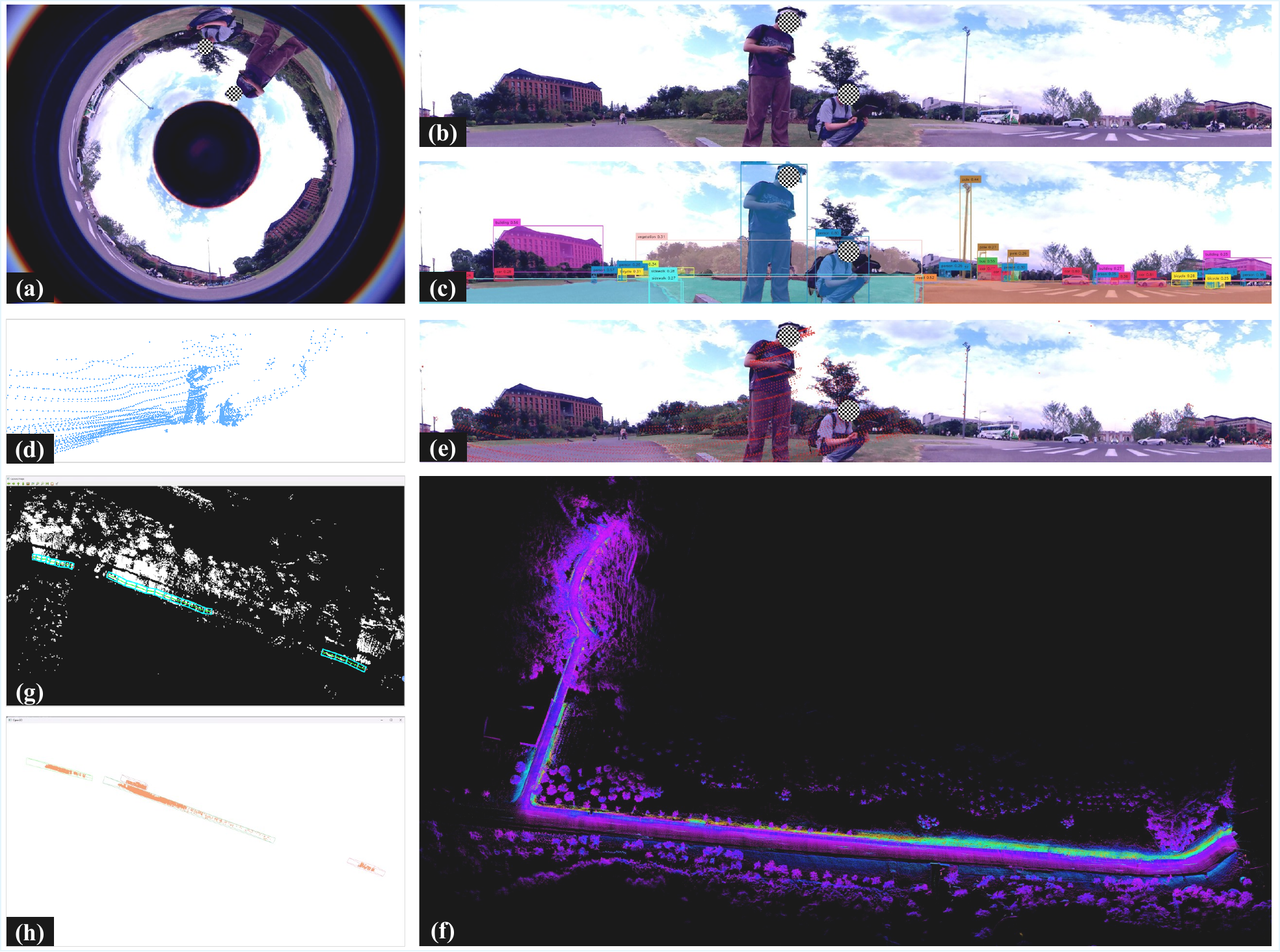}
\caption{\textbf{QuadOcc ground-truth construction pipeline.}
(a) Raw panoramic annulus captured by the PAL camera. 
(b) Equirectangular (ER) panorama obtained by \emph{calibrated} OCam/Taylor~\cite{scaramuzza2006toolbox} unwarping of (a). 
(c) Open-vocabulary segmentation on the ER image to obtain per-pixel class scores (\textit{e.g.}, Grounded-SAM~\cite{ren2024grounded}), yielding $\mathcal{S}_t$ and score maps $\mathbf{Q}_t$. 
(d) A single time-synchronized LiDAR scan $\mathbf{P}_t$. 
(e) LiDAR$\!\to$camera projection followed by \emph{image$\!\to$point} label transfer: pixels at $u=\pi\!\big(T_{L\rightarrow C}\tilde{\mathbf{p}}_L\big)$ assign $(\ell, w)=\mathbf{Q}_t(u)$ to 3D points, producing per-point semantics and confidence. 
(f) LiDAR SLAM~\cite{xu2022fast} mapping for long-range geometry and temporal consistency. 
(g) Human-in-the-loop BEV annotation of moving-object trajectories to form tubes/boxes $\{\Gamma_k\},\{B_{k,t}\}$ for dynamic/static separation. 
(h) Points are associated with trajectory boxes and motion attributes (direction, instance), then voxelized with majority voting and temporal aggregation to yield the refined semantic occupancy volume $\mathbf{V}_{\text{ref}}$.
This pipeline produces $64{\times}64{\times}8$ semantic volumes at $0.4$\,m resolution under a 6-class taxonomy and is used as GT for training/evaluation on \textit{QuadOcc}.}
\label{fig:quad_label}
\end{figure*}

\subsection{Panoramic Semantic Acquisition}
We perform pixel-wise open-vocabulary segmentation on $I^{\mathrm{equi}}_t$ with Grounded-SAM~\cite{ren2024grounded} using a prompt set $\mathcal{P}$ (\textit{e.g.}, ``road'', ``vehicle'', ``vegetation''), producing a semantic map $S_t \in \{1,\dots,C\}^{H\times W}$ with confidence map $Q_t$.

\subsection{LiDAR--Image Label Transfer}
With extrinsics $\mathbf{T}_{L\rightarrow C}\!\in\!\mathrm{SE(3)}$ and the PAL intrinsics, a LiDAR point $\mathbf{p}_L\!\in\!\mathbb{R}^3$ projects to pixel $\mathbf{u}\!=\!\pi(\mathbf{T}_{L\rightarrow C}\,\tilde{\mathbf{p}}_L)$, where $\tilde{\mathbf{p}}_L$ is in homogeneous coordinates and $\pi(\cdot)$ denotes the PAL projection composed with the equirectangular sampling above. We assign the point label:
\begin{equation}
\ell(\mathbf{p}_L) = S_t(\mathbf{u}), \qquad w(\mathbf{p}_L)=Q_t(\mathbf{u}),
\end{equation}
thus converting costly 3D annotation into robust 2D segmentation.

\subsection{Semantic-Guided Dynamic Mapping}
We split points into \emph{static} vs. \emph{dynamic} by semantics $\ell(\cdot)$ (\textit{e.g.}, ``vehicle'', ``pedestrian'' as dynamic). A LiDAR-SLAM~\cite{xu2022fast} backend yields a dense global map $\mathcal{M}$ in a world frame. Dynamic instances form \emph{tubes} $\{\Gamma_k\}$ in space-time; each tube provides a motion axis and per-frame bounding boxes $\mathcal{B}_{k,t}$. Background is integrated as static surfels/voxels; dynamic objects are re-inserted in temporal order to preserve their trajectories with minimal manual interaction.

\subsection{Voxelization and Majority Voting}
We use a fixed occupancy grid $\mathbf{V}\in\{0,\dots,C\}^{X\times Y\times Z}$ (default $64\!\times\!64\!\times\!8$ with $0.4$\,m voxels).
Given all points $\{P_{i,j}\}$ in voxel $V_i$, we take a weighted majority vote:
\begin{equation}
\label{eq:vote}
\begin{aligned}
L_i &= \arg\max_{c}\; \sum_{j} w(P_{i,j})\,\delta\!\big(\ell(P_{i,j}),\,c\big), \\
\mathrm{conf}(V_i) &= \frac{\max\limits_{c}\, \sum_{j} \delta\!\big(\ell(P_{i,j}),c\big)}{\sum_{j} 1}.
\end{aligned}
\end{equation}
where $\delta(\cdot,\cdot)$ is the Kronecker delta and $\mathrm{conf}(\cdot)$ is a per-voxel confidence (later used for quality control).

\begin{figure*}[!t]
\centering
\includegraphics[width=1.0\linewidth]{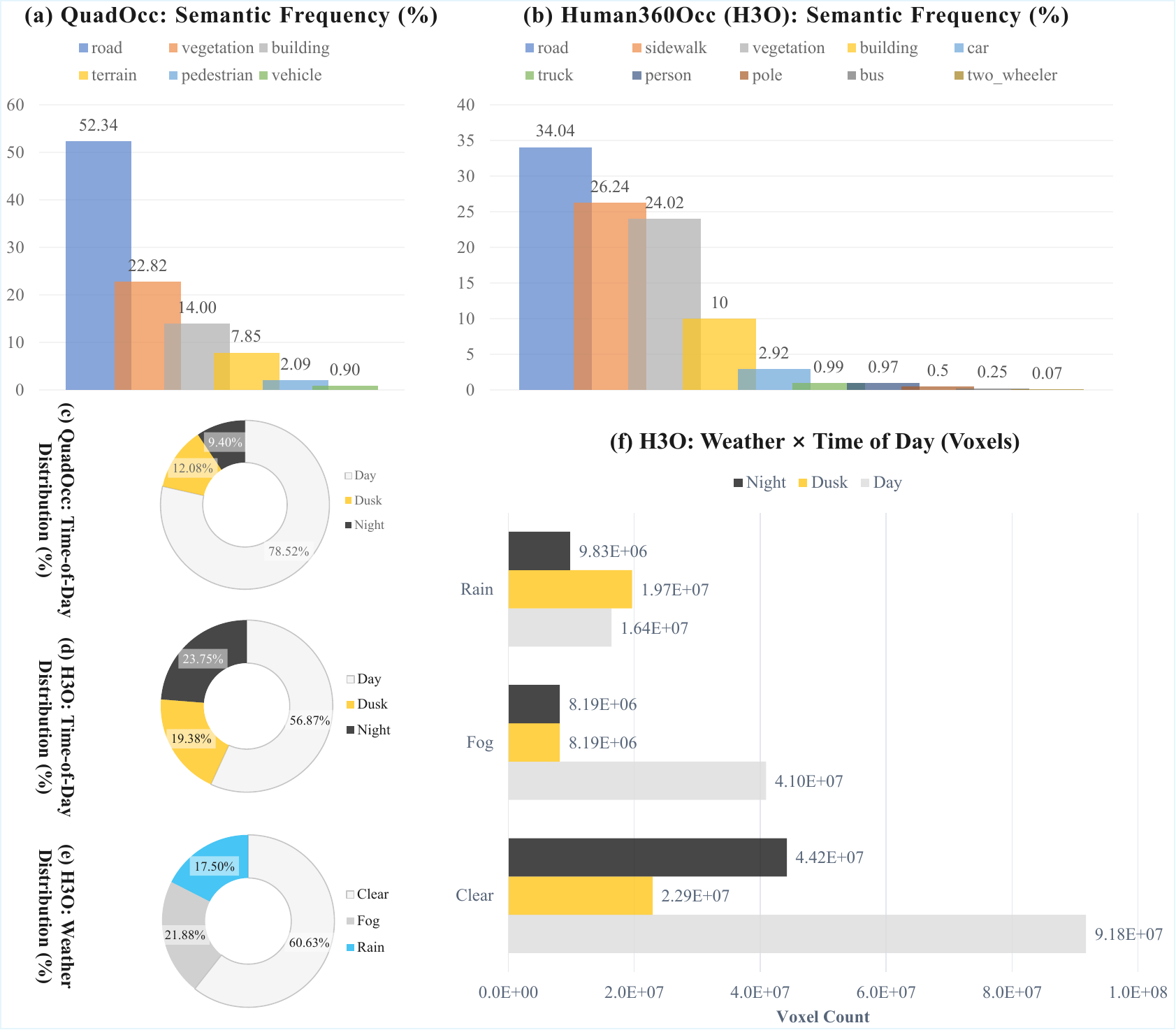}
\vskip -1ex
\caption{\textbf{Dataset statistics for QuadOcc and Human360Occ (H3O).}
(a) QuadOcc semantic frequency (non-empty voxels): strong head classes (\textit{e.g.}, \emph{road} $\approx$ $52.34\%$) with a long tail (\emph{pedestrian}, \emph{vehicle}).
(b) H3O semantic frequency (non-empty voxels): most voxels lie in \emph{road/sidewalk/vegetation/building}, with smaller shares for traffic-related rare classes.
(c) QuadOcc time-of-day split by frames: Day $78.5\%$, Dusk $12.1\%$, Night $9.4\%$.
(d) H3O time-of-day split by voxels: Day $56.9\%$, Dusk $19.4\%$, Night $23.8\%$.
(e) H3O weather split by voxels: Clear $60.6\%$, Fog $21.9\%$, Rain $17.5\%$.
(f) H3O weather $\times$ time of day (voxels): \textbf{clear/day} $91.75M$, \textbf{clear/night} $44.24M$, \textbf{fog/day} $40.96M$ dominate.
\emph{Notes:} (a,b) use non-empty voxel frequencies; (c) is frame-level stats for QuadOcc; (d--f) are voxel-level stats for H3O.}
\label{fig:supp_distribution}
\end{figure*}

\subsection{Temporal Aggregation and Refinement}
To suppress flicker and fill occlusions, we aggregate a temporal window $[t\!-\!n,\,t\!+\!n]$. All points inside the window are first transformed to the current frame by $\mathbf{T}^{\mathrm{cur}}_{\tau}\!\in\!\mathrm{SE(3)}$:
\begin{equation}
\label{eq:temporal-transform}
\mathcal{P}^{\mathrm{cur}}_{\tau} \;=\; \mathbf{T}^{\mathrm{cur}}_{\tau}\,\mathcal{P}_{\tau}, 
\qquad \forall\,\tau\in[t\!-\!n,\,t\!+\!n].
\end{equation}
Here, $\mathcal{P}_{\tau}$ is the LiDAR point set at time $\tau$ and $\mathcal{P}^{\mathrm{cur}}_{\tau}$ its transformed set. After devoxelization and re-voxelization, we accumulate per-class votes in each voxel and take a temporal majority:
\begin{equation}
\label{eq:temporal-vote}
\mathbf{V}_{\mathrm{agg}}[x,y,z] \;=\;
\arg\max_{c} \;\sum_{\tau=t-n}^{t+n} 
\mathbb{I}\!\big(\mathcal{P}^{\mathrm{cur}}_{\tau}(x,y,z)=c\big).
\end{equation}
Here, $\mathbf{V}_{\mathrm{agg}}$ denote the aggregated label volumes. $\mathbb{I}(\cdot)$ denotes the \emph{indicator function}:
$\mathbb{I}(\text{predicate})=1$ if the predicate is true and $0$ otherwise.

\textbf{Quantization-aware 3D mode (our implementation).}
Let $U$ denote the \emph{ignore/unlabeled} index, and define a quantization mask that flags ``empty-vote’’ voxels (no support from any frame):
\begin{equation}
\label{eq:quant-mask}
\mathcal{M}_q[x,y,z] \;=\; 
\mathbb{I}\!\Big(\sum\nolimits_{c}\sum\nolimits_{\tau} 
\mathbb{I}\!\big(\mathcal{P}^{\mathrm{cur}}_{\tau}(x,y,z)=c\big)\;=\;0\Big).
\end{equation}
We then apply a \emph{gated} 3D mode filter only where $\mathcal{M}_q{=}1$. 
Using a cubic neighborhood of size $k{\times}k{\times}k$ (default $k{=}5$), and let the local class histogram $H_c(\mathbf{x})$ be:
\[
\mathbf{x}=(x,y,z)^\top,\qquad 
\mathcal{N}_k=\big\{\boldsymbol{\delta}\in\mathbb{Z}^3:\ \|\boldsymbol{\delta}\|_\infty\le \lfloor k/2\rfloor\big\}.
\]

\begin{equation}
H_c(\mathbf{x}) \;=\; 
\sum_{\boldsymbol{\delta}\in\mathcal{N}_k}
\mathbb{I}\!\big(\mathbf{V}_{\mathrm{agg}}[\mathbf{x}{+}\boldsymbol{\delta}]=c\big).
\end{equation}

The refined label $\mathbf{V}_{\mathrm{ref}}$ is then:
\begin{equation}
\mathbf{V}_{\mathrm{ref}}[\mathbf{x}] \;=\;
\begin{cases}
\underset{c\neq U}{\arg\max}\; H_c(\mathbf{x}), & \mathcal{M}_q[\mathbf{x}]=1,\\[0.4ex]
\mathbf{V}_{\mathrm{agg}}[\mathbf{x}], & \text{otherwise.}
\end{cases}
\end{equation}
\textbf{Ties and safety:} in case of a tie, we keep $\mathbf{V}_{\mathrm{agg}}[\mathbf{x}]$ (no change). If the neighborhood histogram is empty (all $U$), we fall back to a nearest-neighbor fill in Euclidean space with a small radius (default $r_{\max}{=}5$ voxels) and still ignore $U$; if no valid neighbor exists within $r_{\max}$ the voxel remains $U$. Border handling uses a constant pad (out-of-bounds treated as $U$). This quantization-aware gating prevents the mode operation from overwriting confident voxels while reliably filling holes introduced by multi-frame alignment and re-voxelization.

\begin{figure*}[t]
  \centering
  \includegraphics[width=0.8\linewidth]{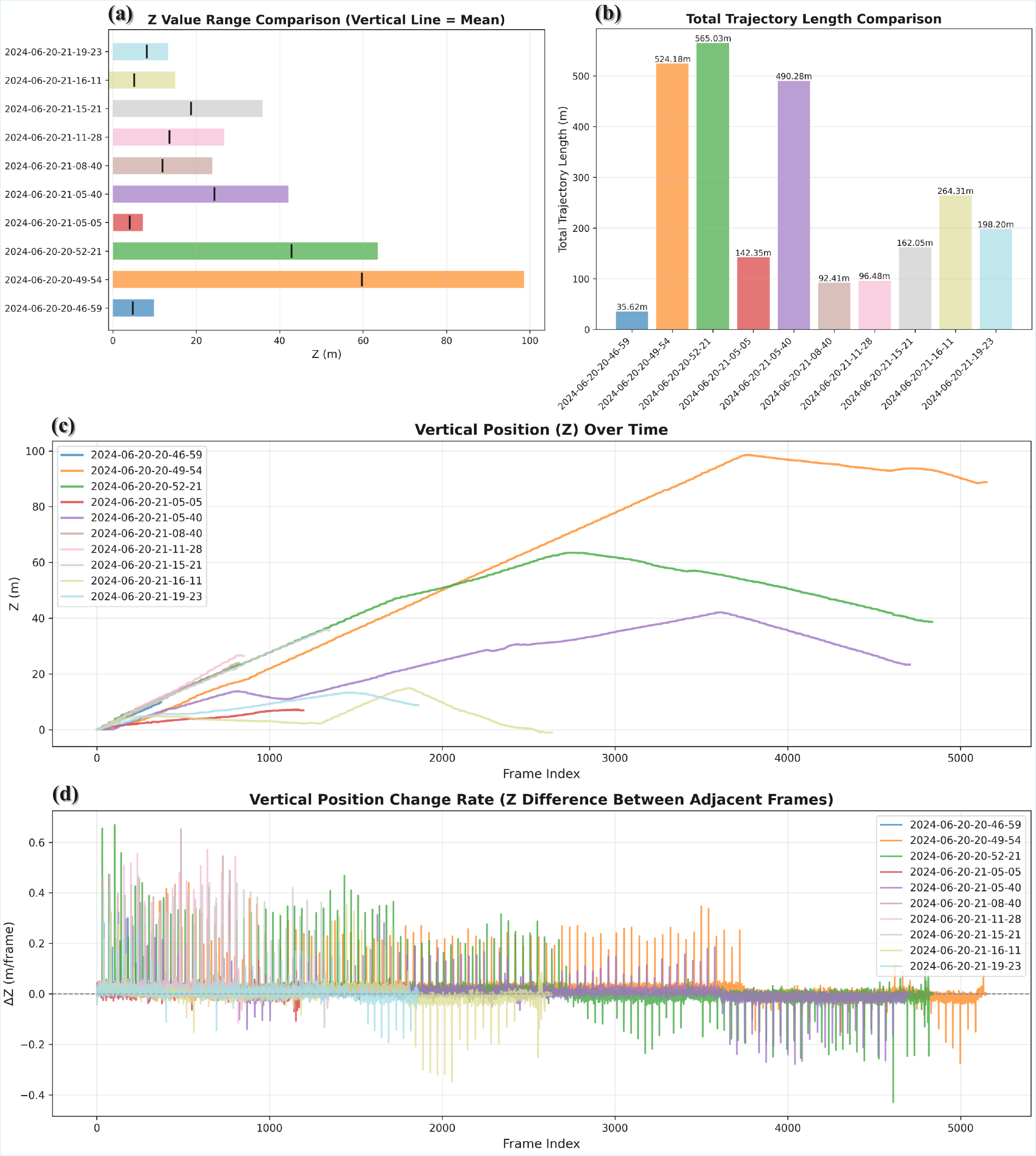}
   \vskip 4ex
  \caption{\textbf{QuadOcc camera pose statistics.}
  Visualization of the ego camera pose distribution on QuadOcc in the canonical grid frame (X forward, Y left, Z up).
  (a)--(d) show complementary views of the aggregated 6-DoF poses over all sequences: top-down ego trajectories, heading statistics, and height/orientation distributions.
  The coverage concentrates within the $12.8$\,m radius and $64{\times}64{\times}8$ voxel grid used for ground-truth construction, confirming that the chosen range and resolution (Sec.~4.7) are well-aligned with real quadruped operation and the gait-induced body motion that motivates our GDC and bi-grid design.}
  \vskip -2ex
  \label{fig:quadocc_pose}
\end{figure*}

\subsection{Data Splits, Resolution, and Taxonomy}
Our legged/humanoid setting prioritizes reliable near-field awareness under tight onboard compute.
We therefore bound the grid to $R=[-X_{\max},X_{\max}]\!\times\![-Y_{\max},Y_{\max}]\!\times\![Z_{\min},Z_{\max}]$
with $X_{\max}\!=\!Y_{\max}\!=\!12.8$\,m, $Z_{\min}\!=\!-2.0$\,m, $Z_{\max}\!=\!1.2$\,m, and a native cubic voxel size
$\Delta\!=\!0.4$\,m, yielding a $64\!\times\!64\!\times\!8$ grid that balances (i) adequate radius for low-speed legged motion
and foothold safety, (ii) memory/latency budgets for real-time inference on embedded GPUs, and
(iii) azimuthal continuity for panoramic cues.
We also standardize a six-class schema \{vehicle, pedestrian, road, building, vegetation, terrain\} to stabilize training under long-tailed class frequencies while keeping labels deployable for onboard planning. The corpus contains ten scenes and $24K$ frames captured under day/dusk/night (the latter also used in stress tests). 

\subsection{Quality Control and Semi-Automatic Labeling}
Our labels originate from multi-frame LiDAR accumulation, Grounded-SAM initialization, and targeted manual fixes in ambiguous regions and around thin structures. During quality control, we flag voxels whose $\mathrm{conf}(V_i)$ or temporal agreement rates fall below thresholds, and prioritize these for human proofreading.\vspace{2pt}

\begin{figure*}[t]
  \centering
  \tikzset{
    block/.style={draw, rounded corners, very thick, align=center,
                  inner sep=2pt, minimum width=3.05cm, minimum height=1.2cm},
    arr/.style={-Latex, very thick},
    tinyanno/.style={font=\scriptsize, inner sep=1pt}
  }
  \begin{tikzpicture}[node distance=0.9cm and 0.6cm]
    \node[block] (cap) {Panoramic Capture\\{\footnotesize Cubemap $\rightarrow$ ER, Gait Bob}\\{\scriptsize Eq.~\ref{eq:gait-bob}}};
    \node[block, right=of cap] (p2p) {ER Depth/Semantics\\$\Rightarrow$ 3D Points\\{\scriptsize Eqs.~\ref{eq:equirect-to-spherical}--\ref{eq:spherical-to-cartesian}}};
    \node[block, right=of p2p] (pose) {Pose Alignment \&\\Grid Frame\\{\scriptsize Eqs.~\ref{eq:grid-frame}, \ref{eq:align-to-grid}}};

    \node[block, below=1.2cm of cap] (agg) {Spatiotemporal Aggregation\\Dynamic Compensation (OBB)\\{\scriptsize Eq.~\ref{eq:inside-obb}}};
    \node[block, right=of agg] (vox) {Voxelization\\{\scriptsize Eq.~\ref{eq:voxel-index}}};
    \node[block, right=of vox] (label) {Label Assignment\\Majority + Density\\{\scriptsize Eq.~\ref{eq:majority}}};
    \node[block, right=of label] (post) {Post-Processing\\OBB Painting / Cleanup\\{\scriptsize Eq.~\ref{eq:paint-obb}}};

    \draw[arr] (cap) -- (p2p);
    \draw[arr] (p2p) -- (pose);
    \draw[arr] (pose.south) |- ++(0,-0.6) -| (agg.north);

    \draw[arr] (agg) -- (vox);
    \draw[arr] (vox) -- (label);
    \draw[arr] (label) -- (post);

    \node[tinyanno, above=0.05cm of cap] {RGB / Depth / Semantics (ER)};
    \node[tinyanno, above=0.05cm of post] {GT Occupancy: 64$\times$64$\times$8 ~ / ~ 128$\times$128$\times$16};
  \end{tikzpicture}
  \caption{\textbf{H3O ground-truth construction pipeline.} Starting from synchronized cubemap panoramas, we stitch to an equirectangular (ER) image and emulate legged motion with a speed-conditioned vertical bob (Eq.~\ref{eq:gait-bob}). ER depth and semantics are back-projected to the camera frame by ER$\!\to\!$sphere$\!\to\!$3D conversion (Eqs.~\ref{eq:equirect-to-spherical}--\ref{eq:spherical-to-cartesian}). We then align points to a canonical grid frame (forward $X$, up $Z$, left $Y$ via $Y$-flip) using a reference pose (Eqs.~\ref{eq:grid-frame}, \ref{eq:align-to-grid}). A spatiotemporal aggregation window stabilizes moving actors through OBB-based local registration and transport to the reference time $\tau$ (Eq.~\ref{eq:inside-obb}), mitigating ghost trails. The aligned points are voxelized within a fixed 3D range to indices $(i,j,k)$ (Eq.~\ref{eq:voxel-index}) and labeled by majority with density safeguards (Eq.~\ref{eq:majority}). Finally, we apply post-processing to ensure complete occupancy of dynamic instances (\emph{OBB painting}) and clean small/noisy components (Eq.~\ref{eq:paint-obb}). Outputs include semantic occupancy at $64{\times}64{\times}8$ and $128{\times}128{\times}16$ over the same spatial bounds, together with per-sequence metadata for exact reproducibility.}

  \label{fig:h3o_pipeline_single}
\end{figure*}

\begin{figure}[t]
\centering
\includegraphics[width=0.95\linewidth]{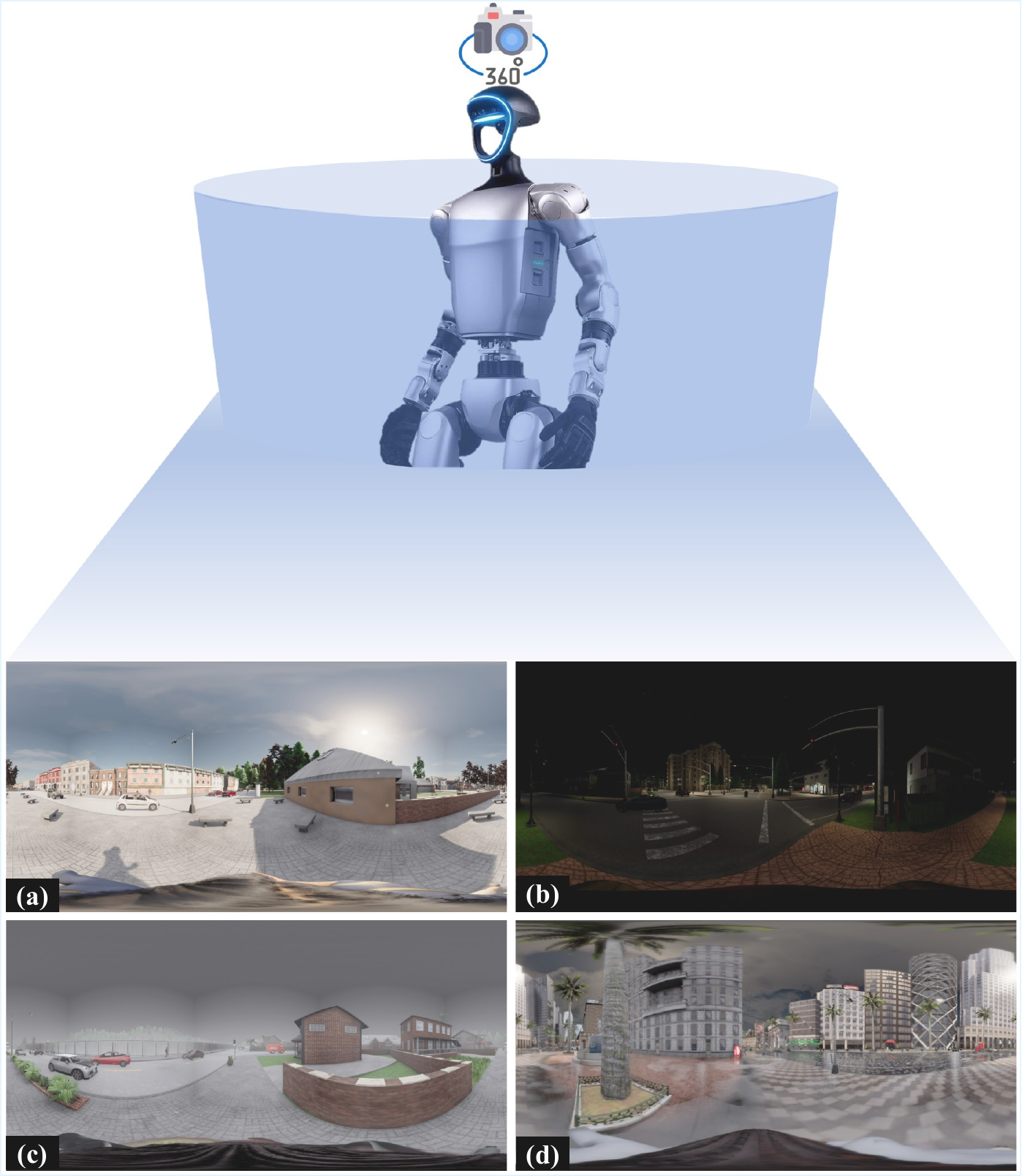}
\vskip -1.5ex
\caption{\textbf{H3O: Weather and lighting diversity.}
Representative equirectangular (ER) panoramas illustrating the variety of conditions covered by \textit{Human360Occ (H3O)}.
(a) Clear—daytime; (b) Clear—night; (c) Fog; (d) Rain/overcast in a metropolitan scene.
All frames are native ER inputs used directly for training/evaluation.
Across H3O, voxel-level distributions span \emph{Clear/Fog/Rain} and \emph{Day/Dusk/Night}, enabling controlled stress-tests under illumination and adverse weather.
Ground-truth supervision is provided at two resolutions (64$\times$64$\times$8 and 128$\times$128$\times$16) over the same spatial bounds.}
\label{fig:weather_h3o}
\vskip -4ex
\end{figure}

\subsection{Analysis: Distribution, Difficulty, and Lighting}
\paragraph{Class imbalance.} On non-empty voxels, road dominates ($\approx52.3\%$) followed by vegetation ($\approx22.8\%$) and building ($\approx14.0\%$), while vehicle/pedestrian are rare ($\approx0.9\%/2.1\%$). This long-tail calls for per-class reweighting or sampling-aware training.
\paragraph{Dynamic vs. static.} Dynamic tubes reduce motion ghosts and improve temporal coherence near moving objects; they also ease targeted manual verification.
\paragraph{Lighting.} Day/dusk/night coverage intentionally stresses robustness. We observe that crepuscular lighting often improves geometric precision (reduced glare), whereas nighttime increases sparsity and aliasing in the far range, recommending stronger temporal priors or test-time refinement.
\paragraph{Confidence-driven curation.} The per-voxel confidence and windowed agreement are reliable indicators for spotting residual errors at class boundaries (\textit{e.g.}, vehicle--road).

\section{Human360Occ: Dataset Construction}
\label{sec:h3o-gt}

\begin{figure*}[t]
  \centering
  \includegraphics[width=\linewidth]{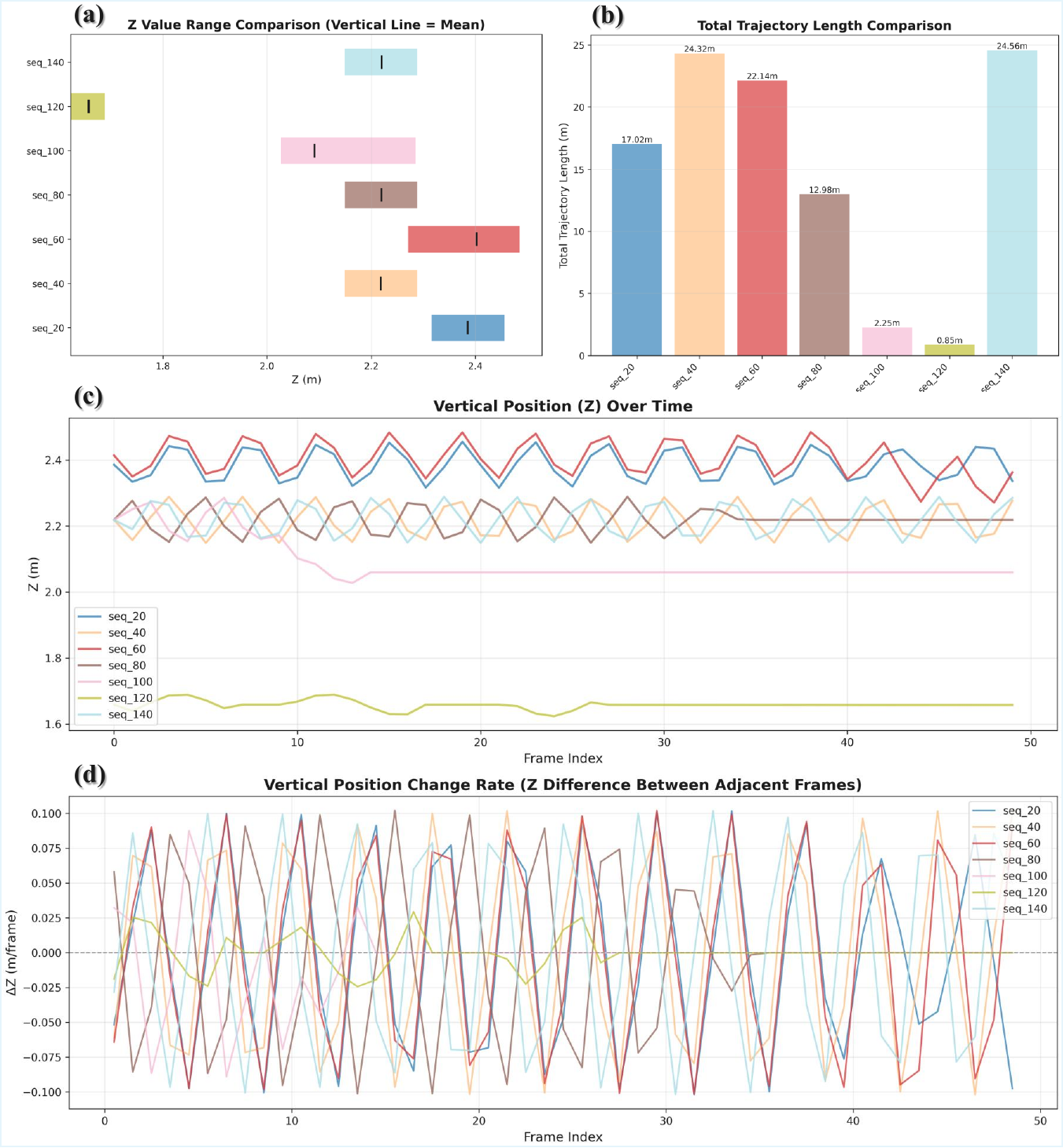}
  \vskip -2.5ex
  \caption{\textbf{Human360Occ (H3O) ego pose distribution.}
  Camera pose statistics for Human360Occ in the same grid frame as used for QuadOcc, aggregated over 160 sequences across 16 CARLA maps and diverse weather/lighting conditions.
  (a)--(d) provide complementary views of the ego trajectories and 6-DoF poses, including top-down coverage and orientation/height distributions.
  The plots highlight that H3O exhibits wide heading coverage, non-trivial vertical bobbing, and varied viewpoints, which together stress-test panoramic SSC under gait-like motion and support the standardized within-/cross-city splits and dual-resolution occupancy volumes used in our benchmarks.}
  \label{fig:h3o_pose}
  \vskip -3.5ex
\end{figure*}

\paragraph{Goal and setting}
Human360Occ (H3O) targets full-surround semantic occupancy for legged/humanoid platforms from a \emph{single} panoramic stream. Each frame provides RGB, metric depth, and semantic panoramas, and the ground-truth (GT) occupancy at two resolutions ($64{\times}64{\times}8$ and $128{\times}128{\times}16$). We standardize within-/cross-city splits and record rich metadata for reproducibility.

\subsection{Panoramic capture}
\noindent\textbf{Cubemap rig and stitching.}
We mount six synchronized virtual cameras in CARLA~\cite{dosovitskiy2017carla} (front/right/back/left/up/down) with unified exposure to avoid face-wise brightness drift. Images from the six faces are stitched into an equirectangular panorama using a calibrated cubemap-to-ER pipeline.

\smallskip\noindent\textbf{Gait-induced motion (bob).}
To emulate first-person legged motion, the camera center undergoes a vertical bob whose amplitude and frequency adapt to walking speed. Let $t$ be time (s), $v(t)$ the ego speed, $A(v)$ the amplitude (m), and $f(v)$ the frequency (Hz). The camera $z$ position is:
\begin{equation}
z(t) \;=\; z_0 \;+\; A\!\big(v(t)\big)\,\cdot\sin\!\big(2\pi \cdot f\!\big(v(t)\big)\cdot t + \phi\big),
\label{eq:gait-bob}
\end{equation}
where $A(v)$ linearly interpolates within $[A_{\min},A_{\max}]$ and $f(v)$ within $[f_{\min},f_{\max}]$ as speed increases.

\smallskip\noindent\textbf{Environment randomization.}
We randomize the environment at the \emph{sequence} level across maps, weather, and time-of-day with fixed quotas for coverage.
Concretely, H3O spans \textbf{16} distinct CARLA maps (10 sequences per map) and totals \textbf{160} sequences (\textbf{8{,}000} frames).
Weather is sampled from \texttt{\{Clear, Fog, Rain\}}, yielding the realized distribution: Clear 97 (\(\approx\)60.6\%), Fog 35 (\(\approx\)21.9\%), Rain 28 (\(\approx\)17.5\%).
Time-of-day is sampled from \texttt{\{day, dusk, night\}}, realized as: day 91 (\(\approx\)56.9\%), dusk 31 (\(\approx\)19.4\%), night 38 (\(\approx\)23.8\%).
For exact reproducibility, each sequence provides a \texttt{sequence\_meta.json} that records the loaded map ID, time-of-day label, weather type and parameters, and simulation tick rate.

\subsection{From panoramas to a registered 3D point cloud}
\noindent\textbf{Depth-to-3D in camera coordinates.}
Given an equirectangular depth map $r(u,v)$ (meters) of size $W{\times}H$ with pixel coordinates $(u,v)\in[0,W)\times[0,H)$, we first map pixels to spherical angles:
\begin{equation}
\theta(u)=\tfrac{2\pi}{W}u-\pi,\qquad
\phi(v)=\tfrac{\pi}{H}v,
\label{eq:equirect-to-spherical}
\end{equation}
and then project to the local camera frame (left-handed, $X$ forward, $Y$ right, $Z$ up):
\begin{equation}
\mathbf{p}_{c}(u,v)\;=\;
\begin{bmatrix}
x\\y\\z
\end{bmatrix}
=
r(u,v)\!
\begin{bmatrix}
\sin\phi\,\cos\theta\\[2pt]
\sin\phi\,\sin\theta\\[2pt]
\cos\phi
\end{bmatrix}.
\label{eq:spherical-to-cartesian}
\end{equation}
We discard invalid measurements (too near/far) and sky pixels and keep $(\mathbf{p}_{c},\ell)$ pairs where $\ell$ is the Cityscapes-style~\cite{cordts2016cityscapes} semantic ID.

\smallskip\noindent\textbf{Pose alignment and grid frame.}
Let $\mathbf{T}_{w\leftarrow c(t)}\in SE(3)$ be the camera pose at time $t$ and define the grid frame so that $X$ is forward, $Z$ is up, and $Y$ points \emph{left} (a $Y$-flip w.r.t. the camera). If $\mathbf{F}_y=\mathrm{diag}(1,-1,1,1)$, then the world-to-grid transform at a \emph{reference} frame $\tau$ is
\begin{equation}
\mathbf{T}_{g\leftarrow w} \;=\; \big(\mathbf{T}_{w\leftarrow c(\tau)}\,\mathbf{F}_y\big)^{-1}.
\label{eq:grid-frame}
\end{equation}
A point captured at time $t$ transforms to the grid frame as
\begin{equation}
\tilde{\mathbf{p}}_{g} \;=\; \mathbf{T}_{g\leftarrow w}\,\mathbf{T}_{w\leftarrow c(t)}\,
\begin{bmatrix}\mathbf{p}_{c}\\1\end{bmatrix}\!,
\qquad \mathbf{p}_{g}=\Pi(\tilde{\mathbf{p}}_{g}),
\label{eq:align-to-grid}
\end{equation}
where $\Pi$ drops the homogeneous coordinate. We aggregate a temporal window of frames around $\tau$ (Sec.~\ref{sec:agg}).

\subsection{Spatiotemporal aggregation with dynamic compensation}
\label{sec:agg}
We stabilize moving actors (cars, walkers) to the reference time $\tau$ to mitigate ``ghost tubes'' in time-accumulated clouds.

\smallskip\noindent\textbf{OBB-based local registration.}
For every dynamic actor, we obtain an oriented bounding box (OBB) at time $t$ via its rigid transform $\mathbf{T}^{(t)}_{\mathrm{box}}$ and extents $\mathbf{e}=(e_x,e_y,e_z)$. A world point $\mathbf{p}_w$ is inside a \emph{scaled} OBB if
\begin{equation}
\big|\big(\mathbf{T}^{(t)}_{\mathrm{box}}\big)^{-1}\,[\mathbf{p}_w;1]\big| \;\preceq\; \mathbf{e}\!\odot\! s \;+\; \mathbf{m},
\label{eq:inside-obb}
\end{equation}
where $s{>}1$ is a scale factor, $\mathbf{m}$ is a per-axis margin (m), and $\preceq$ is element-wise. For points inside, we transport them from time $t$ to $\tau$ by composing actor poses; otherwise we keep the standard camera-pose alignment of \eqref{eq:align-to-grid}. As a fall-back, we optionally \emph{remove} dynamic points from non-reference frames or \emph{paint} current-frame OBBs into the grid (Sec.~\ref{sec:post}).

\subsection{Voxelization and semantic labeling}
\noindent\textbf{Grid bounds and resolution.}
We set an asymmetric vertical range anchored at the ego camera: $Z_{\min}\!=\!-2.0$\,m ensures sufficient ground coverage and small depressions below the head-height camera, while $Z_{\max}\!=\!1.2$\,m captures overhanging obstacles without wasting vertical resolution.
We therefore define a fixed 3D range
\[
\mathcal{R}=[-X_\mathrm{max},X_\mathrm{max}]\times[-Y_\mathrm{max},Y_\mathrm{max}]\times[Z_{\min},Z_{\max}],
\]
with defaults $X_\mathrm{max}=Y_\mathrm{max}=12.8\,$m, $Z_{\min}=-2.0\,$m, $Z_{\max}=1.2\,$m, and a native cubic voxel size $\Delta=0.4\,$m. The native grid size is $\mathbf{N}=(N_x,N_y,N_z)=\big\lfloor(\mathrm{max}-\mathrm{min})/\Delta\big\rceil$, and we also export higher resolutions (\textit{e.g.}, $128{\times}128{\times}16$) over the \emph{same} $\mathcal{R}$. 

\smallskip\noindent\textbf{Indexing.}
A grid index for a point $\mathbf{p}_{g}=(x,y,z)$ is
\begin{equation}
(i,j,k)\;=\;\Big\lfloor\frac{x-x_{\min}}{\Delta}\Big\rfloor,\;
\Big\lfloor\frac{y-y_{\min}}{\Delta}\Big\rfloor,\;
\Big\lfloor\frac{z-z_{\min}}{\Delta}\Big\rfloor.
\label{eq:voxel-index}
\end{equation}

\smallskip\noindent\textbf{Label assignment with density safeguards.}
Let $\mathcal{P}_{i,j,k}$ be the set of aggregated points falling into voxel $(i,j,k)$ and $\ell(\cdot)$ their semantic IDs. We adopt majority voting with a minimum support:
\begin{align}
\hat{c}_{i,j,k} \,&= \arg\max_{c}\,\#\{\,\mathbf{p}\!\in\!\mathcal{P}_{i,j,k}\,:\,\ell(\mathbf{p}){=}c\,\}, \\
\text{keep if } \;\#\mathcal{P}_{i,j,k} \,&\ge \tau,\qquad
\tau \;=\; \max\!\big(N_{\min},\,\lceil\rho_{\min}\,V_{\text{vox}}\rceil\big),
\label{eq:majority}
\end{align}
where $\#S$ denotes the cardinality (number of elements) of a set $S$. $N_{\min}$ is a point-count threshold, $\rho_{\min}$ is an optional density floor (points/m$^3$), and $V_{\text{vox}}$ is the voxel volume. Unless stated otherwise, we use majority voting with $N_{\min}=4$ and optionally enable the density floor.

\subsection{Post-processing}
\label{sec:post}
\noindent\textbf{OBB painting.}
To ensure complete occupancy for current-frame dynamic instances, we fill their OBBs in the grid using a scaled+margin ROI $(s,\mathbf{m})$:
\begin{equation}
\label{eq:paint-obb}
\begin{aligned}
&\text{for voxels } v \in \mathrm{ROI}:\\
&\quad \mathrm{occ}[v] \leftarrow 
\begin{cases}
c_{\text{actor}}, & \text{\tt override},\\
c_{\text{actor}} \ \text{if } \mathrm{occ}[v]=0, & \text{\tt keep}.
\end{cases}
\end{aligned}
\end{equation}
We analogously paint known static vehicles (\emph{environment objects}) with a slightly tighter ROI.

\smallskip\noindent\textbf{Noise removal.}
We remove isolated singletons (no same-label neighbors) and delete small connected components below a voxel-count threshold using spatial-connectivity (we favor 6-connectivity to preserve thin structures).

\smallskip\noindent\textbf{Semantic schema and storage.}
Occupancy volumes are stored as \verb|uint8| NumPy arrays of shape $[N_x,N_y,N_z]$, where $0$ denotes empty and $1{\sim}28$ follow a Cityscapes-style palette (\textit{e.g.}, road/sidewalk/building/vegetation/person/car/truck/bus/twowheeler, \textit{etc.}). We export both native and $128{\times}128{\times}16$ grids, plus optional colored PCDs of voxel centers for visual auditing.

\subsection{Train/val splits and stats}
H3O comprises $160$ sequences / $8K$ frames across $16$ maps under diverse weather and lighting. Each frame includes RGB, metric depth, ER semantics, poses, and GT occupancy at two resolutions, supporting within-city (per-map $8{:}2$) and cross-city ($12{/}4$ maps) protocols.

\subsection{Default configuration}
Unless specified otherwise, we use: time window $T=50$ frames centered at $\tau$; native voxel size $\Delta=0.4$\,m over $\mathcal{R}=[\pm12.8\,\mathrm{m}]$ (XY) and $[-2.0,1.2]$\,m (Z); label assignment \texttt{majority} with $N_{\min}{=}4$ (optionally density-gated); OBB fill for dynamic and static vehicles enabled; and singleton/small-component removal.

\paragraph{Deliverables.}
For each sequence, we provide ER panoramas (RGB/Depth/Semantics), per-frame \texttt{ground\_truth.json} (poses, actor OBBs, ego info), \texttt{sequence\_meta.json}, native and $128{\times}128{\times}16$ occupancy volumes, and optional point-cloud visualizations for quality control.

\section{Implementation Details of Baselines}

\paragraph{General protocol.}
All camera-only baselines are \emph{retrained from scratch} on our panoramic legged-robot benchmarks with \emph{minimal modifications}: we preserve original backbones/heads and only adapt dataset hooks, projection reshaping, voxel grid sizes, and loss toggles required by the panoramic small-volume regime.
Unless otherwise noted, we fix seeds, follow original training schedules, and report results with the default settings stated under each method.

\subsection{Overview}
We organize per-method details in a uniform structure to facilitate verification and extension:

\begin{compactitem}
  \item \textbf{[Method Name]}: a short protocol summary (minimal changes, training schedule).
  \item \textbf{QuadOcc}: dataset- and grid-specific settings for retraining on our QuadOcc split.
  \item \textbf{H3O}: corresponding settings for retraining on H3O.
  \item \textbf{Design rationale}: why these choices are necessary under panoramic, small-volume settings.
\end{compactitem}

\subsection{MonoScene}
\paragraph{Scope.}
We keep \textsc{MonoScene}'s~\cite{cao2022monoscene} original design and adapt only the dataset branch, projection reshape,
and losses required by panoramic inputs and smaller grids; all models are trained from scratch.

\begin{table*}[t]
\centering
\caption{\textbf{QuadOcc vs.\ KITTI~\cite{semantickitti} (MonoScene~\cite{cao2022monoscene}).}}
\small
\resizebox{0.8\linewidth}{!}{
\begin{tabular}{lccp{4.7cm}}
\toprule
Component & KITTI~\cite{semantickitti} & QuadOcc & Rationale \\
\midrule
Grid size & $256{\times}256{\times}32$ & $64{\times}64{\times}8$ & Embedded compute / near-field focus \\
Voxel size & 0.2\,m & 0.4\,m & Balanced radius vs.\ memory \\
Classes & 20 & 6 & legged-relevant labels \\
Relation loss & \texttt{true} & \texttt{false} & Unstable / unhelpful on small grids \\
3D decoder & KITTI default & small-grid CP preserved & Prevent over-downsampling \\
\bottomrule
\end{tabular}
}
\end{table*}

\paragraph{QuadOcc.}
\textbf{Data and taxonomy.}
QuadOcc provides per-frame panoramic RGB and voxelized semantic occupancy within the 3D range
$X{=}{Y}{=}\pm 12.8$\,m, $Z\in[-2.0,\,1.2]$\,m, discretized into a $64{\times}64{\times}8$ grid at $0.4$\,m resolution.
We train on 7 labels with \texttt{empty} at index 0 and six non-empty classes:
\{\texttt{vehicle}, \texttt{pedestrian}, \texttt{road}, \texttt{building}, \texttt{vegetation}, \texttt{terrain}\}.
Images are equirectangular panoramas of size $370{\times}1220$ produced by calibrated OCam/Taylor~\cite{scaramuzza2006toolbox} unwarping of a PAL camera.
We use the official train/val sequences and sample every $5$ frames.

\noindent\textbf{Architecture adaptations.}
We adopt the small-grid CP and standard lifting without extra magnification:
\begin{compactitem}
  \item \emph{Context Prior (CP) for small grids:}
  the CP module (\texttt{CPMegaVoxels}) dynamically adjusts kernel sizes/strides as a function of the minimum side-length;
  if any side is $<3$, it falls back to $1^3$ kernels with unit stride to avoid over-downsampling.
  \item \emph{Logit--GT alignment:}
  the decoder head ensures output logits match the $64{\times}64{\times}8$ GT grid exactly after lifting.
  \item \emph{FLoSP lifting (QuadOcc):}
  we use the \emph{standard} FLoSP reshape consistent with the chosen \texttt{project\_scale}.
\end{compactitem}

\noindent\textbf{Training hyper-parameters.}
We mostly keep the KITTI~\cite{semantickitti} recipe and only adapt what is necessary:
\begin{compactitem}
  \item \emph{Scene/voxel:} \texttt{full\_scene\_size}$=(64,64,8)$, voxel size $0.4$\,m, origin $[-12.8,-12.8,-1.2]$.
  \item \emph{Backbone/decoder:} feature dimension $64$.
  \item \emph{Classes/losses:} $n\_classes{=}7$ (including \texttt{empty});
  enable \texttt{CE\_ssc\_loss}, \texttt{sem\_scal\_loss}, \texttt{geo\_scal\_loss}, and \texttt{fp\_loss} with \texttt{frustum\_size}$=8$;
  \underline{disable} relation loss (\texttt{relation\_loss=false}) due to small grids and panoramic distortions.
  \item \emph{Optimization:} AdamW~\cite{AdamW} with LR$=10^{-4}$, WD$=10^{-4}$; MultiStepLR (\texttt{milestones}$=[20]$, $\gamma=0.1$);
  \texttt{max\_epochs}$=30$, \texttt{batch\_size}$=1$, \texttt{n\_gpus}$=1$.
  \item \emph{Data loading:} \texttt{num\_workers\_per\_gpu}$=16$.
  \item \emph{Augmentation:} color jitter $(0.4,0.4,0.4)$ at train-time; no augmentation at val/test.
\end{compactitem}

\noindent\textbf{Camera and normalization.}
We use OCam/Taylor~\cite{scaramuzza2006toolbox} intrinsics and the precomputed annulus$\rightarrow$equirectangular maps.
RGB normalization: $\mu=\left[\tfrac{123.971}{255},\,\tfrac{123.564}{255},\,\tfrac{164.785}{255}\right]$,
$\sigma=\left[\tfrac{84.271}{255},\,\tfrac{88.170}{255},\,\tfrac{72.966}{255}\right]$.

\noindent\textbf{Loss weighting.}
We compute frequency-based class weights on the QuadOcc training split (non-empty voxels) for the CE term
to stabilize rare dynamic categories and the head class (\texttt{road}).

\noindent\textbf{Design rationale.}
(i) \emph{Small-grid \& dynamic CP:} with $64{\times}64{\times}8$, aggressively strided $3^3$ stacks can collapse features; dynamic kernels retain receptive fields without over-downsampling. %
(iii) \emph{Disable relation loss:} co-occurrence priors that help on KITTI~\cite{semantickitti} tend to over-smooth azimuthal seams and suppress rare classes under panoramas. %
(iv) \emph{jitter on:} moderate color jitter improves robustness.

\paragraph{H3O.}
\begin{table*}[t]
\centering
\caption{\textbf{H3O vs.\ KITTI~\cite{semantickitti} (MonoScene~\cite{cao2022monoscene}).}}
\small
\resizebox{0.8\linewidth}{!}{
\begin{tabular}{lccp{4.3cm}}
\toprule
Component & KITTI~\cite{semantickitti} & H3O & Rationale \\
\midrule
Grid size & $256{\times}256{\times}32$ & $64{\times}64{\times}8$ / $128{\times}128{\times}16$ & Near-field focus vs.\ detail \\
Voxel size & 0.2\,m & 0.4/0.2\,m & Match grid mode and memory \\
Projection scale & 2 & 1 (standard reshape) & Lifting without over-enlargement \\
Classes & 20 & 11 & Stable yet granular labels \\
Relation loss & \texttt{true} & \texttt{false} & Avoid over-smoothing rare classes \\
Batch size & 1 & 2 & Smaller grid $\Rightarrow$ higher throughput \\
\bottomrule
\end{tabular}
}
\end{table*}

\textbf{Projection (FLoSP).}
We use the \emph{standard} FLoSP reshape consistent with \texttt{project\_scale}$=1$.

\noindent\textbf{Voxel grid and scene bounds.}
Two modes are supported: \emph{native} $64{\times}64{\times}8$ at $0.4$\,m/voxel (default) and
\emph{fixed 128} $128{\times}128{\times}16$ at $0.2$\,m/voxel; both share $25.6{\times}25.6{\times}3.2$\,m
with origin $[-12.8,-12.8,-2.4]$\,m.

\noindent \textbf{Remapping and losses.}
We adopt the default 11-class remap (\texttt{car}/\texttt{truck}/\texttt{bus} separated; traffic furniture merged).
We enable \texttt{CE\_ssc\_loss}, \texttt{sem\_scal\_loss}, \texttt{geo\_scal\_loss}, \texttt{fp\_loss} (\texttt{frustum\_size}$=8$),
and \underline{disable} relation loss.

\noindent \textbf{Image pipeline and optimization.}
Panoramas are resized to $352{\times}1216$ and normalized with statistics
($\mu{=}[0.485,0.456,0.406]$, $\sigma{=}[0.229,0.224,0.225]$).
We follow the KITTI~\cite{semantickitti} schedule (AdamW~\cite{AdamW}, LR$=10^{-4}$, WD$=10^{-4}$; MultiStepLR at epoch 20 with $\gamma{=}0.1$; 30 epochs).
The smaller grid permits \texttt{batch\_size}$=2$ and \texttt{num\_workers\_per\_gpu}$=12$ on a single GPU.

\noindent \textbf{Design rationale.}
(i) \emph{Equirect + standard reshape:} no extra magnification is required for uniform azimuthal sampling;
this reduces aliasing. %
(ii) \emph{Two grid modes:} \texttt{native} trades speed for scale; \texttt{fixed128} recovers thin structures. %
(iii) \emph{Relation loss off:} improves stability under panoramic distortions and class imbalance.

\subsection{SGN}
\paragraph{Scope.}
We keep \textsc{SGN}'s~\cite{mei2023camera} transformer-based image-to-voxel lifting and the original detector--head layout, and make only minimal changes required by panoramic inputs and small voxel grids. All models are trained from scratch. Single-GPU batches $\geq$1 are fully supported by us in both training and testing.

\begin{table*}[t]
\centering
\caption{\textbf{KITTI~\cite{semantickitti} vs.\ QuadOcc (SGN~\cite{mei2023camera}).}}
\small
\resizebox{0.8\linewidth}{!}{
\begin{tabular}{lccp{4.9cm}}
\toprule
Component & KITTI~\cite{semantickitti} baseline & QuadOcc (ours) & Rationale \\
\midrule
Grid size (GT) & $256{\times}256{\times}32$ & $64{\times}64{\times}8$ & Near-field grid for embedded budget \\
Voxel size & 0.2\,m & 0.4\,m & Trade detail for memory/speed \\
Bounds (x/y/z) & $[0,{-}25.6,{-}2,\,51.2,25.6,4.4]$ & $[-12.8,{-}12.8,{-}1.2,\,12.8,12.8,2.0]$ & Symmetric, small-volume scenes \\
2D input & Pinhole RGB & ER panorama $370{\times}1220$ & Panoramic coverage \\
2D tokens/PE & Planar PE & ER tokens + spherical (lon/lat) terms & Longitude/latitude-aware tokenization \\
Lifting & X-attn: 3D$\leftarrow$2D & Same (X-attn) & Transformer lifting unchanged \\
3D queries & Per voxel (dense) & Per voxel (dense) & Same mechanism; fewer voxels \\
Classes & 20 & 7 (\texttt{empty}+6) & Dataset taxonomy \\
Losses & CE/sem\_scal/geo\_scal & Same & Stable on small grids \\
\bottomrule
\end{tabular}
}
\end{table*}

\paragraph{QuadOcc.}
\textbf{Data and taxonomy.}
QuadOcc provides single-frame omnidirectional RGB and voxelized semantic occupancy within
$X{=}{Y}{=}\pm 12.8$\,m and $Z\in[-2.0,\,1.2]$\,m, discretized at $0.4$\,m into a $64{\times}64{\times}8$ grid.
We use $7$ labels with \texttt{empty} at index~0 and six non-empty classes:
\{\texttt{vehicle}, \texttt{pedestrian}, \texttt{road}, \texttt{building}, \texttt{vegetation}, \texttt{terrain}\}.
Images are equirectangular panoramas of size $370{\times}1220$ obtained via calibrated OCam/Taylor~\cite{scaramuzza2006toolbox} unwarping.

\noindent\textbf{Architecture adaptations (transformer lifting).}
\begin{compactitem}
  \item \emph{3D queries.} We instantiate one 3D query per voxel center within the native bounds; each query carries a 3D sinusoidal positional encoding.
  \item \emph{2D tokens for panoramas.} ER images are encoded by a 2D CNN and flattened into tokens with 2D positional encodings augmented by spherical (longitude/latitude) terms computed from OCam/Taylor~\cite{scaramuzza2006toolbox} intrinsics.
  \item \emph{Cross-attention lifting.} Multi-head cross-attention lets 3D queries sample informative 2D tokens to produce fused voxel features; The decoder head maps fused features to per-voxel logits aligned with the $64{\times}64{\times}8$ GT grid.
\end{compactitem}

\noindent\textbf{Losses, schedules, and IO.}
We enable \texttt{CE\_ssc\_loss}, \texttt{sem\_scal\_loss}, and \texttt{geo\_scal\_loss}. %
Optimization uses AdamW~\cite{AdamW} (LR $=2{\times}10^{-4}$, WD $=10^{-2}$) with CosineAnnealing and a linear warmup of $500$ iterations; \texttt{total\_epochs}$=48$, \texttt{batch\_size}$=1$, \texttt{workers\_per\_gpu}$=4$. %
RGB normalization uses dataset statistics
$\mu=\left[\tfrac{123.971}{255},\,\tfrac{123.564}{255},\,\tfrac{164.785}{255}\right]$,
$\sigma=\left[\tfrac{84.271}{255},\,\tfrac{88.170}{255},\,\tfrac{72.966}{255}\right]$.

\begin{table*}[t]
\centering
\caption{\textbf{KITTI~\cite{semantickitti} vs.\ H3O (SGN~\cite{mei2023camera}).}}
\small
\resizebox{0.8\linewidth}{!}{
\begin{tabular}{lccp{4.3cm}}
\toprule
Component & KITTI~\cite{semantickitti} baseline & H3O (ours) & Rationale \\
\midrule
Grid size (GT) & $256{\times}256{\times}32$ & $64{\times}64{\times}8$ (\emph{native}) & Matched to human-scale scenes \\
Voxel size & 0.2\,m & 0.4\,m & Memory/speed under panorama \\
Bounds (x/y/z) & $[0,{-}25.6,{-}2,\,51.2,25.6,4.4]$ & $[-12.8,{-}12.8,{-}2.4,\,12.8,12.8,0.8]$ & Symmetric bounds \\
2D input & Pinhole RGB & ER panorama $608{\times}1216$ & Wider FoV under H3O \\
2D tokens/PE & Planar PE & ER tokens + spherical (lon/lat) terms & Azimuth/elevation-aware tokens \\
Lifting & X-attn: 3D$\leftarrow$2D & Same (X-attn) & Transformer lifting unchanged \\
3D queries & Per voxel (dense) & Per voxel (dense) & Same mechanism; fewer voxels \\
Classes & 20 & 11 (default remap) & default mapping on H3O \\
Losses & CE/sem\_scal/geo\_scal & Same & Consistent objectives \\
Batching & $B{=}1$ common & $B{=}2$ (single GPU) & Smaller grid $\Rightarrow$ higher throughput \\
\bottomrule
\end{tabular}
}
\end{table*}

\paragraph{H3O.}
\textbf{Grid modes.}
We support voxel grid mode with the spatial bounds $25.6{\times}25.6{\times}3.2$\,m and origin $[-12.8,-12.8,-2.4]$\,m:
\emph{native} ($64{\times}64{\times}8$ at $0.4$\,m/voxel; default).
The transformer lifting is unchanged: 3D queries at voxel centers cross-attend to ER image tokens.

\noindent\textbf{Remap, losses, and schedule.}
We adopt the default 11-class remap (\texttt{car}/\texttt{truck}/\texttt{bus} separated; traffic furniture merged), and enable the same \texttt{CE/sem\_scal/geo\_scal} losses.
Panoramas are resized to $608{\times}1216$ and normalized with statistics
($\mu{=}[0.485,0.456,0.406]$, $\sigma{=}[0.229,0.224,0.225]$).
We follow a KITTI-style schedule~\cite{semantickitti}: AdamW (LR $=10^{-4}$, WD $=10^{-4}$), MultiStepLR at epoch~20 with $\gamma=0.1$, for $30$ epochs; \texttt{batch\_size}$=2$, \texttt{num\_workers\_per\_gpu}$=12$.

\subsection{VoxFormer}
\paragraph{Scope.}
We follow \textsc{VoxFormer}~\cite{li2023voxformer}'s two-stage design and keep the model (\textsc{VoxFormer-S}) with minimal adaptations for panoramic inputs and small voxel grids. We re-implement the head/encoder hooks to support batch $\geq$1 and retrain all models from scratch. 

\paragraph{QuadOcc.}
\textbf{Data and taxonomy.}
QuadOcc provides single-frame omnidirectional RGB and voxelized semantic occupancy within
$X{=}{Y}{=}\pm 12.8$\,m and $Z\in[-2.0,\,1.2]$\,m, discretized at $0.4$\,m into a $64{\times}64{\times}8$ grid.
We train with $7$ labels (index~0 is \texttt{empty}; non-empty:
\{\texttt{vehicle}, \texttt{person}, \texttt{road}, \texttt{building}, \texttt{vegetation}, \texttt{terrain}\}).
Images are equirectangular panoramas of size $370{\times}1220$ produced by calibrated OCam/Taylor~\cite{scaramuzza2006toolbox} unwarping of a PAL camera.

\noindent\textbf{Architecture adaptations.}
\begin{compactitem}
  \item \emph{Head.} We replace the baseline head with \texttt{VoxFormerHeadQuad} (class count $20\!\rightarrow\!7$; frequency-based class weights). %
  BEV positional encoding (PE): We extensively explored two granularities that are compatible with the repository’s square-PE constraint:
  \emph{(a)} $64{\times}64$ PE for a compact $32{\times}32{\times}4$ BEV query grid (our default for fair comparisons); and
  \emph{(b)} $512{\times}512$ PE for $128{\times}128{\times}16$ queries (the largest setting we tested).
  Despite the substantially larger query sets in (b)—which increase memory and latency considerably—%
  we observed no material accuracy gains on QuadOcc/H3O. In particular, (b) frequently leads to out-of-memory on a 24\,GB RTX\,4090 even at batch size $1$ under panoramic inputs; enabling activation checkpointing/mixed precision improves stability but more than $10{\times}$ wall-clock time with negligible accuracy change. %
  Given the fairness to other transformer-based baselines (\eg, SGN~\cite{mei2023camera}) and the unfavorable compute–benefit trade-off, we \emph{report} the $64{\times}64$ PE results and regard (b) as ablations without qualitative advantage.
  \item \emph{Encoder/Layers.} We adapt reference-point generation
  to QuadOcc bounds and voxel sizes; cross/self attention settings follow the small model.
  \item \emph{Proposals.} Instead of reading proposals from \texttt{img\_metas}, we dynamically voxelize 3D points
  to obtain sparse 3D proposals (Sec.\,Stage-1 below), enabling both pseudo-LiDAR and GT-LiDAR routes.
\end{compactitem}

\noindent\textbf{Stage-1 (depth$\Rightarrow$3D proposals).}
We use two interchangeable routes to generate 3D points for proposal voxelization:
\begin{compactitem}
  \item \emph{Pseudo-LiDAR.} We run \textsc{monodepth2}~\cite{MonoDepth2} on ER panoramas and back-project with OCam/Taylor~\cite{scaramuzza2006toolbox} intrinsics to obtain per-pixel 3D points, then voxelize them into a coarse $64{\times}64{\times}8$ prior.
  \item \emph{GT-LiDAR.} We alternatively voxelize the provided LiDAR points to form the same coarse prior.
\end{compactitem}
Both routes train the identical Stage-1 network (query update via multi-head deformable cross-attention from 3D queries to ER image features). In all cases, Stage-1 outputs $64{\times}64{\times}8$ logits.

\noindent\textbf{Stage-2 (refinement).}
Stage-2 takes Stage-1 logits as priors and refines them with a shallow stack of query updates and 3D convolutions, keeping the same class set and bounds. Unless specified, reported results include the \emph{GT-LiDAR Stage-1} variant followed by Stage-2.

\noindent\textbf{Training hyper-parameters.}
We keep the VoxFormer-S recipe and adapt only what is necessary:
\begin{compactitem}
  \item \emph{Scene/voxel.} \texttt{full\_scene\_size}$=(64,64,8)$, voxel size $0.4$\,m, origin $[-12.8,-12.8,-1.2]$, eval range $25.6$\,m.
  \item \emph{Transformer (S).} embed\_dims$=128$, cross/self layers$=3/2$, sampled points per head$=8$, FFN$=1024$.
  \item \emph{Optimization.} AdamW (LR $=2{\times}10^{-4}$, WD $=10^{-2}$), CosineAnnealing with $500$ warmup iters, \texttt{total\_epochs}$=20$,
  \texttt{batch\_size}$=4$, \texttt{workers\_per\_gpu}$=4$, gradient clip (max\_norm$=35$).
  \item \emph{Losses.} We enable \texttt{CE\_ssc\_loss}, \texttt{sem\_scal\_loss}, and \texttt{geo\_scal\_loss}.
\end{compactitem}

\noindent\textbf{Image pipeline and normalization.}
Panoramas are $370{\times}1220$; standard color jitter $(0.4,0.4,0.4)$ at train-time; no val/test augmentation.
RGB normalization uses QuadOcc statistics
$\mu=\left[\tfrac{123.971}{255},\,\tfrac{123.564}{255},\,\tfrac{164.785}{255}\right]$,
$\sigma=\left[\tfrac{84.271}{255},\,\tfrac{88.170}{255},\,\tfrac{72.966}{255}\right]$.

\noindent\textbf{Class weighting.}
We compute frequency-based weights on the QuadOcc training split (non-empty voxels) for the CE term to stabilize rare dynamic classes and the head class (\texttt{road}). 

\begin{table*}[t]
\centering
\caption{\textbf{KITTI~\cite{semantickitti} vs.\ QuadOcc (VoxFormer-S~\cite{li2023voxformer}).}}
\small
\resizebox{0.8\linewidth}{!}{
\begin{tabular}{lccp{5.0cm}}
\toprule
Component & KITTI baseline & QuadOcc (ours) & Rationale \\
\midrule
Grid size (GT) & $256{\times}256{\times}32$ & $64{\times}64{\times}8$ & Embedded compute / near-field focus \\
Voxel size & 0.2\,m & 0.4\,m & Balanced radius vs.\ memory \\
BEV PE $\rightarrow$ queries & $128{\times}128{\times}16$ & $32{\times}32{\times}4$ (default); $128{\times}128{\times}16$ (ablations) & Larger PE yields heavy memory/latency with no material gains \\
Classes & 20 & 7 & Panoramic, legged-relevant labels \\
Proposal source & meta/file & voxelized (pseudo/GT) Depth & Rule out reproduction artifacts \\
Batch size & 1 & 4 & Smaller grid $\Rightarrow$ higher throughput \\
\bottomrule
\end{tabular}
}
\end{table*}

\noindent\textbf{Design rationale.}
\emph{(i) BEV positional encoding.}
We made a best-effort push to larger PE (up to $512{\times}512$) to increase the number of BEV queries and potential receptive-field coverage. However, under omnidirectional inputs and small near-field grids, the resulting memory/latency costs are disproportionately high and do not translate into consistent accuracy gains; thus we favor the $64{\times}64$ PE for fair and efficient comparison. %
\emph{(ii) Proposal voxelization (pseudo vs.\ GT).}
Pseudo-LiDAR from \textsc{monodepth2}~\cite{MonoDepth2} preserves the monocular setting but introduces depth noise that weakens sparse-query coverage; training Stage-1 with GT-LiDAR proposals improves stability yet still does not close the gap to stronger transformer baselines~\cite{mei2023camera} on QuadOcc/H3O. %
\emph{(iii) Minimal deltas from the baseline.}
We keep transformer depths/widths unchanged and only modify class hooks, reference-point generation, and proposal construction; this avoids capacity confounds and keeps results reproducible under panoramic small-volume scenes.
\paragraph{H3O.}
\textbf{Data and taxonomy.}
H3O provides single-frame omnidirectional RGB and voxelized semantic occupancy within
$X{=}{Y}{=}\pm 12.8$\,m and $Z\in[-2.4,\,0.8]$\,m, discretized at $0.4$\,m into a native $64{\times}64{\times}8$ grid.
We adopt the default 11-class remap with \texttt{empty} at index~0 and ten non-empty classes:
\{\texttt{road}, \texttt{sidewalk}, \texttt{building}, \texttt{vegetation}, \texttt{car}, \texttt{truck}, \texttt{bus}, \texttt{two\_wheeler}, \texttt{person}, \texttt{pole}\}.
Panoramas are equirectangular (ER).

\noindent\textbf{Architecture adaptations (omni variants).}
\begin{compactitem}
  \item \emph{Head.} We use \texttt{VoxFormerHeadOmni} (class count $20\!\rightarrow\!11$; frequency-based class weights).
  \item \emph{Encoder/Layers.} We operate with ER cameras and
  \emph{consume precomputed} per-voxel, per-view image correspondences from \texttt{img\_metas}:
  \texttt{projected\_pix} stores the pixel coordinates $(u,v)$ of voxel centers on each ER image, and
  \texttt{fov\_mask} flags whether a voxel lies inside the valid field-of-view. These caches:
  (i) avoid recomputation of spherical projection in the forward pass,
  (ii) faithfully adapt to the equirectangular model, and
  (iii) naturally extend to multi-camera ER inputs by concatenating per-view tokens and masks.
  Reference-point generation is adapted to the symmetric bounds $[-12.8,12.8]^2\times[-2.4,0.8]$.
  \item \emph{BEV positional encoding (PE) $\rightarrow$ query grid.}
  The repository enforces a square PE whose side equals the BEV query side. Given H3O's native $64{\times}64{\times}8$ GT,
  we evaluate two compatible query configurations:
  (a) $64{\times}64$ PE $\Rightarrow$ $32{\times}32{\times}4$ queries (default for fairness/efficiency), and
  (b) $512{\times}512$ PE $\Rightarrow$ $128{\times}128{\times}16$ queries (largest tested; ablation).
\end{compactitem}

\noindent\textbf{Stage-1 (depth$\Rightarrow$3D proposals).}
We consider two proposal routes within the H3O bounds:
\begin{compactitem}
  \item \emph{Pseudo-depth (ER).} Back-project ER depth from \textsc{MonoDepth2}~\cite{MonoDepth2} and voxelize the 3D points.
  \item \emph{Ground-truth depth (ER).} To rule out reproduction artifacts, we also voxelize proposals from \emph{ground-truth} depth maps and train \textsc{VoxFormer-S}~\cite{li2023voxformer} accordingly; we \emph{report} its accuracy alongside the pseudo-depth variant.
\end{compactitem}
Stage-1 performs multi-head deformable cross-attention from 3D queries to ER image features and outputs $64{\times}64{\times}8$ logits.

\noindent\textbf{Stage-2 (refinement).}
Stage-2 refines Stage-1 logits using a shallow stack of query updates and 3D convolutions under the same bounds/classes. Unless otherwise specified, H3O results correspond to Stage-1$\rightarrow$Stage-2.

\noindent\textbf{Training hyper-parameters.}
\begin{compactitem}
  \item \emph{Scene/voxel.} \texttt{full\_scene\_size}$=(64,64,8)$, voxel size $0.4$\,m, origin $[-12.8,-12.8,-2.4]$, eval range $25.6$\,m.
  \item \emph{Transformer (S).} embed\_dims$=128$, cross/self layers$=3/2$, sampled points per head$=8$, FFN$=1024$.
  \item \emph{Optimization.} AdamW (LR $=2{\times}10^{-4}$, WD $=10^{-2}$), CosineAnnealing with $500$ warmup iters,
  \texttt{total\_epochs}$=24$, \texttt{batch\_size}$=1$, \texttt{workers\_per\_gpu}$=4$, gradient clip (max\_norm$=35$).
  \item \emph{Losses.} \texttt{CE\_ssc\_loss}, \texttt{sem\_scal\_loss}, \texttt{geo\_scal\_loss} are enabled.
\end{compactitem}

\noindent\textbf{Image pipeline and normalization.}
Panoramas are resized to $608{\times}1216$; we use statistics for normalization
($\mu{=}[0.485,0.456,0.406]$, $\sigma{=}[0.229,0.224,0.225]$). No test-time augmentation is used.

\begin{table*}[t]
\centering
\caption{\textbf{KITTI~\cite{semantickitti} vs.\ H3O (VoxFormer-S~\cite{li2023voxformer}).}}
\small
\resizebox{0.8\linewidth}{!}{
\begin{tabular}{lccp{4.0cm}}
\toprule
Component & KITTI baseline & H3O (ours) & Rationale \\
\midrule
Grid size (GT) & $256{\times}256{\times}32$ & $64{\times}64{\times}8$ & Native near-field \\
Voxel size & 0.2\,m & 0.4 & Match grid mode and memory \\
BEV PE $\rightarrow$ queries & $128{\times}128{\times}16$ & $32{\times}32{\times}4$ (default); $128{\times}128{\times}16$ (ablation) & Square-PE constraint; larger PE = heavy cost \\
Omni projection & pinhole & ER & Deterministic sampling \\
Proposals & meta/file & voxelized (pseudo/GT) & Rule out reproduction artifacts \\
\bottomrule
\end{tabular}
}
\end{table*}

\noindent\textbf{Design rationale.}
\emph{(i) Cached ER correspondences.}
Precomputing \texttt{projected\_pix} and \texttt{fov\_mask} yields deterministic cross-attention sampling, eliminates
redundant spherical projection during the forward pass, and scales to multi-camera ER inputs—this is the key omni feature
we introduce to strengthen VoxFormer~\cite{li2023voxformer} on H3O. %
\emph{(ii) BEV PE under square-constraint.}
We pushed the BEV PE to $512{\times}512$ (\textit{i.e.}, $128{\times}128{\times}16$ queries) to maximize coverage; however, the substantial memory/latency increase did not translate into consistent accuracy gains on H3O, mirroring our QuadOcc findings. %
\emph{(iii) Depth-driven proposals (pseudo and GT).}
Using \emph{ground-truth} depth removes the confound from depth estimation noise and confirms that the performance gap is \emph{not} due to reproduction artifacts: even with GT-depth proposals and Stage-2 refinement, \textsc{VoxFormer-S}~\cite{li2023voxformer} remains markedly weaker than stronger transformer baselines (\textit{e.g.}, OccFormer~\cite{zhang2023occformer}, SGN~\cite{mei2023camera}) in our panoramic setting.
\subsection{OccFormer}
\paragraph{Scope.}
We follow \textsc{OccFormer}~\cite{zhang2023occformer} and preserve the original architecture
(2D backbone + LSS-style view transformer~\cite{LSS} + 3D encoder/decoder + Mask2Former-style head~\cite{Mask2Former} with a DETR decoder~\cite{DETR}).
To adapt to panoramic inputs and small voxel grids in QuadOcc, we only modify the dataset branch, spatial bounds,
class taxonomy, and sampling/normalization. All models are trained from scratch; the core network depths/widths remain unchanged.

\paragraph{QuadOcc.}
\textbf{Data and taxonomy.}
QuadOcc provides single-frame omnidirectional RGB and voxelized semantic occupancy within
$X{=}{Y}{=}\pm 12.8$\,m and $Z\in[-1.2,\,2.0]$\,m, discretized at $0.4$\,m into a $64{\times}64{\times}8$ grid.
We train on $7$ labels (index~0 is \texttt{unlabeled}; non-empty classes:
\{\texttt{vehicle}, \texttt{person}, \texttt{road}, \texttt{building}, \texttt{vegetation}, \texttt{terrain}\}).
Images are unfolded equirectangular panoramas; Input resolution is $(384,1280)$; We use dataset-specific normalization
(mean $[123.971,\,123.564,\,164.785]$, std $[84.271,\,88.170,\,72.966]$).

\noindent\textbf{Architecture/data adaptations.}
\begin{compactitem}
  \item \emph{Camera/input.} We switch \texttt{camera\_used} from \texttt{['left']} (KITTI~\cite{semantickitti}) to \texttt{['images\_unfold']} to consume unfolded panoramic RGB.
  \item \emph{View transformer bounds.} We set \texttt{point\_cloud\_range}$=[-12.8,-12.8,-1.2,\,12.8,12.8,2.0]$ and
  \texttt{occ\_size}$=[64,64,8]$ (voxel size $0.4$\,m). This symmetrizes the near-field volume relative to KITTI’s~\cite{semantickitti} $51.2{\times}51.2{\times}6.4$\,m.
  \item \emph{Classes.} We replace the $20$-class SemanticKITTI taxonomy with the $7$-class QuadOcc taxonomy in the head and evaluator.
\end{compactitem}

\noindent\textbf{Model components (unchanged).}
We keep the KITTI~\cite{semantickitti} baseline components: \texttt{CustomEfficientNet-B7} backbone, \texttt{SECONDFPN} image neck,
\texttt{ViewTransformerLiftSplatShootVoxel} for image$\rightarrow$voxel lifting, \texttt{OccupancyEncoder} (3D),
\texttt{MSDeformAttnPixelDecoder3D} BEV neck, \texttt{Mask2FormerOccHead}, and a $9$-layer \texttt{DetrTransformerDecoder}.

\noindent\textbf{Training hyper-parameters.}
\begin{compactitem}
  \item \emph{Optimization.} AdamW with LR $=10^{-4}$, weight decay $=10^{-2}$; gradient clipping with \texttt{max\_norm}$=20$.
  \item \emph{Schedule.} StepLR with milestones at epochs $[20,\,25]$; total $30$ epochs.
  \item \emph{Batching.} \texttt{samples\_per\_gpu}$=4$ (vs.\ 1 on KITTI~\cite{semantickitti}) and \texttt{workers\_per\_gpu}$=8$, enabled by the smaller grid.
  \item \emph{Losses.} Loss weights follow the baseline: \texttt{loss\_cls}$=2.0$, \texttt{loss\_mask}$=5.0$, \texttt{loss\_dice}$=5.0$.
  We keep class weighting consistent with the 7-class setup (background down-weighted).
\end{compactitem}

\noindent\textbf{I/O and evaluation.}
Input resolution is $(384,1280)$; we evaluate on the official QuadOcc validation split (KITTI~\cite{semantickitti} baseline uses \texttt{test}).
Unless noted, we report single-model, single-scale results without test-time augmentation.

\begin{table*}[t]
\centering
\caption{\textbf{KITTI~\cite{semantickitti} vs.\ QuadOcc (OccFormer~\cite{zhang2023occformer}).}}
\small
\resizebox{0.8\linewidth}{!}{
\begin{tabular}{lccp{4.6cm}}
\toprule
Component & KITTI~\cite{semantickitti} baseline & QuadOcc (ours) & Rationale \\
\midrule
Grid size (GT) & $256{\times}256{\times}32$ & $64{\times}64{\times}8$ & Near-field focus; compute budget \\
Voxel size & 0.2\,m & 0.4\,m & Match smaller radius, reduce tokens \\
Point-cloud range & $[0,{-}25.6,{-}2,\,51.2,25.6,4.4]$ & $[-12.8,{-}12.8,{-}1.2,\,12.8,12.8,2.0]$ & Symmetric bounds for panoramas \\
Classes & 20 & 7 & Quad taxonomy (coarser labels) \\
Camera & left (pinhole) & images\_unfold (panorama) & Fit ER/unfolded panoramas \\
Batch size & 1 & 4 & Lower res $\Rightarrow$ higher throughput \\
\bottomrule
\end{tabular}
}
\end{table*}

\noindent\textbf{Design rationale.}
\emph{(i) Symmetric near-field bounds.}
QuadOcc emphasizes a $25.6{\times}25.6{\times}3.2$\,m volume; centering and halving KITTI’s~\cite{semantickitti} bounds removes pinhole-centric bias
and avoids wasting tokens outside the useful radius. %
\emph{(ii) Dataset-specific normalization.}
Quad statistics align the unfolded ER appearance with the backbone’s expectations and improve stability.
\emph{(iii) Minimal deltas.}
We keep the architecture intact (EffNet-B7~\cite{EfficientNet} $\rightarrow$ Mask2FormerOccHead~\cite{Mask2Former}) to avoid capacity confounds; changes are isolated to
data, bounds, taxonomy, and sampling so that comparisons remain fair and reproducible.
\paragraph{H3O.}
\textbf{Data and taxonomy.}
H3O provides omnidirectional RGB and voxelized semantic occupancy within
$X{=}{Y}{=}\pm 12.8$\,m and $Z\in[-2.4,\,0.8]$\,m, discretized at $0.4$\,m into a native
$64{\times}64{\times}8$ grid. We adopt the 11-class remap with \texttt{empty} at index~0 and non-empty:
\{\texttt{road}, \texttt{sidewalk}, \texttt{building}, \texttt{vegetation}, \texttt{car}, \texttt{truck},
\texttt{bus}, \texttt{two\_wheeler}, \texttt{person}, \texttt{pole}\}. Images are panoramic ER (equidistant
cylindrical) frames; normalization uses H3O statistics (mean $[123.971,123.564,164.785]$, std $[84.271,88.170,72.966]$).

\noindent\textbf{Architecture/data adaptations.}
\begin{compactitem}
  \item \emph{Camera/input.} \texttt{camera\_used} is switched from \texttt{['left']} (KITTI~\cite{semantickitti}) to \texttt{['panorama\_rgb']} (ER panorama).
  \item \emph{View transformer.} We replace the pinhole version with \texttt{ViewTransformerLiftSplatShootVoxelH3O}
  and its ER-aware geometry routine \texttt{get\_geometry\_h3o}, which maps pixel $(u,v)$ to azimuth--elevation
  $(\theta,\phi)$ (\textit{e.g.}, $\theta=2\pi(u/W-0.5)$, $\phi=\pi\,(v/H)$) and then to 3D rays before lifting and splatting.
  \item \emph{Spatial bounds and resolution.} \texttt{point\_cloud\_range}$=[-12.8,-12.8,-2.4,\,12.8,12.8,0.8]$,
  \texttt{occ\_size}$=[64,64,8]$ (voxel size $0.4$\,m). Symmetric near-field bounds replace KITTI’s~\cite{semantickitti} forward-biased setup.
  \item \emph{Pipelines.} Training uses \texttt{CreateDepthFromLiDAR(dataset='h3o')} to provide LSS~\cite{LSS} depth supervision and
  \texttt{LoadH3OAnnotation}; evaluation uses \texttt{LoadH3OAnnotation}. 
\end{compactitem}

\noindent\textbf{Components (kept as baseline).}
\texttt{CustomEfficientNet-B7} backbone, \texttt{SECONDFPN} image neck,
\texttt{OccupancyEncoder} (3D), \texttt{MSDeformAttnPixelDecoder3D} BEV neck,
\texttt{Mask2FormerOccHead}, and a 9-layer \texttt{DetrTransformerDecoder}.

\noindent\textbf{Training hyper-parameters.}
\begin{compactitem}
  \item \emph{Optimization.} AdamW (LR $=10^{-4}$, weight decay $=10^{-2}$); gradient clipping with \texttt{max\_norm}$=20$.
  \item \emph{Schedule.} StepLR with milestones at epochs $[20,\,25]$; total $30$ epochs.
  \item \emph{Batching and sampling.} \texttt{samples\_per\_gpu}$=4$, \texttt{workers\_per\_gpu}$=8$.
\end{compactitem}

\noindent\textbf{I/O and evaluation.}
Input resolution is $(640,1280)$; we evaluate on the official \texttt{val} split of H3O and select
\texttt{save\_best='h3o\_SSC\_mIoU'}. We report single-model, single-scale results without test-time augmentation.

\begin{table*}[t]
\centering
\caption{\textbf{KITTI~\cite{semantickitti} vs.\ H3O (OccFormer~\cite{zhang2023occformer}).}}
\small
\resizebox{0.8\linewidth}{!}{
\begin{tabular}{lccp{4.9cm}}
\toprule
Component & KITTI~\cite{semantickitti} baseline & H3O (ours) & Rationale \\
\midrule
Grid size (GT) & $256{\times}256{\times}32$ & $64{\times}64{\times}8$ & Small near-field; computation budget \\
Voxel size & 0.2\,m & 0.4\,m & Match smaller radius; reduce tokens \\
Point-cloud range & $[0,{-}25.6,{-}2,\,51.2,25.6,4.4]$ & $[-12.8,{-}12.8,{-}2.4,\,12.8,12.8,0.8]$ & Symmetric bounds for 360° panoramas \\
Camera model & pinhole (left) & ER panorama & Correct geometry for 360° input \\
View transformer & LSS-voxel (pinhole) & LSS-voxel (ER) & Azimuth/elevation $\rightarrow$ 3D rays \\
cam\_channels & 33 & 27 & Match H3O camera metadata \\
Batch size & 1 & 4 & Lower res $\Rightarrow$ higher throughput \\
\bottomrule
\end{tabular}
}
\end{table*}

\noindent\textbf{Design rationale.}
\emph{(i) ER-aware lifting.} The equidistant cylindrical model avoids systematic azimuthal bias and preserves uniform
horizontal sampling; \texttt{get\_geometry\_h3o} converts $(u,v)$ to $(\theta,\phi)$ and then to 3D rays for accurate lift--splat. %
\emph{(ii) Symmetric bounds at small radius.} Centered $25.6{\times}25.6{\times}3.2$\,m is better matched to 360{\textdegree} human-scale scenes than KITTI’s~\cite{semantickitti} forward-biased frustum. %
\emph{(iii) Minimal deltas.} By keeping backbone, necks, encoder, and head identical to the baseline, improvements (or failures) can be attributed to geometry/data adaptations rather than capacity changes.
\subsection{LMSCNet}
\paragraph{Scope.}
We follow \textsc{LMSCNet}~\cite{roldao2020lmscnet}'s original 3D CNN design and keep the architecture intact (multi-scale encoder--decoder for semantic scene completion).
To adapt to panoramic inputs and the native small voxel grid in QuadOcc, we only modify channel widths and the loss, and plug in the Quad dataset/config.
All models are trained from scratch with identical depths and kernels as the baseline.

\paragraph{QuadOcc.}
\textbf{Data and taxonomy.}
QuadOcc provides single-frame semantic occupancy within
$X{=}{Y}{=}\pm 12.8$\,m and $Z\in[-2.0,\,1.2]$\,m, discretized at $0.4$\,m into a native $64{\times}64{\times}8$ grid
(\emph{implementation layout}: $(W,H,D){=}(64,8,64)$).
We train on $7$ labels (index~0 is \texttt{empty}; non-empty:
\{\texttt{vehicle}, \texttt{person}, \texttt{road}, \texttt{building}, \texttt{vegetation}, \texttt{terrain}\}).

\noindent\textbf{LiDAR input and alignment.}
Unlike monocular settings, our \textsc{LMSCNet}~\cite{roldao2020lmscnet} input is the \emph{real}, time-synchronized Livox Mid360 scan captured together with the PAL panoramic camera.
Each scan is transformed to the benchmark frame using the calibrated rigid extrinsics $T_{\text{cam}\leftarrow\text{lidar}}\!\in\!SE(3)$:
$\mathbf{X}_{\text{cam}}=T_{\text{cam}\leftarrow\text{lidar}}\mathbf{X}_{\text{lidar}}$.
We then \emph{crop} to the QuadOcc volume $X{=}{Y}{=}\pm12.8$\,m, $Z\in[-2.0,\,1.2]$\,m and \emph{voxelize} at $0.4$\,m to produce a native
$64{\times}64{\times}8$ occupancy tensor.
The origin and axes match the ground-truth convention (origin $[-12.8,-12.8,-1.2]$\,m), so that voxel indices align \emph{exactly} with the GT occupancy region.
Out-of-bounds points are discarded; extrinsics are applied consistently across all scans in a sequence.
This setup eliminates projection drift, ensures geometric consistency with the PAL camera, and provides a strong LiDAR-aligned prior for semantic scene completion.

\noindent\textbf{Architecture adaptations.}
\begin{compactitem}
  \item \emph{Channel expansion (core).} Let $C_{\mathrm{in}}$ denote the input feature channels at the encoder entrance.
  We expand the base width to $f \,{=}\, 8\,C_{\mathrm{in}}$ (baseline uses $f \,{=}\, C_{\mathrm{in}}$).
  The first encoder block is updated to \texttt{Conv2d($C_{\mathrm{in}} \!\rightarrow\! f$)}$\rightarrow$\texttt{ReLU}$\rightarrow$\texttt{Conv2d($f \!\rightarrow\! f$)}$\rightarrow$\texttt{ReLU}.
  At the decoder end, the fusion layer \texttt{conv1\_1} maps the concatenated multi-scale features back to \texttt{$C_{\mathrm{in}}$} instead of \texttt{$f$}
  to keep the output head unchanged.
  (We tried $2\times$ widening; it underperformed, hence the $8\times$ choice.)
  \item \emph{Loss.} We use \texttt{CrossEntropyLoss} for the $1{:}1$ scale.
  \item \emph{Logit alignment.} The decoder produces logits that are voxel-for-voxel aligned with the native $64{\times}64{\times}8$ GT grid
  (our tensors follow $(W,H,D)$ with $H{=}8$).
\end{compactitem}

\noindent\textbf{Training hyper-parameters.}
Unless otherwise stated, we keep the official recipe and change only what is necessary for QuadOcc:
\begin{compactitem}
  \item \emph{Dataset/config.} \texttt{QuadSSC}, grid $(W,H,D)=(64,8,64)$, \texttt{num\_classes}$=7$, \texttt{FLIPS}$=\,$\texttt{true}.
  \item \emph{Optimization.} Adam~\cite{Adam} (LR $=10^{-3}$, $\beta{=}(0.9,0.999)$), no weight decay; power-iteration LR schedule per epoch with \texttt{LR\_POWER}$=0.98$.
  \item \emph{Batching.} \texttt{train\_bs}$=8$, \texttt{val\_bs}$=8$, \texttt{num\_workers}$=8$.
  \item \emph{Schedule.} \texttt{epochs}$=80$.
\end{compactitem}

\begin{table*}[t]
\centering
\caption{\textbf{KITTI}~\cite{semantickitti} \textbf{baseline vs.\ QuadOcc (LMSCNet}~\cite{roldao2020lmscnet}\textbf{).}}
\small
\resizebox{0.8\linewidth}{!}{
\begin{tabular}{lccp{4.6cm}}
\toprule
Component & KITTI~\cite{semantickitti} baseline & QuadOcc (ours) & Rationale \\
\midrule
Grid size (GT) & $256{\times}256{\times}32$ & $64{\times}64{\times}8$ & Small near-field grid \\
Voxel size & 0.2\,m & 0.4\,m & Memory/compute budget \\
Base width $f$ & $C_{\mathrm{in}}$ & $8\,C_{\mathrm{in}}$ & Recover capacity lost by small $H{=}8$ \\
Enc.\ first conv & $f{\rightarrow}f$ & $C_{\mathrm{in}}{\rightarrow}f$ & Widen from the input stage \\
Dec.\ last conv & $\cdot{\rightarrow}f$ & $\cdot{\rightarrow}C_{\mathrm{in}}$ & Keep head I/O unchanged \\
Batch size & 2--4 & 8 & Smaller grid $\Rightarrow$ higher throughput \\
Epochs & 60--80 & 80 & Converge under widened channels \\
\bottomrule
\end{tabular}
}
\end{table*}

\noindent\textbf{Design rationale.}
\emph{(i) Width-for-depth trade-off.} QuadOcc collapses the vertical axis to $H{=}8$ bins; naive downscaling severely reduces 3D capacity.
Widening the base to $8{\times}$ restores representational power without changing depth or kernel shapes. %
\emph{(ii) Input-aware widening.} Starting from \texttt{$C_{\mathrm{in}}{\rightarrow}f$} lets the network capture richer low-level cues that are otherwise
lost on the shallow vertical dimension, while mapping back to \texttt{$C_{\mathrm{in}}$} preserves the original head. %
\emph{(iii) Minimal deltas, consistent I/O.} We keep depths/kernels unchanged, align logits exactly to the native grid, and modify only widths, thus ensuring fairness and reproducibility.
\subsection{SSCNet}
\paragraph{Scope.}
We follow \textsc{SSCNet~\cite{song2017semantic}}’s fully 3D CNN design (encoder--decoder with multi-scale skip connections) and keep depths, kernel sizes, and heads unchanged. 
To fit QuadOcc’s panoramic near-field grid, we minimally modify the upsampling head and the loss/balancing, and we plug in the Quad dataset branch. 
All models are trained from scratch.

\paragraph{QuadOcc.}
\textbf{Data and taxonomy.}
QuadOcc provides single-frame voxelized semantic occupancy within 
$X{=}{Y}{=}\pm 12.8$\,m and $Z\in[-2.0,\,1.2]$\,m, discretized at $0.4$\,m into a native $64{\times}64{\times}8$ grid 
(\emph{implementation layout}: $(W,H,D)=(64,8,64)$). 
We adopt the Quad taxonomy with \texttt{num\_classes}$=7$ for occupied voxels; \emph{free} voxels are encoded as label \texttt{255} and \emph{ignored} by the CE loss (thus not counted as a learnable class).

\noindent\textbf{LiDAR input and alignment.}
Our \textsc{SSCNet}~\cite{song2017semantic} and \textsc{SSCNet-full}~\cite{roldao2020lmscnet} variant consumes the \emph{real}, time-synchronized Livox Mid360 scan acquired together with the PAL panoramic camera.
Each scan is rigidly transformed to the benchmark/camera frame using calibrated extrinsics $T_{\text{cam}\leftarrow\text{lidar}}\!\in\!SE(3)$:
$\mathbf{X}_{\text{cam}}=T_{\text{cam}\leftarrow\text{lidar}}\mathbf{X}_{\text{lidar}}$.
We then \emph{crop} to the QuadOcc volume $X{=}{Y}{=}\pm 12.8$\,m, $Z\in[-2.0,\,1.2]$\,m and \emph{voxelize} at $0.4$\,m to form a native
$64{\times}64{\times}8$ occupancy tensor (implementation layout $(W,H,D)=(64,8,64)$).
The origin and axes follow the ground-truth convention (origin $[-12.8,-12.8,-1.2]$\,m), so that voxel indices align \emph{exactly} with the GT occupancy region.
Out-of-bounds points are discarded, and the same extrinsics are applied consistently across a sequence.
The resulting single-channel free/occupied grid serves as the network input, while the CE loss supervises the $7$ semantic classes with free voxels ignored (\texttt{label}$=255$).
This setup removes projection drift, enforces geometric consistency with the PAL rig, and provides a strong LiDAR-aligned prior for semantic scene completion.

\noindent\textbf{Architecture adaptations.}
\begin{compactitem}
  \item \emph{Deconvolution head $\rightarrow$ Upsample\,+\,Conv (stability fix).} 
  The baseline uses a transposed convolution to recover the native grid 
  (\texttt{ConvTranspose3d(128$\rightarrow$C, k=4, s=4)}). 
  We replace it with \texttt{Upsample(scale\_factor=4, mode='nearest')} followed by \texttt{Conv3d(128$\rightarrow$C, k=3, s=1, p=1)}. 
  This avoids loss-time shape/NaN issues while preserving the target stride and receptive field.
  \item \emph{Logit alignment.} 
  The decoder outputs voxel-aligned logits at the native $64{\times}64{\times}8$ resolution (our tensors follow $(W,H,D)$ with $H{=}8$).
\end{compactitem}

\noindent\textbf{Training hyper-parameters.}
\begin{compactitem}
  \item \emph{Scene/grid.} $(W,H,D)=(64,8,64)$, voxel size $0.4$\,m, \texttt{num\_classes}$=7$, free$=255$ (ignored).
  \item \emph{Optimization.} Adam~\cite{Adam} (LR $=10^{-3}$, $\beta=(0.9,0.999)$), no weight decay; constant LR schedule.
  \item \emph{Batching.} \texttt{train\_bs}$=8$, \texttt{val\_bs}$=8$, \texttt{num\_workers}$=8$.
  \item \emph{Schedule.} \texttt{epochs}$=80$.
\end{compactitem}

\begin{table*}[t]
\centering
\caption{\textbf{KITTI~\cite{semantickitti} baseline vs.\ QuadOcc (SSCNet-full~\cite{song2017semantic}).}}
\small
\resizebox{0.8\linewidth}{!}{
\begin{tabular}{lccp{4.9cm}}
\toprule
Component & KITTI~\cite{semantickitti} baseline & QuadOcc (ours) & Rationale \\
\midrule
Grid size (GT) & $256{\times}256{\times}32$ & $64{\times}64{\times}8$ (impl.\ $(64,8,64)$) & Small near-field volume \\
Voxel size & 0.2\,m & 0.4\,m & Memory/compute budget \\
Upsampling head & ConvTranspose3d ($\times 4$) & Upsample $\times 4$ + Conv3d & Avoid NaNs/shape issues; same stride \\
Labels (free) & $0$ (free), occupied $>0$ & $255$ (free), occupied $<255$ & Match Quad label semantics \\
Batch size & 4 & 8 & Lower res $\Rightarrow$ higher throughput \\
Epochs & 60--80 & 80 & Convergence at native width/depth \\
\bottomrule
\end{tabular}
}
\end{table*}

\begin{table*}[t]
\centering
\caption{\textbf{KITTI~\cite{semantickitti} baseline vs.\ QuadOcc (OccRWKV~\cite{wang2024occrwkv}).}}
\small
\resizebox{0.8\linewidth}{!}{
\begin{tabular}{lccp{4.8cm}}
\toprule
Component & KITTI~\cite{semantickitti} baseline & QuadOcc (ours) & Rationale \\
\midrule
Grid size & $256{\times}256{\times}32$ & $64{\times}64{\times}8$ & Near-field, low-memory grid \\
Voxel size & 0.2\,m & 0.4\,m & Token reduction for panoramic scenes \\
Bounds (X/Y/Z) & $[0,51.2]/{-}25.6{:}25.6/{-}2{:}4.4$ & $[0,25.6]/{-}12.8{:}12.8/{-}1.2{:}2.0$ & Match Quad near-field convention \\
Loss & CE $+$ aux.\ seg & CE (w/ class weights) $+$ Lovasz & Stable single-head supervision \\
Batch size & 2 & 2 & Compute budget \\
Epochs & 80 & 80 & Same convergence horizon \\
\bottomrule
\end{tabular}
}
\vskip -1ex
\end{table*}

\noindent\textbf{Design rationale.}
\emph{(i) Numerically stable upsampling.} 
Replacing deconvolution with Upsample\,+\,Conv preserves the $\times4$ stride yet eliminates checkerboard artifacts and loss-time instability observed with transposed convolutions on small $H{=}8$ grids. %
\emph{(ii) Minimal deltas, aligned outputs.} 
Keeping the encoder/decoder unchanged and aligning logits exactly to the native grid ensures that performance differences come from data/label semantics and small-grid geometry rather than capacity shifts. 

\subsection{OccRWKV}
\paragraph{Scope.}
We follow the original \textsc{OccRWKV}~\cite{wang2024occrwkv} pipeline (point-cloud preprocessing $\rightarrow$ BEV UNet branch $\rightarrow$ 3D completion branch, with RWKV-style recurrent blocks) and keep depths/kernels unchanged. 
To fit panoramic near-field grids in QuadOcc, we introduce Quad-specific modules, LiDAR-driven inputs, explicit voxel coordinate ordering, and a simplified loss. 
All models are trained from scratch.

\paragraph{QuadOcc.}
\textbf{Data and taxonomy.}
QuadOcc provides single-frame voxelized semantic occupancy within the near-field bounds
$X\!\in\![0,\,25.6]$\,m, $Y\!\in\![-12.8,\,12.8]$\,m, $Z\!\in\![-1.2,\,2.0]$\,m, discretized at $0.4$\,m into a native 
$64{\times}64{\times}8$ grid. We adopt $7$ semantic classes (non-empty) and treat free space as \texttt{255} (ignored by CE).

\noindent\textbf{LiDAR input and alignment.}
Our \textsc{OccRWKV}~\cite{wang2024occrwkv} variant consumes the \emph{real}, time-synchronized Livox Mid360 scan captured together with the PAL panoramic camera. 
Each scan is rigidly transformed to the benchmark/camera frame via calibrated extrinsics $T_{\text{cam}\leftarrow\text{lidar}}\!\in\!SE(3)$:
$\mathbf{X}_{\text{cam}}=T_{\text{cam}\leftarrow\text{lidar}}\mathbf{X}_{\text{lidar}}$. 
We \emph{crop} to the QuadOcc volume, \emph{voxelize} at $0.4$\,m to obtain the native $64{\times}64{\times}8$ occupancy tensor, and align the origin/axes to the GT convention. 
Out-of-bounds points are discarded, and the same extrinsics are applied consistently per sequence.

\noindent\textbf{Architecture/data adaptations.}
\begin{compactitem}
  \item \emph{LiDAR-driven inputs.} The network ingests point clouds from \texttt{PCD['1\_1']} and dense occupancy from \texttt{3D\_OCCUPANCY}; labels are read from \texttt{3D\_LABEL['1\_1']}.
  \item \emph{Occupancy$\,+$BEV fusion.} We treat the $D{=}8$ occupancy slices as channels (\,$B{\times}D{\times}H{\times}W$\,) and concatenate them with BEV features (\,$B{\times}C_{\text{bev}}{\times}H{\times}W$\,) along the channel dimension, yielding a $B{\times}(D{+}C_{\text{bev}}){\times}H{\times}W$ tensor for the 2D BEV UNet.
  \item \emph{Loss.} We remove auxiliary semantic-point loss and keep a \emph{single} voxel-wise objective: class-weighted cross-entropy (CE) with \texttt{ignore\_index=255} plus \texttt{Lovasz-Softmax}. 
  Final logits are permuted to $(B, C, W, D, H)$ as required by the head before loss.
\end{compactitem}

\noindent\textbf{Training hyper-parameters.}
\begin{compactitem}
  \item \emph{Dataset/config.} \texttt{QuadOcc}; grid $[W,H,D]=[64,64,8]$; voxel size $0.4$\,m; bounds as above; \texttt{FLIPS=true}; modalities: \texttt{PCD}, \texttt{3D\_OCCUPANCY}, \texttt{3D\_LABEL}.
  \item \emph{Optimization.} Adam~\cite{Adam} (LR $=10^{-3}$, $\beta=(0.9,0.999)$), no weight decay; power-iteration schedule per epoch with \texttt{LR\_POWER}$=0.98$.
  \item \emph{Batching/schedule.} \texttt{batch\_size}$=2$, \texttt{num\_workers}$=8$, \texttt{epochs}$=80$.
\end{compactitem}

\noindent\textbf{Design rationale.}
\emph{(i) Simplified loss.}
Dropping the auxiliary semantic segmentation loss branch focuses capacity and stabilizes training; adding Lovasz complements CE on class-imbalanced voxels. %
\emph{(v) Minimal deltas.}
Network depths/widths are unchanged; improvements (or failures) can thus be attributed to geometry/data choices rather than capacity changes.
\section{Discussions}

\subsection{Limitations and Potential Solutions}

\paragraph{Spatial resolution and fine-grained interaction.}
Our semantic occupancy is defined on a $64{\times}64{\times}8$ grid with $0.4$\,m voxels around the ego, which is sufficient for navigation, foothold selection, and path planning on legged/humanoid platforms with moderate speed and limited payload compared to intelligent vehicles. However, this resolution is not ideal for tasks that require fine-grained contact reasoning, such as precise grasping, object re-arrangement, or manipulation in cluttered shelves. At this scale, small objects and thin structures are often represented by only a few voxels, which amplifies label noise and makes it difficult to capture geometry with sub-voxel accuracy.

A straightforward remedy is to increase the grid resolution (\textit{e.g.}, $128{\times}128{\times}16$ under the same metric bounds), but our resolution study shows that this leads to a significant increase in memory and latency, and can even hurt mIoU due to optimization difficulty. Instead, a more promising direction is \emph{hierarchical} or \emph{adaptive} occupancy. One option is to maintain a coarse global grid (as in OneOcc) for full-surround situational awareness, while allocating high-resolution local volumes around the end-effectors (hands and feet) or task-relevant regions selected by an attention or proposal mechanism. Another option is to decouple geometry and semantics: a coarse semantic grid can be refined locally by an implicit or Gaussian-based representation~\cite{3dgs} that super-resolves surfaces where higher accuracy is required. Both strategies preserve the efficiency and robustness of the current pipeline while opening the door to fine-grained interaction.

\paragraph{Dataset coverage and sim-to-real gap.}
QuadOcc is moderate-scale and rooted in a campus-like environment, and Human360Occ (H3O) is purely simulated. While this pairing is useful for studying cross-domain robustness and robot morphology changes, it does not fully cover the diversity of real humanoid deployment scenarios (\textit{e.g.}, dense indoor offices, homes, or highly dynamic crowds). In particular, our humanoid-style evaluation currently relies on simulated occupancy labels and thus inherits the biases of the simulator and rendering stack. This introduces an inevitable sim-to-real gap for humanoid robots with panoramas.

Mitigating this gap will require a combination of larger-scale real-world 360{\textdegree} occupancy datasets (\textit{e.g.}, panoramic rigs mounted on humanoids or human operators) and sim-to-real adaptation techniques. Possible options include domain randomization and style augmentation in the panoramic image space, occupancy-level adversarial training between simulated and real volumes, and semi-supervised self-training where the model bootstraps dense occupancy from sparse or partial real sensors. Another complementary direction is to leverage weak labels, such as traversability or contact affordances, as auxiliary supervision for regions where dense voxel labels are missing.

\paragraph{Single-frame design and motion modeling.}
Our current formulation predicts occupancy from a single panoramic frame. This design choice avoids hard assumptions on the robot's motion model, simplifies deployment, and is already robust to gait-induced jitter via GDC. However, it also limits the ability to accumulate evidence over time, to explicitly reason about dynamic objects, and to exploit long-term temporal context. In particular, fast-moving agents (pedestrians, vehicles, other robots) are treated as static at each frame, and any temporal consistency emerges only implicitly from the training data.

A natural extension is to move from single-frame inputs to short panoramic clips, and to lift them into a 4D spatio-temporal occupancy volume. Architecturally, this could be realized by adding a recurrent~\cite{shi2024beyond} or transformer-style temporal module on top of OneOcc, or by introducing causal 3D/4D convolutions over occupancy sequences. To keep the embodied footprint manageable, one promising strategy is to maintain a low-rate, wide-range occupancy memory complemented by higher-rate local updates near the robot, rather than processing dense video at full resolution.

\subsection{Failure Case Analysis}
\begin{figure}[t]
  \centering
  \includegraphics[width=\linewidth]{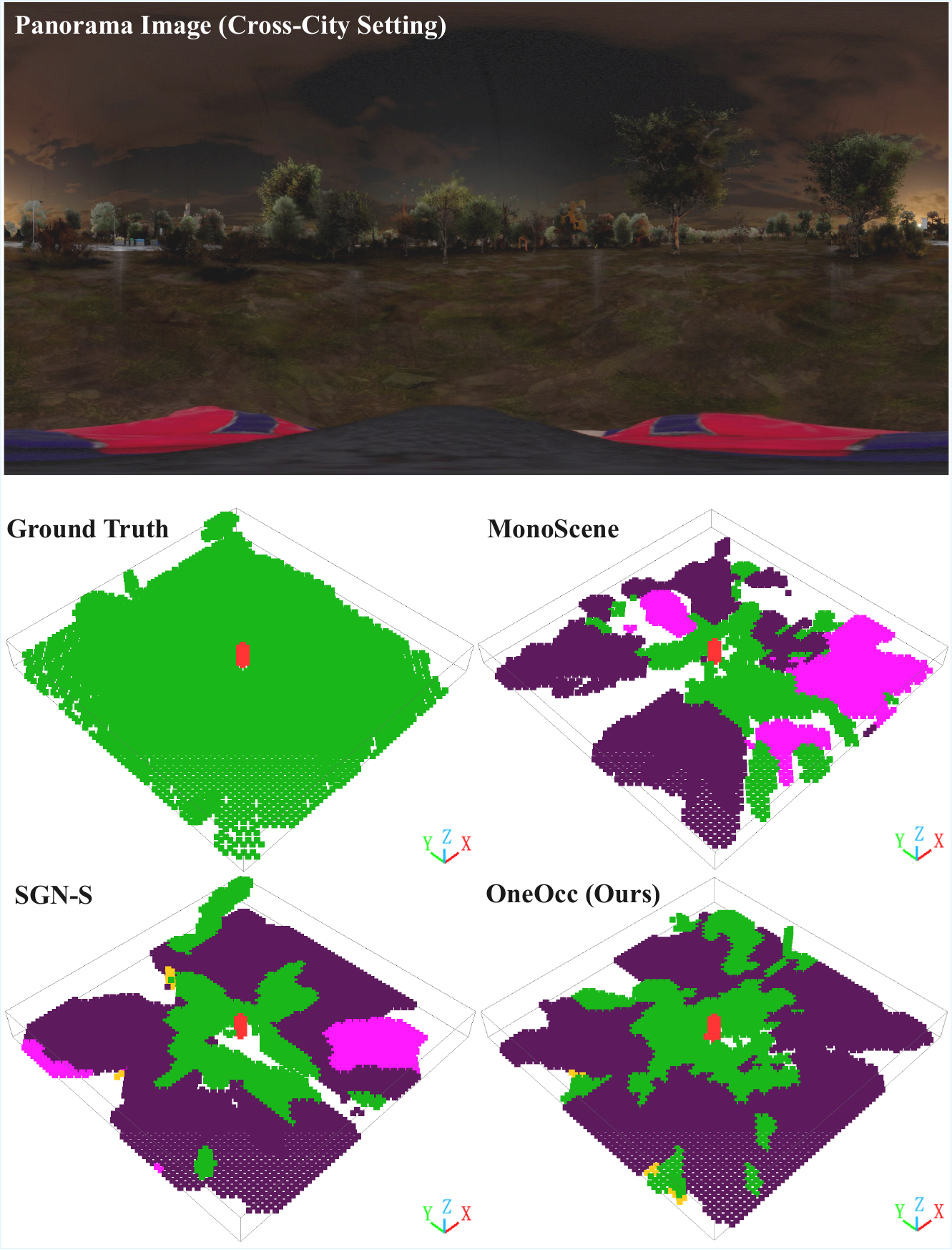}
  \caption{\textbf{Failure case on H3O-Heter (cross-city) in a rainy dusk scene dominated by vegetation.} From left to right: ground-truth occupancy, predictions from MonoScene~\cite{cao2022monoscene}, SGN-S~\cite{mei2023camera}, and OneOcc (Ours). OneOcc alleviates but does not eliminate severe semantic misclassification (vegetation predicted as sidewalk) under this rare combination of layout, weather, and time-of-day.}
  \label{fig:failure-case}
\end{figure}

Figure~\ref{fig:failure-case} illustrates a representative failure case from the challenging H3O-Heter split, where the robot traverses a corridor that is almost entirely covered by vegetation and spans multiple maps under the cross-city configuration. This scene is further compounded by adverse conditions: it is captured at dusk, in rain, with low illumination and strong specular highlights on wet surfaces. While our training data already covers diverse weather and lighting conditions, sequences that are simultaneously \emph{vegetation-dominated}, \emph{cross-map}, and \emph{rainy at dusk} are extremely rare, and most trajectories follow road- or sidewalk-centric layouts in clearer visibility. As a result, all methods face a substantial distribution shift in both geometry and appearance.

In this example, the ground-truth occupancy assigns the majority of near-field voxels to \emph{vegetation} (class~4, green), and virtually no large dynamic objects such as \emph{car} (class~5), \emph{truck} (class~6), or \emph{bus} (class~7). In contrast, vision-based baselines (\eg, MonoScene~\cite{cao2022monoscene}, SGN-S~\cite{mei2023camera}) hallucinate extensive road and sidewalk regions in front of the agent and along the sides. These errors reflect a strong reliance on co-occurrence priors learned from more typical urban scenes: given a forward-looking corridor with lane-like edges and headlight-like highlights, the models default to ``road + sidewalk'' even when the geometry and appearance actually correspond to wet vegetation. OneOcc substantially reduces these artifacts: the free/occupied structure is better aligned with the ground truth, and most spurious roads are removed, yet it still misclassifies sizable vegetation regions as sidewalk, leading to severe semantic errors despite reasonably accurate geometry.

This failure pattern highlights two important insights. First, even with improved architectural bias and stronger cross-city robustness, semantic occupancy prediction remains fundamentally data-driven: models inherit biases from the training distribution and struggle when encountering rare combinations of layout, weather, and time-of-day. Closing this gap will likely require larger-scale and more diverse 360$^\circ$ occupancy datasets that explicitly target under-represented environments (\textit{e.g.}, forests, parks, off-road trails) under diverse conditions (night, adverse weather, seasonal changes), as well as more balanced class statistics. Second, the dominant error here is \emph{semantic} rather than geometric: the model correctly infers that space is occupied, but assigns the wrong category. This suggests that future \emph{open-vocabulary occupancy}~\cite{tan2023ovo} formulations, where volumetric geometry is coupled with flexible semantic embeddings instead of a fixed closed set of labels, could be beneficial. By leveraging open-set vision-language representations, such models may better adapt their semantic partition of the occupancy volume to novel environments (for example, distinguishing different types of vegetation or terrain) while reusing the same geometric backbone as OneOcc. Combined with stronger data diversity, such open-vocabulary occupancy could mitigate, though likely not completely eliminate, the kind of semantic failure observed in this case.

\begin{figure*}[t]
    \centering
    \includegraphics[width=\textwidth]{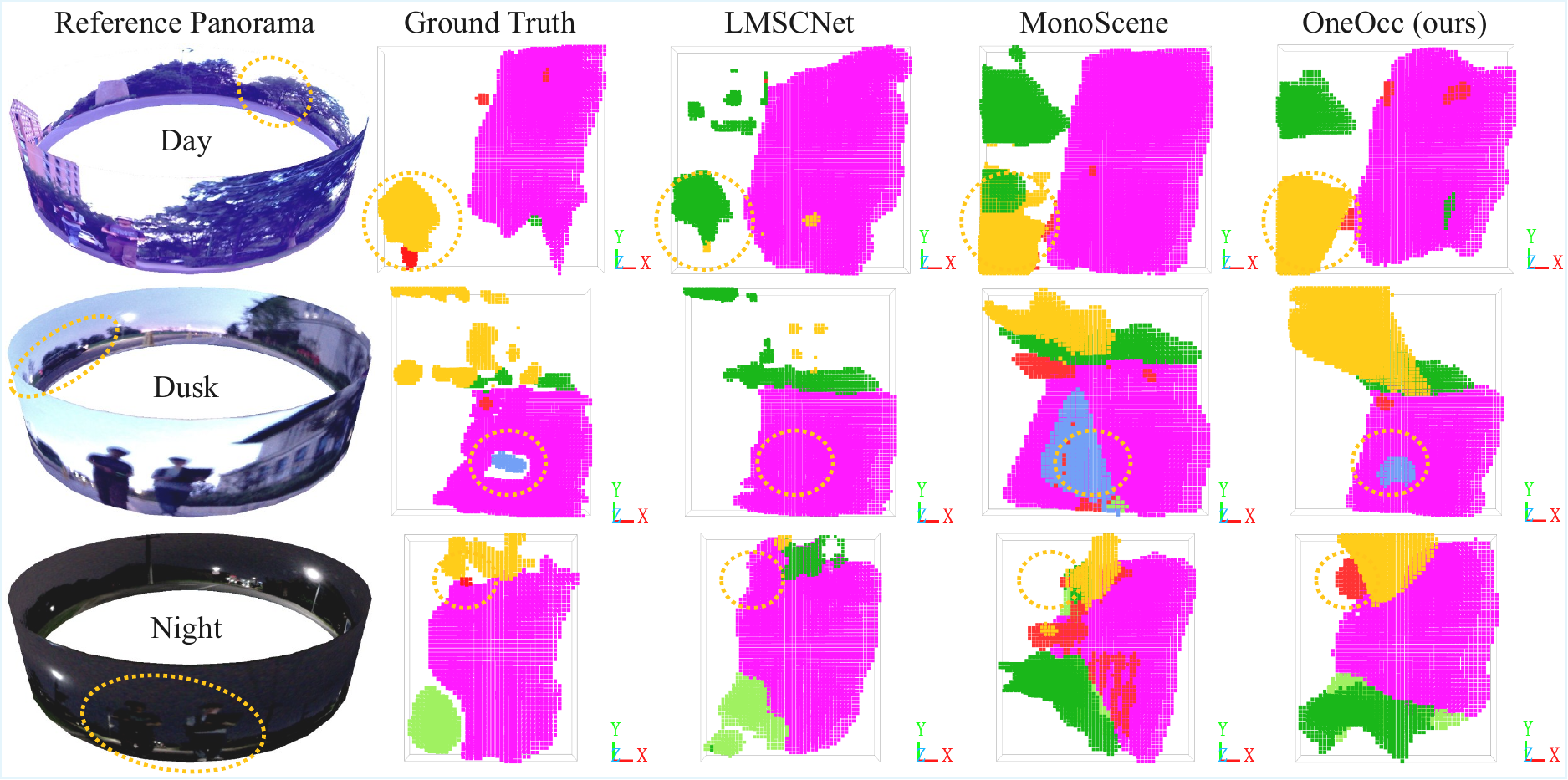}
    \caption{\textbf{Qualitative comparison on QuadOcc under different times of day.}
    From left to right: reference panorama, ground-truth occupancy, LMSCNet~\cite{roldao2020lmscnet}, MonoScene~\cite{cao2022monoscene}, and OneOcc (ours). From top to bottom: day, dusk, and night.
    All methods gradually degrade from day to dusk and further to night, where low illumination and glare introduce severe ambiguities.
    Compared to baselines, OneOcc reduces perspective sampling artifacts and yields more coherent occupied structures and semantics, especially for dynamic classes such as vehicles and pedestrians.}
    \label{fig:quad-timeofday-qual}
\end{figure*}

\subsection{Range-wise Safety Analysis}
\label{sec:dist_binned_analysis}

To further examine whether the gains of OneOcc are confined
to a particular depth interval or remain valid in
safety-critical local regions, we report distance-binned
mIoU comparisons against MonoScene~\cite{cao2022monoscene}
on both QuadOcc and H3O-Heter.
Following the same evaluation protocol as the main paper,
we restrict non-empty voxels to three nested spatial ranges
centered at the agent: \emph{Far} ($\pm 12.8$m),
\emph{Mid} ($\pm 6.4$m), and \emph{Near} ($\pm 3.2$m).

\begin{figure}[t]
\centering
\includegraphics[width=\linewidth]{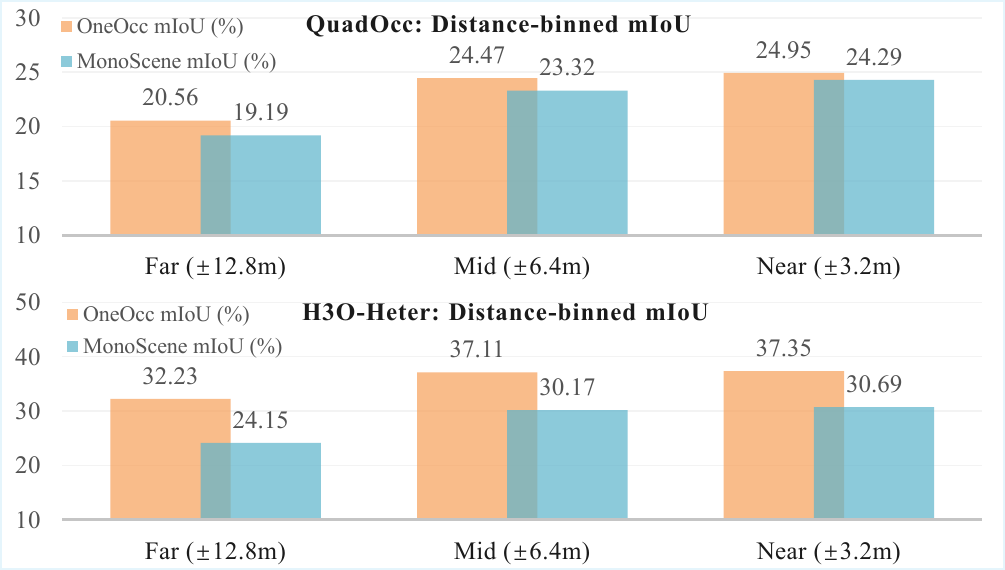}
\vskip -1.5ex
\caption{Distance-binned mIoU comparisons on QuadOcc and H3O-Heter.
We evaluate non-empty voxels within three nested spatial ranges centered at the agent:
Far ($\pm 12.8$m), Mid ($\pm 6.4$m), and Near ($\pm 3.2$m).
OneOcc consistently outperforms MonoScene~\cite{cao2022monoscene}
across all ranges on both benchmarks, including the near-field region that is most relevant to local navigation and foothold safety.}
\label{fig:dist_binned}
\vskip -1.0ex
\end{figure}
Figure~\ref{fig:dist_binned} shows that OneOcc consistently
outperforms MonoScene across all distance bins on both
benchmarks.
On QuadOcc, the mIoU improves from 19.19 to 20.56 in the
Far range, from 23.32 to 24.47 in the Mid range, and from
24.29 to 24.95 in the Near range.
On H3O-Heter, the corresponding gains are larger:
24.15 $\rightarrow$ 32.23 (Far),
30.17 $\rightarrow$ 37.11 (Mid),
and 30.69 $\rightarrow$ 37.35 (Near).
These results indicate that the benefits of OneOcc are not
limited to a specific distance regime, but remain stable from
long-range completion to local occupancy reasoning.
From a robotics perspective, the Near range is especially
important because it is most directly related to short-horizon
collision avoidance, foothold selection, and local traversability.
Importantly, OneOcc still maintains positive margins in this
region on both datasets: +0.66 mIoU on QuadOcc and
+6.66 mIoU on H3O-Heter.
This suggests that the proposed design does not merely improve
far-field continuity or suppress panoramic artifacts at a global
scale, but also provides more reliable local semantic occupancy
in the region most relevant to embodied safety.
A second trend is that both methods generally improve from
Far to Near.
This is expected, since smaller spatial ranges contain less
severe long-range ambiguity, fewer occlusion-induced completion
errors, and denser effective evidence around the agent.
However, the gap between OneOcc and MonoScene remains
consistently positive throughout.
This supports the claim that the gains of OneOcc arise from
better geometry-aware lifting and panoramic representation,
rather than from overfitting to a narrow subset of voxels or
to only far-range context.
Overall, the distance-binned analysis complements the main
benchmark tables by showing that OneOcc improves semantic
occupancy prediction not only globally, but also in the
near-field regime that is most critical for safe embodied
deployment.

\subsection{Scaling to Other Scenarios and Modalities}

\paragraph{From legged to diverse robot morphologies.}
Although QuadOcc and H3O are designed for quadruped and humanoid-like settings, the core architectural ideas in OneOcc --- dual-projection fusion, bi-grid voxelization, and lightweight 3D decoding --- are not specific to a particular robot morphology. For wheeled platforms with forward-biased sensing, the same framework can be used with asymmetric voxel bounds that allocate more range in the forward direction while preserving 360{\textdegree} continuity, or with cropped panoramas when only 180{\textdegree} coverage is available. For aerial robots or platforms with elevated cameras, the vertical bounds of the voxel grid can be expanded to better capture tall structures and multi-level environments.

\paragraph{Indoor, outdoor, and extreme conditions.}
Our datasets focus on outdoor campus-style and urban-like scenes. Deployments in cluttered indoor environments introduce different statistics: shorter ranges, denser occlusions, and more small-scale objects. Scaling OneOcc to such scenarios mainly requires reconfiguring the voxel bounds and semantics, and retraining on appropriate indoor 360{\textdegree} data. Meanwhile, operation under extreme conditions (severe rain/snow, low light, lens contamination) may benefit from explicit robustness techniques such as test-time adaptation~\cite{shi2025offboard}, uncertainty-aware occupancy heads, or sensor health monitoring that can gracefully degrade occupancy outputs when the panoramic signal is unreliable.

\paragraph{Multi-modal extensions.}
OneOcc is intentionally vision-only, which is attractive for platforms with strict payload and power budgets. Nevertheless, in settings where additional sensors are available, the proposed architecture can serve as a backbone for multi-modal occupancy. LiDAR or depth maps can be encoded into 2D or 3D features and fused with the panoramic features before or after bi-grid voxelization, while radar and event cameras~\cite{guo2025event} can complement the panoramic input with long-range or high-dynamic-range motion cues. A promising direction is to treat each modality as an additional ``view'' in the View2View sampling scheme, letting the model learn how to combine them into a unified 3D grid without hard-coded fusion rules.

\begin{figure*}[!t]
    \centering
    \includegraphics[width=\textwidth]{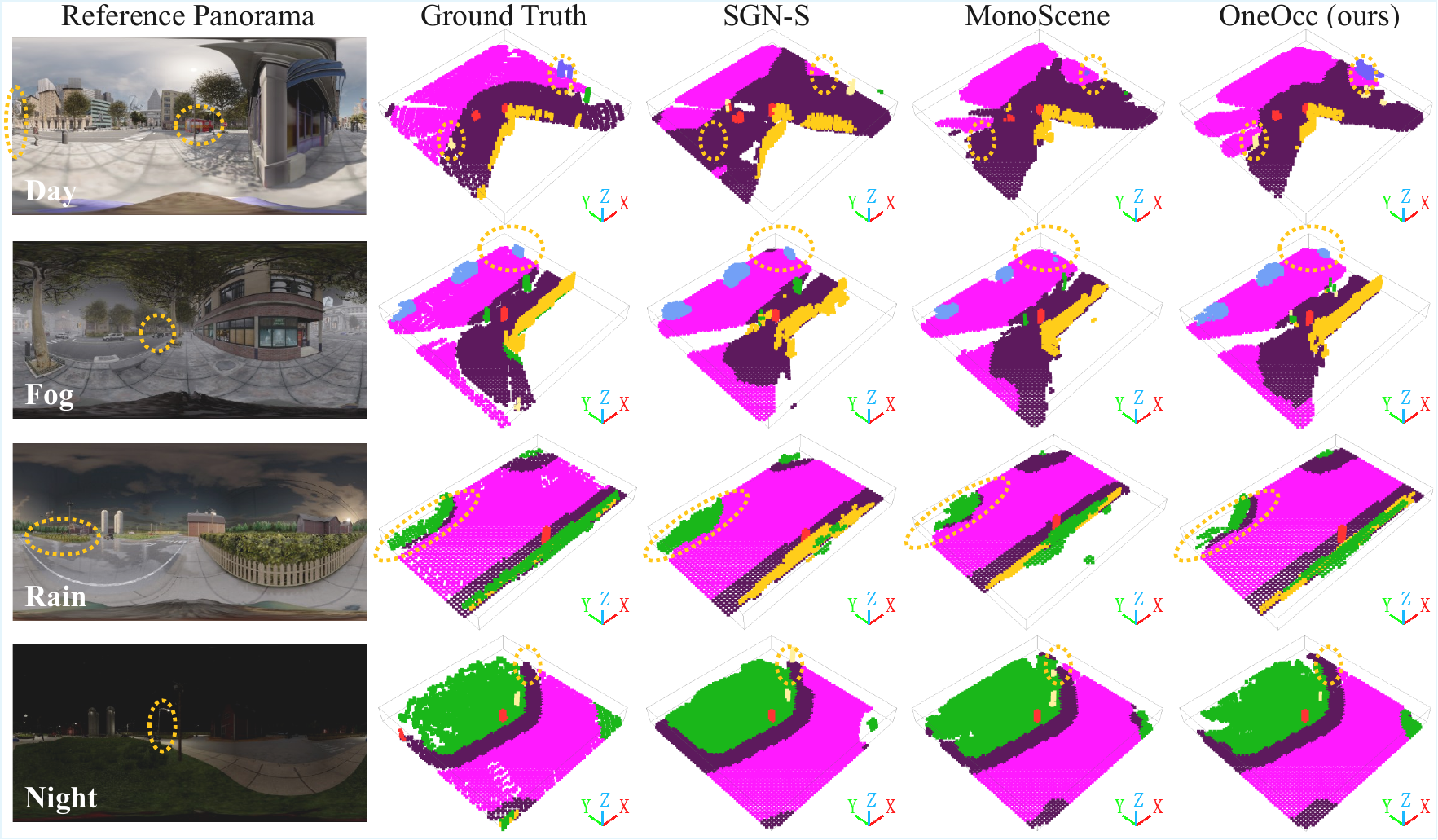}
    \caption{\textbf{Qualitative comparison on H3O-Heter under diverse weather and illumination.}
From left to right: reference panorama, ground-truth occupancy, SGN-S~\cite{mei2023camera}, MonoScene~\cite{cao2022monoscene}, and OneOcc (ours). From top to bottom: clear day, fog, rain, and night.
The heter split introduces strong cross-map distribution shift and long-tailed conditions; the performance of all methods degrades under fog/rain and becomes the most brittle at night.
Notably, the circled distant lamp post is outside the OCC sensing range in the ground truth (the Night row).
SGN-S and MonoScene overestimate their depth and mistakenly project it into the occupancy volume, whereas OneOcc avoids this long-range depth hallucination and yields fewer false positives, while still recovering thin structures (\textit{e.g.}, poles) within range.}

    \label{fig:h3o-weather-qual}
\end{figure*}

\subsection{Societal Impacts}

Dense 3D semantic occupancy opens up positive applications for legged and humanoid robots. More reliable 360{\textdegree} perception can improve safety when robots operate in proximity to humans, reduce collisions with obstacles in cluttered environments, and enable new capabilities in search-and-rescue, inspection, and assistive robotics. In such contexts, a geometry-aware volumetric representation is often more interpretable and task-aligned than raw images, and can form a safer intermediate layer for downstream planning or language-conditioned policies.

At the same time, always-on full-surround perception could raise some societal questions. A robot equipped with a panoramic camera and a strong occupancy predictor could be used for pervasive surveillance or long-term monitoring of public or semi-private spaces. While our representation focuses on occupancy and coarse semantics rather than identity, it still encodes where humans and vehicles are likely to be, and could be combined with other modules to infer more sensitive attributes. Moreover, our datasets are biased toward specific geographies, weather patterns, and human behaviors, which may lead to failure modes or unfair performance in underrepresented environments. Responsible deployment will require transparency about sensing capabilities, adherence to local privacy regulations, careful consideration of where such robots are allowed to operate, and continued auditing of failure cases, especially near vulnerable populations.

\subsection{Future Work}

Beyond the immediate extensions discussed above, we see three main research directions.

\paragraph{Temporal occupancy, flow, and world models.}
First, moving from static occupancy to temporal occupancy sequences is a key step toward \emph{world models}~\cite{zheng2024occworld,zuo2025gaussianworld,feng2025gaussian} for legged and humanoid robots. Instead of predicting only the current occupancy, one can learn to forecast future volumes conditioned on past panoramas and robot actions, or to infer an \emph{occupancy flow}~\cite{liu2024let,chen2025alocc} that encodes voxel-wise 3D motion over time. Such a flow field would provide a sparse, geometry-aware analogue of scene flow, tailored to dynamic obstacles and multi-contact locomotion. Training such models will likely require a mix of supervised signals (from synthetic data or reconstructed trajectories) and self-supervised objectives that enforce temporal consistency and physical plausibility.

\paragraph{Adaptive resolution and manipulation-centric perception.}
Second, we plan to investigate adaptive-resolution schemes that bridge the gap between navigation-centric occupancy and manipulation-centric understanding. This includes foveated occupancy around hands and feet, dynamic refinement of regions with high uncertainty or contact likelihood, and hybrid explicit--implicit models where a coarse grid is locally refined into meshes or signed distance fields for grasp synthesis and force reasoning. A tight coupling between such perception modules and whole-body control could enable humanoid robots to seamlessly transition from long-range navigation to precise object interaction.

\paragraph{Data, generalization, and policy integration.}
Finally, scaling to broader real-world deployment will require richer data and tighter integration with decision-making. On the data side, we advocate for collecting real 360$^\circ$ occupancy datasets on humanoid and other platforms, with diverse environment types and behaviors, and for exploring self-supervised pretraining on unlabeled panoramic video. On the policy side, occupancy grids can act as an intermediate tokenization layer for vision-language-action models and reinforcement learning policies, allowing language instructions and low-level control to be grounded in a common geometric space. Jointly training such policies and occupancy predictors, while preserving modularity and interpretability, is an exciting avenue toward more capable and trustworthy embodied agents.

\section{More Visualizations}

Figures~\ref{fig:quad-timeofday-qual} and~\ref{fig:h3o-weather-qual} provide additional qualitative comparisons between OneOcc and representative vision-based SSC baselines under diverse illumination and weather conditions. On QuadOcc, we report results for daytime, dusk, and nighttime scenes. Across all methods, the prediction quality degrades noticeably as the illumination decreases: daytime scenes yield the cleanest geometry and semantics, dusk introduces stronger ambiguity around boundaries and small objects, and nighttime remains consistently the most challenging setting due to low signal-to-noise ratio, headlight glare, and reduced texture cues. In particular, existing perspective-based lifting methods~\cite{cao2022monoscene} often suffer from pronounced projection artifacts and over-smoothed structures at dusk and night, leading to inconsistent free/occupied patterns and spurious semantic regions (\textit{e.g.}, hallucinated road/sidewalk blobs). OneOcc exhibits improved robustness along this axis: by leveraging panoramic View2View fusion and bi-grid voxelization, it mitigates the typical perspective-sampling artifacts observed in FLOSP-style pipelines, and preserves more coherent occupied structures and semantic layouts at dusk and night. Notably, dynamic categories such as cars and pedestrians are more reliably localized, and their occupied footprints are less fragmented compared to MonoScene~\cite{cao2022monoscene} and LiDAR-based LMSCNet~\cite{roldao2020lmscnet}.

On H3O-Heter, we further evaluate generalization under cross-map scenes and adverse weather, including clear daytime, fog, rain, and nighttime. Similar to QuadOcc, all methods deteriorate as the environment becomes more visually ambiguous. Fog reduces contrast and texture, causing baselines to miss thin vertical structures and to blur object boundaries; rain introduces specular reflections and appearance shifts that amplify road-centric priors; nighttime again triggers the largest failures. While baselines frequently collapse thin objects into the background or misclassify them into dominant terrain classes, OneOcc retains sharper geometric contours and more faithful semantics across conditions.
We additionally observe a characteristic depth overestimation error of prior methods in the heter split. As highlighted in Fig.~\ref{fig:h3o-weather-qual}, a distant lamp post (pole) lies beyond the predefined OCC sensing range in the ground truth. However, SGN-S~\cite{mei2023camera} and MonoScene~\cite{cao2022monoscene} incorrectly estimate its depth and ``pull'' the pole into the occupancy volume, generating spurious occupied voxels. In contrast, OneOcc does not introduce such false positives, suggesting that the panoramic View2View fusion and bi-grid lifting lead to more calibrated long-range depth reasoning under cross-map distribution shift.

Overall, these results corroborate our quantitative findings: OneOcc maintains stronger structural consistency and semantic fidelity under both illumination and weather shifts, though extreme low-light conditions remain an open challenge for monocular panoramic SSC.

\end{document}